\documentclass[a4paper]{article}
\usepackage[a4paper,top=3cm,left=3cm,right=2cm,bottom=2cm]{geometry}
\usepackage{amsmath}
\usepackage{amsthm}
\usepackage{graphicx}
\usepackage{natbib}
\usepackage{algorithm}
\usepackage{setspace}
\usepackage{fancyhdr}
\usepackage{amssymb}
\usepackage{setspace}
\usepackage[usenames,dvipsnames]{xcolor}

\usepackage{lscape}
\usepackage{graphics}
\usepackage{graphicx}
\usepackage{caption}
\usepackage{subcaption}
\usepackage[noend]{algpseudocode}
\usepackage{longtable}
\usepackage{multirow}
\usepackage{url}
\usepackage{array}
\usepackage{tikz}
\usepackage{rotating}
\usepackage{morefloats}
\usepackage{colortbl}
\usepackage[normalem]{ulem}
\usepackage[titletoc,title]{appendix}
\usepackage{hyperref}
\usepackage[vertfit]{breakurl} 

\hypersetup{
	colorlinks=true,
	linkcolor=blue,
	citecolor=blue,
	urlcolor=blue
}

\usetikzlibrary{patterns}
\usetikzlibrary{arrows}
\newcolumntype{L}[1]{>{\raggedright\let\newline\\\arraybackslash\hspace{0pt}}m{#1}}
\newcolumntype{C}[1]{>{\centering\let\newline\\\arraybackslash\hspace{0pt}}m{#1}}
\newcolumntype{R}[1]{>{\raggedleft\let\newline\\\arraybackslash\hspace{0pt}}m{#1}}

\newcommand{\up}[1]{\raisebox{1.3ex}[0pt]{#1}}



\begin{document}
\sloppy




\large
\title{A unified heuristic and an annotated bibliography for a large class of earliness-tardiness scheduling problems}

\author{{\bf Arthur Kramer, Anand Subramanian}\\
Departamento de Engenharia de Produ\c{c}\~ao\\ Universidade Federal da Para{\'i}ba, Brazil\\
arthurhfrk@gmail.com, anand@ci.ufpb.br
}

\date{}

\maketitle

\vspace{-0.5cm}
\begin{center}
Working Paper, UFPB -- December 2016 \\
\end{center}
\vspace{0.5cm}

\begin{abstract}
This work proposes a unified heuristic algorithm for a large class of earliness-tardiness (E-T) scheduling problems. We consider single/parallel machine E-T problems that may or may not consider some additional features such as idle time, setup times and release dates. In addition, we also consider those problems whose objective is to minimize either the total (average) weighted completion time or the total (average) weighted flow time, which arise as particular cases when the due dates of all jobs are either set to zero or to their associated release dates, respectively. The developed local search based metaheuristic framework is quite simple, but at the same time relies on sophisticated procedures for efficiently performing local search according to the characteristics of the problem. We present efficient move evaluation approaches for some parallel machine problems that generalize the existing ones for single machine problems. The algorithm was tested in hundreds of instances of several E-T problems and particular cases. The results obtained show that our unified heuristic is capable of producing high quality solutions when compared to the best ones available in the literature that were obtained by specific methods. Moreover, we provide an extensive annotated bibliography on the problems related to those considered in this work, where we not only indicate the approach(es) used in each publication, but we also point out the characteristics of the problem(s) considered. Beyond that, we classify the existing methods in different categories so as to have a better idea of the popularity of each type of solution procedure.

\end{abstract}

%
%
%
\onehalfspace


\section{Introduction}
\label{sec:Introduction}

Scheduling problems have been widely studied in the literature over the past 50 years, receiving considerable attention from many scholars and practitioners around the world.
Motivated by complex real-life problems faced by different types of companies, as well as by the challenge in solving them, a huge number of scheduling variants and solution procedures were proposed during this half-century period. Hence, it is safe to say that \emph{Scheduling} is one of the most important subjects in the fields of Operations Research and Management Science.

A particular branch of the scheduling theory arises in the context of Just-In-Time (JIT) manufacturing systems. Problems of this branch of scheduling are generally referred to as earliness-tardiness (\mbox{E-T}) scheduling problems, where penalties are incurred if a job is completed before or after its due date. Moreover, they often include several additional features commonly found in the scheduling literature such as multiple machines (identical, uniform, unrelated), release dates, sequence-dependent setup times, among others.  

Weighted \mbox{E-T} problems are usually $\mathcal{NP}$-hard, since they include the single machine total weighted tardiness scheduling problem ($1||\sum w_jT_j$) as a particular case, which is known to be $\mathcal{NP}$-hard \citep{Lawler1977, Lenstra1977}. 
Therefore, solving this type of problems to optimality is an extremely hard task. Nevertheless, there has been a continuous effort towards the development of efficient exact algorithms for this type of problems \citep{SourdKedad-Sidhoum2003, Sourd2005, Sourd2008, TanakaFujikuma2008, Sourd2009, Tanakaetal2009, Pessoaetal2010, TanakaFujikuma2012, Tanaka2012}. In some particular situations, often involving a single machine environment without sequence-dependent setup times, they seem to be very effective, even solving instances with up to 300 jobs \citep{Tanakaetal2009, Tanaka2012}, but in most cases their application are still limited to small/medium-size instances. Heuristic algorithms are thus the natural alternative for trying to generate high quality feasible solutions within an acceptable computational time.

The large majority of heuristic algorithms proposed for scheduling problems were devised for a particular variant. While there are unified methods for polynomial problems \citep{Leyvandetal2010} as well as polynomial time approximation schemes \citep{Kumaretal2009, Epsteinetal2013} and exact approaches \citep{ChenPowell1999, TanakaFujikuma2008, TanakaFujikuma2012, Lin2013} for classes of $\mathcal{NP}$-hard problems, we are not aware of general heuristics developed for similar purposes, as opposed, for example, to the field of vehicle routing, where there has been a trend in the development of efficient unified heuristics \citep{Cordeauetal2001, RopkePisinger2006, PisingerRopke2007, CordeauMaischberger2012, Subramanianetal2013, Vidaletal2013, Vidaletal2014, DerigsVogel2014}. 

Vehicle routing and single/parallel-machine scheduling problems have some similarities in the sense that in both problems one should determine the order in which the tasks (deliver goods to customers, processing a job in a machine, etc.) must be performed. However, there is a clear distinction between these two problems, especially when it comes to the objective function. In classical vehicle routing problems the objective is frequently to minimize the sum of the travel costs, whereas in scheduling problems the variety of objectives is more notable, ranging from minimizing the makespan or the total weighted completion time to minimizing the total weighted \mbox{(earliness-)tardiness} and so on. These differences on the objective functions have a direct influence, for instance, on the local search performance of a heuristic. In standard vehicle routing problems, a move can be often computed in constant time, while in scheduling problems more complex and specific procedures are generally required for evaluating the cost of a move. Moreover, while some aspects such as idle (waiting) times do not usually (directly) affect the objective function value in, for example, feasible solutions of vehicle routing problems with time windows, they may cause a considerable impact on the solution cost of E-T scheduling problems. These are just few examples of many of the issues that arise in scheduling problems that are generally relatively easier to handle in vehicle routing algorithms. 


In addition, a heuristic that works well for scheduling problems without idle time may end up having a poor performance  when applied to a variant in which idle time is allowed, mainly because in the latter one should solve a subproblem known as \emph{timing problem} \citep{Vidaletal2015b} that consists of determining the optimal start times of the jobs given a sequence. In fact, parts of a heuristic algorithm that were not designed to allow idle time should be substantially redesigned to address this point. Note that the opposite can also happen, that is, a heuristic designed to handle idle time may turn out to have a slow performance when considering instances without idle time.

In summary, it seems more challenging to devise a unified heuristic for standard scheduling problems than for vehicle routing problems, which may explain the lack of this kind of algorithms in the scheduling literature. Furthermore, we strongly believe that unified methods are extremely important in practice. For example, commercial packages must be designed to be both robust and general enough to efficiently deal with many real-life problems. Also, these type of methods can be a good source of reference if one intends to evaluate the performance of specially-tailored algorithms for some particular variants.

This work attempts to start filling the aforementioned gap by proposing a unified heuristic algorithm for single/parallel-machine scheduling problems. By ``unified" we mean that several specific ingredients are put together into a single framework that is capable of efficiently solving a large class of distinct scheduling problems. However, due to the countless number of variants existing in the literature, we decided to focus our attention on a subset of them, in particular, E-T problems without preemption. We consider a class of problems that may or may not consider earliness penalties, sequence-dependent setup times, release dates, due dates and idle time, resulting in a broad range of variants that include other type of objectives such as minimizing the total weighted tardiness; total (average) weighted completion time; and total (average) weighted flow time. 

The developed heuristic algorithm extends the one of \citet{Subramanian2012, Subramanianetal2013} that was successfully applied to solve a class of vehicle routing problems. The conceptual idea of the proposed method is quite simple, however, in contrast to the original algorithm, some specific parts rely on sophisticated procedures \citep{Ibarakietal2005, Ibarakietal2008, ErgunOrlin2006, Liaoetal2012} for efficiently performing local search according to the characteristics of the problem. One particular contribution of this work is the development of efficient move evaluation approaches for a class of parallel machine problems. Such approaches generalize the ideas presented in \cite{ErgunOrlin2006} and \cite{Liaoetal2012} for single machine total weighted tardiness problems. The algorithm was tested in benchmark instances of several E-T problems. The results obtained show that our unified heuristic is capable of producing high quality solutions when compared to the best ones available in the literature that were obtained by specific methods.

We also provide an extensive annotated bibliography on the problems related to those considered in this work, where we not only indicate the approach(es) used in each publication, but we also point out the characteristics of the problem(s) considered. In addition, we classify the heuristic and exact methods in different categories so as to have a better idea of the popularity of each type of solution procedure.

The remainder of this work is organized as follows. Section \ref{sec:RelatedWork} presents an annotated bibliography on E-T works related to the class of problems considered in this research. Section \ref{sec:Problems} specifies the range of problems solve by our unified heuristic whose detailed description can be found in Section \ref{sec:Heuristic}. Computational experiments are reported in Section \ref{sec:Results}. Finally, Section \ref{sec:Conclusions} contains the concluding remarks.

\section{Problems considered}
\label{sec:Problems}

In this section we enumerate the class of E-T problems, including some particular cases, that our unified heuristic is capable of solving. We start by characterizing the general problem followed by a list of particular cases. 

Let $J = \{1,\dots,n\}$ be a set of jobs to be scheduled on a set of unrelated parallel machines given by $M = \{1,\dots,m\}$. For each job $j \in J$, let $p_j^k$, $d_j$, $r_j$, $w'_j$ and $w_j$  be its processing time in machine $k \in M$, due date, release date, earliness penalty weight and tardiness penalty weight, respectively. Also, let $s_{ij}^{k}$ be the setup time required before starting to process job $j \in J$ if $j$ is scheduled immediately after job $i \in J$ in machine $k \in M$. 
The objective is to minimize $\sum w'_jE_j + w_jT_j$, where $E_j = \max\{d_j - C_j, 0\}$ and $T_j = \max\{C_j - d_j, 0\}$ are the earliness and tardiness of a job $j \in J$, respectively, that depends on its completion time $C_j$. Idle time is allowed to be inserted between two consecutive jobs. According to the notation suggested by \citet{Graham1979}, this problem can be referred to as $R|r_j, s_{ij}^k|\sum w'_jE_j + w_jT_j$.

A large number of problems arise as a special case of the problem described above, including well-known single machine problems without sequence-dependent setup times such as $1||\sum w_jT_j$, $1|r_j|\sum w_jT_j$, $1||\sum w'_jE_j + w_jT_j$,  $1|r_j|\sum w'_jE_j + w_jT_j$,  $1|r_j|\sum  w_jC_j$ that can be efficiently solved by the exact algorithm of \citet{TanakaFujikuma2012} for instances with up to 200 jobs (sometimes even 300 jobs as in the case of $1||\sum w_jT_j$). Note that the latter problem, which consists of minimizing the average weighted completion time, is a particular case of problem $1|r_j|\sum w_jT_j$ when all due dates are admitted to be zero, i.e., $T_j = \max\{C_j-0,0\} = C_j, \forall j \in J$. Although the proposed heuristic is capable of dealing with these problems, its performance is simply not as good as some exact algorithms such as the one of \citet{TanakaFujikuma2012}. In fact, it is really hard to devise heuristic algorithms with a superior or at least equivalent performance than the exact ones for these problems, even for large size instances. Our heuristic still finds high quality solutions for most of the existing instances, but not as fast as the state-of-the-art exact methods. For brevity, we decided not to report computational results for these problems.   

Another type of objective function that arise as particular case of $\sum w_jT_j$ is when one aims at minimizing the average weighted flow time, which is given by $\sum w_j F_j = \sum w_j (C_j - r_j)$. Note that by setting $d_j = r_j, \forall j \in J$, we obtain $T_j = \max\{C_j-r_j,0\} = F_j, \forall j \in J$.

Table \ref{tab:Problems} lists some of the main problems that appear as special case of problem $R|r_j, s_{ij}^k|\sum w'_jE_j + w_jT_j$ and where our heuristic can be applied, including those mentioned above for the sake of completeness. Problems like $1||w_jC_j$ do not appear in the table because we only considered $\mathcal{NP}$-hard problems. We also do not explicitly include problems with unitary weights because they are simply particular cases of the weighted problems. For example, problem $1|s_{ij}|\sum T_j$ is well studied in the literature, but in principle any algorithm developed for problem $1|s_{ij}|\sum w_jT_j$ can be used to solve the former one. In fact, some recent works on problem $1|s_{ij}|\sum w_jT_j$ \citep{Kirliketal2012, Tanaka2013, Subramanian2014, Xuetal2014} also considered the version with unitary weights.

\begin{table}[!ht]
  \centering
  \onehalfspacing
  \small
  \caption{Problems considered}
  \begin{tabular}{llll}
    \hline
    \multicolumn{1}{c}{Single machine} & \multicolumn{1}{c}{Identical machines} & \multicolumn{1}{c}{Uniform machines} & \multicolumn{1}{c}{Unrelated machines} \\
    \hline
    $1||\sum w_jT_j$ & $P||\sum  w_jT_j$ & $Q||\sum w_jT_j$ & $R||\sum w_jT_j$ \\
    $1|r_j|\sum w_jT_j$ & $P|r_j|\sum w_jT_j$ & $Q|r_j|\sum w_jT_j$ & $R|r_j|\sum w_jT_j$ \\
    $1|s_{ij}|\sum w_jT_j$ & $P|s_{ij}|\sum w_jT_j$ & $Q|s_{ij}^{k}|\sum w_jT_j$ & $R|s_{ij}^{k}|\sum w_jT_j$ \\
    $1|r_j,  s_{ij}|\sum w_jT_j$ & $P|r_j,  s_{ij}|\sum w_jT_j$ & $Q|r_j,  s_{ij}^{k}|\sum w_jT_j$ & $R|r_j,s_{ij}^{k}|\sum w_jT_j$ \\
    $1||\sum w'_jE_j+w_jT_j$ & $P||\sum w'_jE_j+w_jT_j$ & $Q||\sum w'_jE_j+w_jT_j$ & $R||\sum w'_jE_j+w_jT_j$ \\
    $1|r_j|\sum w'_jE_j+w_jT_j$ & $P|r_j|\sum w'_jE_j+w_jT_j$ & $Q|r_j|\sum w'_jE_j+w_jT_j$ & $R|r_j|\sum w'_jE_j+w_jT_j$ \\
    $1|s_{ij}|\sum w'_jE_j+w_jT_j$ & $P|s_{ij}|\sum w'_jE_j+w_jT_j$ & $Q|s_{ij}^{k}|\sum w'_jE_j+w_jT_j$ & $R|s_{ij}^{k}|\sum w'_jE_j+w_jT_j$ \\
    $1|r_j,  s_{ij}|\sum w'_jE_j+w_jT_j$ & $P|r_j,  s_{ij}|\sum w'_jE_j+w_jT_j$ & $Q|r_j,  s_{ij}^{k}|\sum w'_jE_j+w_jT_j$ & $R|r_j,  s_{ij}^{k}|\sum w'_jE_j+w_jT_j$ \\
    $1|r_j|\sum w_jC_j$ & $P||\sum w_jC_j$ & $Q||\sum w_jC_j$ & $R||\sum w_jC_j$ \\
    $1|s_{ij}|\sum w_jC_j$ & $P|r_j|\sum w_jC_j$ & $Q|r_j|\sum w_jC_j$ & $R|r_j|\sum w_jC_j$ \\
    $1|r_j,  s_{ij}|\sum w_jC_j$ & $P|s_{ij}|\sum w_jC_j$ & $Q|s_{ij}^{k}|\sum w_jC_j$ & $R|s_{ij}^{k}|\sum w_jC_j$ \\
    $1|r_j|\sum w_jF_j$ & $P|r_j,  s_{ij}|\sum w_jC_j$ & $Q|r_j,  s_{ij}^{k}|\sum w_jC_j$ & $R|r_j,  s_{ij}^{k}|\sum w_jC_j$ \\
    $1|r_j, s_{ij}|\sum w_jF_j$ & $P|r_j|\sum w_jF_j$ & $Q|r_j|\sum w_jF_j$ & $R|r_j|\sum w_jF_j$ \\
		& $P|r_j,  s_{ij}|\sum w_jF_j$ & $Q|r_j,  s_{ij}^{k}|\sum w_jF_j$ & $R|r_j,  s_{ij}^{k}|\sum w_jF_j$ \\    
    \hline  
  \end{tabular}      
  \label{tab:Problems}
\end{table}

We did not perform computational experiments for all problems, not only for the reasons mentioned above, but also due to lack of publicly available instances. In addition, for some particular problems in which there are available instances, the authors who proposed them did not report lower/upper bounds.

\section{An annotated bibliography for E-T problems related to those considered in this work}
\label{sec:RelatedWork}

There is a vast literature on \mbox{(E-)T} scheduling. The first publications started to appear in the 1980s and they were surveyed by \citet{BakerScudder1990}. Another survey was later performed by \citet{LauffWerner2004} for E-T problems in a multi-machine environment. Recently, \citet{Ratlietal2013} presented an overview on mathematical formulations and heuristics proposed for E-T problems but on a single machine environment. In these three works, the authors only reviewed problems with common due dates for all jobs.  An entire book devoted to E-T scheduling was written by \citet{Jozefowska2007} where the author compiled the main models and algorithms for this class of problems, including those with common and individual due dates, respectively. The latter case is not only more general but also considered to be more challenging, since the first one has some properties that facilitates scheduling decisions. For example, it is known that idle time does not appear in the optimal solution of some variants where common due dates are considered, which makes the problem easier to be solved. 

Moreover, problems with other characteristics such as learning effects, preemption, maintenance, deteriorating jobs, etc., are not considered in this section. We also do not consider approximation algorithms for problems that aim at minimizing either the total weighted completion time (in this case  we refer the reader to the survey of \cite{ChekuriKhanna2004}) or the total flow time (see \cite{Kellereretal1996, LeonardiRaz2007}). In addition, we only list the works published in the last 25 years.
Finally, despite all our efforts in performing a complete enumeration of the large related literature, we cannot ensure that the annotated bibliography presented in this section include all the available relevant work.

\subsection{Single machine environment}
\label{sec:Single}

A considerable amount of the research involving \mbox{(E-)T} problems in the literature consider a single machine environment. Since there is a very large number of published works, it becomes rather impractical (and also it is beyond our goal) to provide a detailed description of each of them. Instead, we summarize, in chronological order, most of these works in Table \ref{tab:SingleMachineWorks}, where we specify the solution approach; the scheduling characteristics such as the existence of sequence-dependent setup-times ($s_{ij}$), release dates ($r_j$) and idle time (IT); and the type of objective considered by each method, namely (non-)weighted tardiness ($T_j$), (non-)weighted earliness and tardiness ($E_j + T_j$), (non-)weighted flow time ($F_j$) and (non-)weighted completion time ($C_j$).

\begin{center}
{\footnotesize
\renewcommand{\arraystretch}{1.10}
\setlength{\arrayrulewidth}{2\arrayrulewidth}
\begin{longtable}{>{\centering\arraybackslash}m{4.5cm} >{\centering\arraybackslash}m{4.5cm} c c c c c c c}
\caption{Single machine environment --- list of works}\\
\hline
Reference & Approach(es) & $s_{ij}$ & $r_j$ & $T_j$ & $E_j + T_j$ & $C_j$ & $F_j$ & IT \\
\hline
\endfirsthead
\caption[]{(\textit{Continued})}\\
\hline
Reference & Approach(es) & $s_{ij}$ & $r_j$ & $T_j$ & $E_j + T_j$ & $C_j$ & $F_j$ & IT \\
\hline
\endhead
\hline \multicolumn{9}{r}{\textit{Continued on next page}} \\
\endfoot
\endlastfoot
    \citet{Raman1989}	& Constructive Heuristic  &  \checkmark &  & \checkmark &  & & &\\
    \citet{Dyer1990}  & Formulations and valid inequalities &   & \checkmark &  &  & \checkmark & & \checkmark \\
		\citet{PottsVanWassenhove1991}  & SA, Constructive and randomized interchanging heuristics & & & \checkmark &  &  & & \checkmark \\
    \citet{Yano1991}  & Heuristics, DP &   &  &  & \checkmark & & & \checkmark\\
    \citet{Belouadah1992}  & B\&B &   & \checkmark &  &  & \checkmark & & \checkmark \\
    \citet{Chu1992}  & B\&B &   & \checkmark &  &  &  & \checkmark &  \\
    \citet{Chu1992b}  & B\&B &   & \checkmark & \checkmark  &  &  &  &  \\
    \citet{Rubin1995}	& GA  &  \checkmark &  & \checkmark &  & & &\\
    \citet{DellaCroce1995}	& Local search, Lower bounds  &  & \checkmark & \checkmark & \checkmark & \checkmark & &\\
    \citet{Tan1997}	& SA  &  \checkmark &  & \checkmark &  &&\\
    \citet{Li1997}  & B\&B + LR, Heuristic &   &  &  & \checkmark & & & \\
    \citet{Lee1997}	& Constructive Heuristic  &  \checkmark &  & \checkmark &  & & &\\
    \citet{Akturk2000} & B\&B  &   & \checkmark  & \checkmark &  & & &\\
    \citet{Franca2001}	& GA, MA  &  \checkmark &  & \checkmark &  & & &\\
    \citet{Wan2002} & TS  &   &   &  & \checkmark & & & \checkmark  \\
    \citet{Gagne2002}	& ACO  &  \checkmark &  & \checkmark &  & & &\\
    \citet{Congrametal2002}	& Dynasearch, Iterated dynasearch, ILS  &  &  & \checkmark &  & & &\\
    \citet{Feldmann2003}	& EA, SA, Threshold Accepting &    &  &  & \checkmark & & &\\
    \citet{SourdKedad-Sidhoum2003} & B\&B + LR  &   &   &  & \checkmark & & & \checkmark\\
    \citet{Guoetal2004}	& Experimental analysis with an approximation algorithm &   & \checkmark &  &  &  & \checkmark &  \\
    \citet{Grossoetal2004}	& Dynasearch, Eliminations rules &   &  & \checkmark &  &  & &  \\    
    \citet{Sourd2005}	& Time-indexed formulation, B\&B + LR, Multi-start heuristic &   \checkmark &  &  & \checkmark & & & \checkmark\\
    \citet{Gagne2005}	& TS + VNS  &  \checkmark &  & \checkmark &  & & &\\
    \citet{Cicirello2005}	& Heuristics  &  \checkmark &  & \checkmark &  & & &\\
    \citet{Masonetal2005} & Moving block heuristic &    &  &  & \checkmark & & & \checkmark\\
    \citet{ErgunOrlin2006}  & Fast neighborhood search &  &  & \checkmark &  & & & \\
    \citet{Esteve2006} & Beam Search  &   & \checkmark  &  & \checkmark & & & \checkmark\\
    \citet{Cicirello2006}	& GA  &  \checkmark &  & \checkmark &  & & &\\
    \citet{Gupta2006}	& GRASP, PR, Problem Spaced-Based Local Search  &  \checkmark &  & \checkmark &  & & &\\
    \citet{Hendel2006}	& Efficient Local Search		&    &  &  & \checkmark & & & \checkmark\\
    \citet{Sourd2006}	& Dynasearch	&   \checkmark & \checkmark &  & \checkmark & & & \checkmark\\
    \citet{Bulbul2007}  & Preemption based relaxation + Heuristic &  & \checkmark &  & \checkmark & & & \checkmark \\
    \citet{MHallah2007}	& GA + Hill-Climbing + SA &    &  &  & \checkmark & & &\\
    \citet{LiaoJuan2007}	& ACO  &  \checkmark &  & \checkmark &  & & &\\
    \citet{Liao2007}	& VNS + TS &    &  &  & \checkmark & & &\\
    \citet{Lin2007}	& GA, SA, TS  &  \checkmark &  & \checkmark &  & & &\\
    \citet{Tsai2007}	& GA  &  & \checkmark & & \checkmark & & & \checkmark\\
		\citet{PanandShi2008}	& Hybrid approach: B\&B + DP + CP  &  & \checkmark & & & \checkmark & & \checkmark\\
		\citet{TanakaFujikuma2008} & SSDP  &   & \checkmark  &  & \checkmark & & & \checkmark\\
    \citet{Anghinolfi2008}	& ACO &   \checkmark &  & \checkmark &  & & &\\
    \citet{Sourd2008}	& B\&B + LR &   & \checkmark &  & \checkmark & & &\checkmark\\
    \citet{Valente2008}	& Beam Search  &  \checkmark &  & \checkmark &  & & &\\
    \citet{Bigras2008}	& Formulations, B\&B  &  \checkmark &  & \checkmark &  & & &\\
    \citet{Anghinolfi2009}	& PSO &   \checkmark &  & \checkmark & & & &\\
    \citet{Tasgetiren2009}	& EA &   \checkmark &  & \checkmark & & & &\\
    \citet{Arroyo2009}	& ILS + GRASP  &  \checkmark &  & \checkmark &  & & &\\
    \citet{Ying2009}	& Iterated Greedy algorithm  &  \checkmark &  & \checkmark &  & & & \\
    \citet{Sourd2009} & B\&B + LR  &   &   &  & \checkmark & & & \checkmark\\
    \citet{Tanakaetal2009} & SSDP  &  &   & \checkmark & \checkmark & & & \\
    \citet{Geiger2010}  & Empirical Analysis &  \checkmark &  & \checkmark &  & & &\\
    \citet{Bozejko2010}	& Parallel SS  &  \checkmark &  & \checkmark &  & & &\\
    \citet{KedadSidhoum2010}& ILS, Fast Neighborhood Search&    &  &  & \checkmark & & & \checkmark\\
    \citet{RONCONI2010} & B\&B  &   &   &  & \checkmark & & &\\
    \citet{Yoon2011}	& Constructive heuristics &    &  & \checkmark &  & & &\\
    \citet{Mandahawi2011}	& ACO  &  \checkmark &  & \checkmark &  & & &\\
    \citet{Kirliketal2012}	& VNS &   \checkmark &  & \checkmark &  & & &\\
    \citet{Liaoetal2012}  & Fast neighborhood Search & \checkmark  &  & \checkmark &  & & & \\
    \citet{Sioud2012}	& Hybrid GA & \checkmark &  & \checkmark &  & & &\\
    \citet{TanakaFujikuma2012} & SSDP  &   & \checkmark  & \checkmark & \checkmark & \checkmark & & \checkmark\\
    \citet{Tanaka2012} & SSDP  &   & \checkmark  &  & \checkmark & & & \checkmark\\
    \citet{Sioud2012}	& GA + ACO  &  \checkmark &  & \checkmark &  & & &\\
    \citet{Wan2013} & Proof of strongly $\mathcal{NP}$-Hardness for problem $1||E_j + T_J$  &   &   &  & \checkmark & & & \checkmark\\
    \citet{Tanaka2013}	& SSDP  &   \checkmark &  & \checkmark &  & & &\\
    \citet{Xu2013}		& ILS &   \checkmark &  & \checkmark &  & & &\\
    \citet{Subramanian2014}	& ILS  &  \checkmark &  & \checkmark &  & & &\\
    \citet{DengGu2014}  & ILS &  \checkmark &  & \checkmark &  & & &\\
    \citet{Guo2015}  & SS &  \checkmark &  & \checkmark &  & & &\\
    \citet{SubramanianFarias2015}	& Efficient local search limitation strategy  &  \checkmark &  & \checkmark &  & & &\\
\hline\noalign{\smallskip}
\label{tab:SingleMachineWorks}
\end{longtable}
}
\end{center}

From Table \ref{tab:SingleMachineWorks}, we can observe that the wide range of heuristic procedures for single machine problems proposed in the literature can be classified as follows:

\begin{itemize}
  \item Constructive or improvement heuristics \citep{Raman1989, PottsVanWassenhove1991, Yano1991, DellaCroce1995, Li1997, Lee1997, Cicirello2005, Masonetal2005, Sourd2005, Ying2009, Yoon2011}.
  \item Efficient local search procedures \citep{Congrametal2002, Grossoetal2004, ErgunOrlin2006, Hendel2006, Sourd2006, KedadSidhoum2010, Liaoetal2012, SubramanianFarias2015}.
  \item Population based metaheuristics:
  \begin{itemize}
    \item Genetic Algorithms (GAs) \citep{Rubin1995, Cicirello2006, Lin2007, Tsai2007, Sioud2012}.  
    \item Evolutionary Algorithms (EAs) \citep{Feldmann2003, Tasgetiren2009}. 
    \item Memetic Algorithms (MAs) \citep{Franca2001}. 
    \item Ant Colony Optimization (ACO) \citep{Gagne2002, LiaoJuan2007, Anghinolfi2008, Mandahawi2011}. 
    \item Particle Swarm Optimization (PSO) \citep{Anghinolfi2009}.
    \item Scatter Search (SS) \citep{Bozejko2010, Guo2015}.
  \end{itemize}
  \item Local search based metaheuristics: 
  \begin{itemize}
    \item Simulated Annealing (SA) \citep{PottsVanWassenhove1991, Tan1997, Feldmann2003, Lin2007}. 
    \item Tabu Search (TS) \citep{Wan2002, Lin2007}. 
    \item Variable Neighborhood Search (VNS) \citep{Kirliketal2012}. 
    \item Iterated Local Search (ILS) \citep{Congrametal2002, Ying2009, KedadSidhoum2010, Xu2013, Subramanian2014}. 
    \item Greedy Randomized Adaptive Search Procedure (GRASP) \citep{Gupta2006}.
  \end{itemize}
  \item Hybrid metaheuristics \citep{Gagne2005, MHallah2007, Liao2007, Arroyo2009, Sioud2012}.
  \item Mathematical Programming based heuristics \citep{Bulbul2007}.
  \item Alternative methods such as Beam Search (BS) \citep{Esteve2006, Valente2008}, Spaced-Based Local Search (SBLS) and Path-Relinking (PR) \citep{Gupta2006}.
\end{itemize}

As for the exact methods, they mostly consist of a partial combination between Dynamic Programming (DP), Lagrangian Relaxation (LR), Branch-and-Bound (B\&B) and Mixed Integer Programming (MIP) formulations, and they can be categorized as follows:

\begin{itemize}
  \item B\&B based on domination rules \citep{Belouadah1992, Chu1992, Chu1992b, Akturk2000, RONCONI2010}.
  \item DP + LR, such as the Successive Sublimation Dynamic Programming (SSDP) algorithm of Tanaka et al. \citep{TanakaFujikuma2008, Tanakaetal2009, Tanaka2012, TanakaFujikuma2012, Tanaka2013}.
  \item B\&B + LR, such as the methods of \citet{SourdKedad-Sidhoum2003, Sourd2005, Sourd2008, Sourd2009} and \citet{Li1997}.
	\item B\&B + DP + Constraint Programming (CP) \citep{PanandShi2008}.
  \item B\&B over linear programming (LP) relaxations based on Column Generation (CG) and/or cutting planes \citep{Bigras2008}.
\end{itemize}


\subsection{Parallel machine environment}
\label{sec:Parallel}

In this section we list the related works that considered parallel machines. Tables \ref{tab:IdenticalParallelMachineWorks}, \ref{tab:UniformParallelMachineWorks} and \ref{tab:UnrelatedParallelMachineWorks} summarize these works according to the environment, namely identical, uniform and unrelated parallel machine. 

\begin{center}
{\footnotesize
\renewcommand{\arraystretch}{1.10}
\setlength{\arrayrulewidth}{2\arrayrulewidth}
\begin{longtable}{>{\centering\arraybackslash}m{4.5cm} >{\centering\arraybackslash}m{4.5cm} c c c c c c c}
\caption{Identical parallel machine environment --- list of works}\\
\hline
Reference & Approach(es) & $s_{ij}$ & $r_j$ & $T_j$ & $E_j + T_j$ & $C_j$ & $F_j$ & IT \\
\hline
\endfirsthead
\caption[]{(\textit{Continued})}\\
\hline
Reference & Approach(es) & $s_{ij}$ & $r_j$ & $T_j$ & $E_j + T_j$ & $C_j$ & $F_j$ & IT \\
\hline
\endhead
\hline \multicolumn{9}{r}{\textit{Continued on next page}} \\
\endfoot
\endlastfoot
    \citet{Webster1992} & Lower bounds & & \checkmark  & & & & \checkmark &\\
    \citet{Webster1993} & Optimal priority rules for a special case of problem $P|r_j|\sum w_j F_j$ & & \checkmark  & & & & \checkmark &\\
    \citet{Webster1995} & Lower bounds & & \checkmark  & & & & \checkmark &\\
    \citet{BelouadahPotts1994} & B\&B + LR &  &  &  & & \checkmark & &\\
    \citet{LeePinedo1997}	& Preprocessing phase + Apparent tardiness cost with setups heuristic (ATCS) + SA &   \checkmark &  & \checkmark &  & & &\\
    \citet{Koulamas1997}	& Lower bounds, SA &  &  & \checkmark & & & &\\
    \citet{AzizogluKirca1998}	& B\&B &  &  & \checkmark &  & & &\\
    \citet{AzizogluKirca1999}	& B\&B &  &  &  &  & \checkmark & &\\
    \citet{ChenPowell1999}	& CG &  &  &  &  & \checkmark & &\\
    \citet{Radhakrishnanetal2000}	& SA & \checkmark &  &  & \checkmark & & & \\
    \citet{Eometal2002}	& EDD + ATCS + TS & \checkmark &  & \checkmark & & & & \\
    \citet{YalaouiChu2002}	& B\&B &  &  & \checkmark & & & \\
    \citet{SunWang2003}	& DP, Constructive heuristic &  &  &  & \checkmark & & &\\
    \citet{Kimetal2006}	& TS & \checkmark & \checkmark & \checkmark & & & & \checkmark\\
    \citet{OmarTeo2006}	& MIP Formulation & \checkmark &  &  & \checkmark & & & \checkmark \\
    \citet{YalaouiChu2006}	& B\&B &  & \checkmark & & & \checkmark & & \checkmark\\
    \citet{ShimKim2007a}	& B\&B &  &  & \checkmark &  & & & \\
    \citet{KedadSidhoum2008} & Time-indexed formulation, LR, CG, Efficient local search &  & \checkmark &  & \checkmark & & &\checkmark\\
    \citet{Feng2008}	& Squeaky Wheel Optimization (SWO) & \checkmark &  &  & \checkmark & & &\checkmark\\
    \citet{RiosSolis2008}	& Exponential Neighborhood Search &  &  &  & \checkmark & & &\\
    \citet{Pfund2008}	& Apparent tardiness cost with setups and ready times (ATCSR) heuristic & \checkmark & \checkmark & \checkmark &  & & &\checkmark\\
    \citet{Nessahetal2008}	& B\&B &  & \checkmark & & & \checkmark & & \checkmark\\
    \citet{Biskupetal2008}	& Constructive Heuristic &  &  & \checkmark & & & & \\
    \citet{Rodrigues2008}	& ILS &  &  & \checkmark & & & & \\
    \citet{TanakaAraki2008}	& B\&B + LR&  &  & \checkmark & & & & \\
    \citet{Baptisteetal2008}	& Lower Bounds &  & \checkmark & \checkmark & & \checkmark & & \checkmark\\
    \citet{Mason2009}	& Moving block heuristic  &  &  &  & \checkmark & & & \checkmark\\
    \citet{Pessoaetal2010}	& Time-indexed formulation, Branch-cut-and-price &  &  & \checkmark & & & &\\
    \citet{Jouglet2011}	& B\&B &  & \checkmark & \checkmark & & & & \checkmark\\
    \citet{MHallah2012}	& MIP, Hybrid heuristic (steepest descent + GA + SA) &  &  &  & \checkmark & & & \checkmark\\
    \citet{DellaCroce2012}	& ILS + Very Large Neighborhood Search  &  &  & \checkmark &  & &\\
    \citet{Amorim2013}	& Hybrid GA &  &  &  & \checkmark & &\\
    \citet{Amorim2013b}	& GA, ILS, PR &  &  &  & \checkmark & &\\
    \citet{Schaller2014}	& TSs, GAs, B\&B  & \checkmark &  & \checkmark &  & &\\
    \citet{Xietal2015}	& Look-ahead constructive heuristic & \checkmark & \checkmark & \checkmark & & & & \checkmark\\
\hline\noalign{\smallskip}
\label{tab:IdenticalParallelMachineWorks}
\end{longtable}
}
\end{center}

\begin{center}
{\footnotesize
\renewcommand{\arraystretch}{1.10}
\setlength{\arrayrulewidth}{2\arrayrulewidth}
\begin{longtable}{>{\centering\arraybackslash}m{4.5cm} >{\centering\arraybackslash}m{4.5cm} c c c c c c c}
\caption{Uniform parallel machine environment --- list of works}\\
\hline
Reference & Approach(es) & $s_{ij}^k$ & $r_j$ & $T_j$ & $E_j + T_j$ & $C_j$ & $F_j$ & IT \\
\hline
\endfirsthead
\caption[]{(\textit{Continued})}\\
\hline
Reference & Approach(es) & $s_{ij}^k$ & $r_j$ & $T_j$ & $E_j + T_j$ & $C_j$ & $F_j$ & IT \\
\hline
\endhead
\hline \multicolumn{9}{r}{\textit{Continued on next page}} \\
\endfoot
\endlastfoot
    \citet{Webster1992} & Lower bounds & & \checkmark  & & & & \checkmark &\\
    \citet{Guinet1995}	& SA &  &  & \checkmark &  &  & & \\
    \citet{AzizogluKirca1998}	& B\&B &  &  & \checkmark &  & & &\\
    \citet{AzizogluKirca1999}	& B\&B &  &  &  &  & \checkmark & &\\
    \citet{ChenPowell1999}	& CG &  &  &  &  & \checkmark & & \\
    \citet{Sivrikaya1999}	& GA &   \checkmark & \checkmark &  & \checkmark & & & \checkmark\\
    \citet{Balakrishnanetal1999}	& Mathematical formulation + Benders' decomposition& \checkmark & \checkmark &  & \checkmark & & & \\
    \citet{BiskupFeldmann2001}	& Benchmark instances	&  &  &  & & \checkmark & & \\
    \citet{Bilgeetal2004}	& TS & \checkmark & \checkmark & \checkmark & & & & \\
    \citet{Anghinolfi2007}	& TS+SA+VNS & \checkmark & \checkmark & \checkmark & & & & \\
    \citet{Armentano2007}	& GRASP		& \checkmark & \checkmark & \checkmark & & & &\\
    \citet{Raja2008}	& SA + Fuzzy Logic	& \checkmark &  &  & \checkmark &  & & \\
    \citet{Yousefi2013}	& Imperialist Competitive Algorithm &  &  &  & \checkmark &  & & \checkmark\\
    \citet{Lin2013}  & Models / Formulations &  &  &  &  & \checkmark & & \\
    \citet{Lietal2014}	& Agent-based algorithm + Lower Bounds	&  & \checkmark &  & & \checkmark & & \checkmark \\
\hline\noalign{\smallskip}
\label{tab:UniformParallelMachineWorks}
\end{longtable}
}
\end{center}
\vspace{-0.75cm}
\begin{center}
{\footnotesize
\renewcommand{\arraystretch}{1.10}
\setlength{\arrayrulewidth}{2\arrayrulewidth}
\begin{longtable}{>{\centering\arraybackslash}m{4.5cm} >{\centering\arraybackslash}m{4.5cm} c c c c c c c }
\caption{Unrelated parallel machine environment --- list of works}\\
\hline
Reference & Approach(es) & $s_{ij}^k$ & $r_j$ & $T_j$ & $E_j + T_j$ & $C_j$ & $F_j$ & IT \\
\hline
\endfirsthead
\caption[]{(\textit{Continued})}\\
\hline
Reference & Approach(es) & $s_{ij}^k$ & $r_j$ & $T_j$ & $E_j + T_j$ & $C_j$ & $F_j$ & IT \\
\hline
\endhead
\hline \multicolumn{9}{r}{\textit{Continued on next page}} \\
\endfoot
\endlastfoot
    \citet{Webster1992} & Lower bounds & & \checkmark  & & & & \checkmark &\\
    \citet{ZhuHeady2000} & MIP formulation & \checkmark &  &  & \checkmark & & &\\
    \citet{BankWerner2001} & Constructive + iterative heuristics, SA &  & \checkmark &  & \checkmark & & & \checkmark\\
    \citet{Wengetal2001} & Constructive heuristics & \checkmark &  &  & & \checkmark & & \\
    \citet{Liawetal2003} & B\&B &  &  & \checkmark & & & &\\
    \citet{Zhouetal2007} & ACO &  &  & \checkmark & & & &\\
    \citet{ShimKim2007b}	& B\&B &  &  & \checkmark &  & & & \\
    \citet{Logendranetal2007} & Six algorithms based on TS & \checkmark & \checkmark & \checkmark &  & & & \\
    \citet{AkyolBayhan2008} & Neural network & \checkmark &  &  & \checkmark & & \\
    \citet{LiYang2009} & Survey & \checkmark & \checkmark &  & \checkmark & & \\
    \citet{ValladaRuiz2012} & MIP formulation, GA & \checkmark &  &  & \checkmark & & & \checkmark\\
    \citet{Leeetal2013} & TS & \checkmark &  & \checkmark &  & & \\
    \citet{Polyakovskiy2014} & Multi-Agent System Heuristic &  &  &  & \checkmark & & & \checkmark\\ 
    \citet{Nogueira2014}        & GRASP + PR + ILS & \checkmark &  &  & \checkmark & & & \checkmark\\
    \citet{LinHsieh2014} & Modified ATCSR and Electromagnetism-like Algorithm (EMA)  & \checkmark & \checkmark & \checkmark & & & &\\
		\citet{BulbulandSen2016} & Preemptive relaxation + Benders' decomposition & & & & & \checkmark & &\\
		\citet{SenBulbul2015} & Preemptive relaxation + Benders' decomposition + solver SiPS/SiPSi \citep{Tanakaetal2009} for obtaining upper bounds & & \checkmark & \checkmark & \checkmark & & & \checkmark \\
\hline\noalign{\smallskip}
\label{tab:UnrelatedParallelMachineWorks}
\end{longtable}
}
\end{center}

The heuristic approaches proposed for parallel machine problems can be classified as follows:

\begin{itemize}
  \item Constructive and/or improvement heuristics \citep{LeePinedo1997, BankWerner2001, Wengetal2001, SunWang2003, Feng2008, Biskupetal2008, KedadSidhoum2008, Pfund2008, Mason2009, Xietal2015}.
  \item Population based metaheuristics: 
  \begin{itemize}
    \item GA \citep{Sivrikaya1999, ValladaRuiz2012, Amorim2013, Amorim2013b, Schaller2014}.
    \item EA \citep{Yousefi2013, LinHsieh2014}.
    \item ACO \citep{Zhouetal2007}.
  \end{itemize}
  \item Local search based metaheuristics: 
  \begin{itemize}
    \item SA \citep{Guinet1995, LeePinedo1997, Koulamas1997, Radhakrishnanetal2000, BankWerner2001}.
    \item TS \citep{Eometal2002, Bilgeetal2004, Kimetal2006, Logendranetal2007, Leeetal2013, Schaller2014}.
    \item ILS \citep{Rodrigues2008, Amorim2013b}.
    \item GRASP \citep{Armentano2007}.
    \item Large Neighborhood Search (LNS) \citep{RiosSolis2008}.
  \end{itemize}
  \item Hybrid metaheuristics \citep{Anghinolfi2007, DellaCroce2012, MHallah2012, Nogueira2014}.
  \item Artificial Intelligence based heuristics \citep{AkyolBayhan2008, Raja2008, Polyakovskiy2014, Lietal2014}.
  \item Mathematical Programming based heuristics \citep{KedadSidhoum2008, SenBulbul2015}.
\end{itemize}

The exact methods developed for parallel machine problems can be classified as follows:

\begin{itemize}
  \item B\&B based on domination rules \citep{AzizogluKirca1998, AzizogluKirca1999, YalaouiChu2002, Liawetal2003, YalaouiChu2006,  ShimKim2007b, ShimKim2007a, Jouglet2011, Nessahetal2008, Schaller2014}.
  \item B\&B + LR \citep{BelouadahPotts1994, TanakaAraki2008}.
  \item B\&B over linear programming (LP) relaxations based on CG and/or cutting planes \citep{ChenPowell1999, Pessoaetal2010}.
  \item Compact MIP formulations \citep{Balakrishnanetal1999, ZhuHeady2000, OmarTeo2006, ValladaRuiz2012, Lin2013}.
  \item Preemption based relaxation \citep{BulbulandSen2016}.
\end{itemize}

\section{The unified heuristic algorithm}
\label{sec:Heuristic}

The proposed unified heuristic, called UILS, is mostly based on ILS \citep{Lourencoetal2002}. This metaheuristic basically alternates between intensification (local search) and diversification (perturbation) procedures in order to escape from local optima. We modified the original ILS algorithm by allowing multiple restarts of the method. Previous works showed that this type of implementation, along with a Randomized Variable Neighborhood Descent (RVND) approach in the local search phase, yielded high quality results for different kinds of problems \citep{Subramanian2010, Silvaetal2012, Pennaetal2013, SubramanianBattarra2013, Martinellietal2013, Subramanian2014, Vidaletal2015, SubramanianFarias2015}, most of them in the field of routing, including the unified algorithms presented in \cite{Subramanian2012, Subramanianetal2013}.

Given the previous successful implementations of the multi-start ILS-RVND algorithm, we decided to extend this method to E-T scheduling problems. However, as pointed out in Section \ref{sec:Introduction}, several issues arise when dealing with scheduling problems that usually do not appear in vehicle routing problems. The main adaptation was in the local search phase where we implemented a tailored move evaluation approach according to the characteristics of the problem, such as the existence or not of earliness penalties, idle time, release dates and sequence-dependent setup times. This is one of the key aspects for the versatility and potential scalability of the proposed heuristic when facing problems with distinct features.

\begin{figure}[!ht]
 \begin{center}
  \includegraphics[scale=0.35]{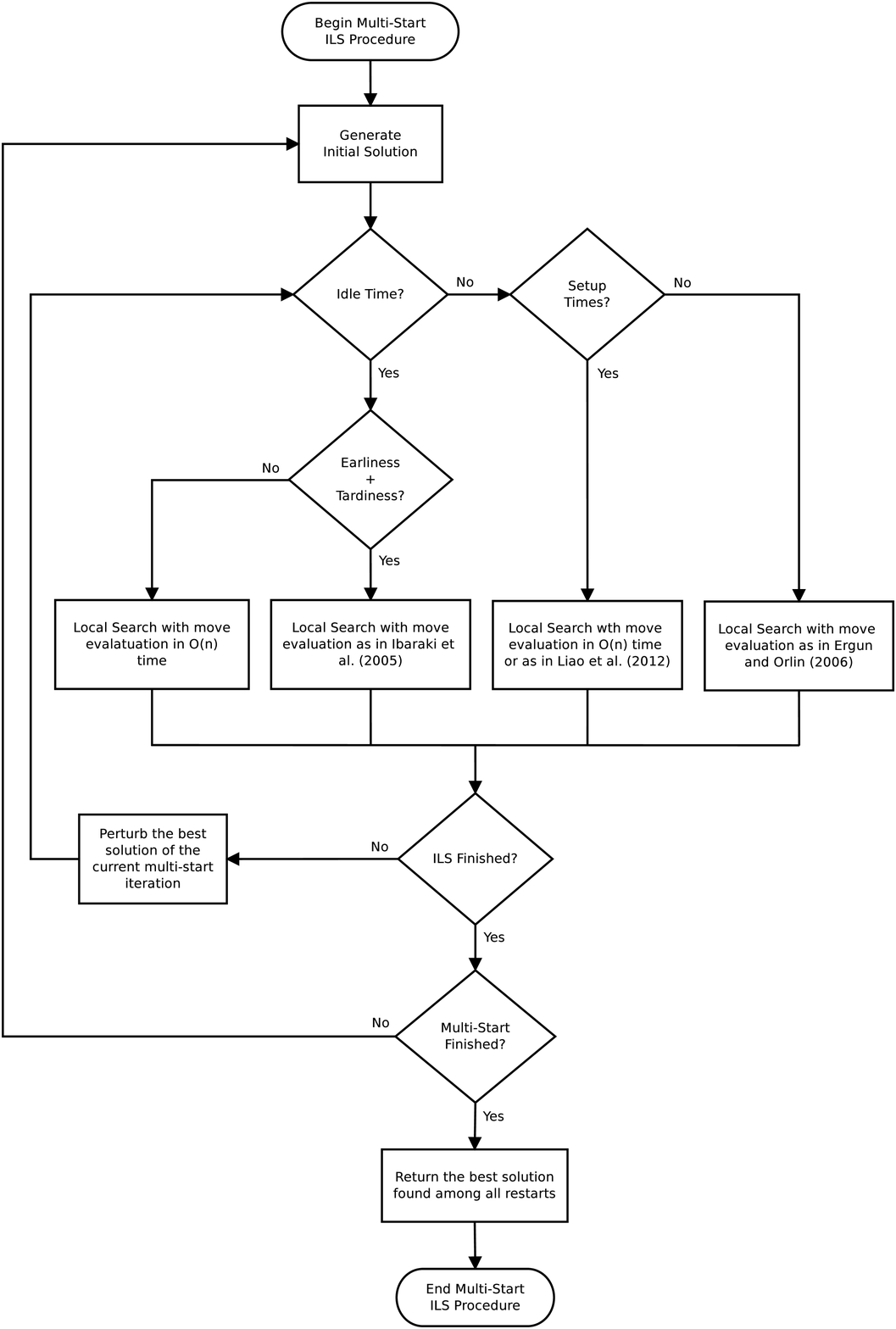}\\
  \caption{Flowchart of the unified heuristic}
\label{fig:ILS-Flowchart}
 \end{center}  
\end{figure}

The multi-start heuristic starts by generating an initial solution using a very simple greedy randomized or a completely randomized insertion procedure. Next, a local search is performed using RVND, but with a specific move evaluation scheme that depends on the characteristics of the problem. On the one hand, when idles times are not considered, the choice of the move evaluation scheme to be used depends on the existence of setup times. On the other hand,  when idle times are taken into account, the decision on the move evaluation scheme to be used depends on the presence of earliness penalties. Note that in this latter case, the existence of setup times does not affect such decision.
Moreover, if the maximum number of consecutive perturbations without improvements ($I_{ILS}$) is not achieved (ILS stopping criterion), then the algorithm modifies the incumbent solution of the current multi-start iteration by applying a perturbation mechanism and then it restarts the local search procedure from that perturbed solution.
Otherwise, the algorithm restarts from the beginning. If the maximum number of restarts ($I_R$) is achieved then the heuristic stops and returns the best solution found.

\subsection{Initial Solutions}
\label{sec:InitialSolutions}

For problems involving multiple machines, the initial solutions are generated at random, whereas for single machine problems without release date the solutions are built either at random or following the earliest due date (EDD) method. When release dates are considered, the initial solutions are generated using the earliest release date (ERD) criterion. For the sake of diversification, EDD and ERD were implemented based on the greedy randomized approach of the constructive phase of GRASP \citep{FeoResende1995}. 

It is worth mentioning that, based on preliminary experiments, the use of more sophisticated methods for building initial solutions do not significantly affect, on average, neither the quality of the final solution, nor the speed of convergence towards a local optimum. Hence, for simplicity, we decided to keep the constructive procedure as simple as possible.

\subsection{Local Search}
\label{sec:LocalSearch}

The local search is performed by a RVND procedure \citep{Mladenovic1997, Subramanian2012}, which consists of randomly choosing an unexplored neighborhood to proceed with the search every time another one fails to find an improved solution. Otherwise, all neighborhoods are  allowed to be selected. 

In what follows, we describe the neighborhood structures used in RVND, as well as the  move evaluation procedures which can vary depending on the characteristics of the problem.



\subsubsection{Neighborhood structures}
\label{sec:Neighborhoods}

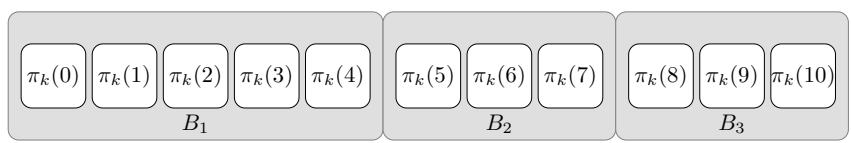
\begin{figure}[!ht]
\centering
  \begin{tikzpicture}[thick, scale=0.85, every node/.style={transform shape}, >=stealth', dot/.style = {draw, fill = white, circle, inner sep = 0pt, minimum size = 4pt}]
    \draw[rounded corners, very thin, color=black!50, fill=gray!25] (0,1) rectangle (5.8,3);\node (B1) at (2.9,1.25){$B_1$};
    \draw[rounded corners, very thin, fill=white] (0.2,1.5) rectangle (1.2,2.5);\node (A) at (0.7,2){$\pi_k(0)$};
		\draw[rounded corners, very thin, fill=white] (1.3,1.5) rectangle (2.3,2.5);\node (B) at (1.8,2){$\pi_k(1)$};
		\draw[rounded corners, very thin, fill=white] (2.4,1.5) rectangle (3.4,2.5);\node (C) at (2.9,2){$\pi_k(2)$};
		\draw[rounded corners, very thin, fill=white] (3.5,1.5) rectangle (4.5,2.5);\node (D) at (4,2){$\pi_k(3)$};
		\draw[rounded corners, very thin, fill=white] (4.6,1.5) rectangle (5.6,2.5);\node (E) at (5.1,2){$\pi_k(4)$};
		
    \draw[rounded corners, very thin, color=black!50, fill=gray!25] (5.8,1) rectangle (9.4,3);\node (B2) at (7.6,1.25){$B_2$};
    \draw[rounded corners, very thin, fill=white] (6,1.5) rectangle (7,2.5);\node (F) at (6.5,2){$\pi_k(5)$};
		\draw[rounded corners, very thin, fill=white] (7.1,1.5) rectangle (8.1,2.5);\node (G) at (7.6,2){$\pi_k(6)$};
		\draw[rounded corners, very thin, fill=white] (8.2,1.5) rectangle (9.2,2.5);\node (H) at (8.7,2){$\pi_k(7)$};

		\draw[rounded corners, very thin, color=black!50, fill=gray!25] (9.4,1) rectangle (13,3);\node (B3) at (11.2,1.25){$B_3$};
    \draw[rounded corners, very thin, fill=white] (9.6,1.5) rectangle (10.6,2.5);\node (I) at (10.1,2){$\pi_k(8)$};
		\draw[rounded corners, very thin, fill=white] (10.7,1.5) rectangle (11.7,2.5);\node (J) at (11.2,2){$\pi_k(9)$};
		\draw[rounded corners, very thin, fill=white] (11.8,1.5) rectangle (12.8,2.5);\node (K) at (12.3,2){$\pi_k(10)$};

  \end{tikzpicture}
	\caption{Example of block representation}
	\label{fig:block}
\end{figure}

With a view of better illustrating the neighborhood structures used in our algorithm we resort to a block representation, which is commonly adopted in the scheduling literature. Let $\pi_k = (\pi_k(0),\pi_k(1),\dots,\pi_k(n_k))$ be a sequence of $n_k+1$ jobs associated with a machine $k \in M$, with $\pi_k(0) = 0$ as a dummy job that is necessary for considering an eventual setup $s^k_{0\pi_k(1)}$ for processing the initial job of the sequence. A block $B$ consists of subsequence of consecutive jobs. An example involving 10 jobs divided into three blocks ($B_1 = (\pi_k(0),\pi_k(1),\pi_k(2),\pi_k(3),\pi_k(4))$, $B_2 = (\pi_k(5),\pi_k(6),\pi_k(7))$, $B_3 = (\pi_k(8),\pi_k(9),\pi_k(10))$) is shown in Fig. \ref{fig:block}.

The neighborhoods used in RVND are based on insertion and swap moves involving subsequences of jobs (blocks) and they are described next.

\begin{itemize}
	\item $l$-block insertion intra-machine --- consists of moving (reinserting) a block of size $l$, starting from job $\pi_k(i)$, to the position after job $\pi_k(j)$ in the same machine. A block ($B_2$) is moved forward (after $B_3$) when $i = 1,\dots, n_k-l$ and $j=i+l,\dots, n_k$, if $i < j$; and moved backward ($B_3$ before $B_2$) when $i = 2,\dots,n_k+1-l$ and $j=0,\dots,n_k-l$, if $i > j$, as depicted in Figs. \ref{fig:l-blockInsertionFwd} and \ref{fig:l-blockInsertionBack}, respectively.

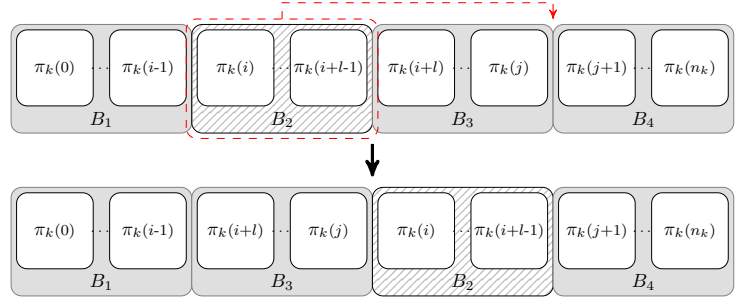
\begin{figure}[!ht]
	\centering
  \begin{tikzpicture}[thick, scale=0.72, every node/.style={transform shape}, >=stealth', dot/.style = {draw, fill = white, circle, inner sep = 0pt, minimum size = 4pt}]
		\draw[rounded corners, very thin, color=black!50, fill=gray!25] (0,3.5) rectangle (3.3,5.5);\node (B1) at (1.65,3.75){$B_1$};
    \draw[rounded corners, very thin, fill=white] (0.1,4) rectangle (1.5,5.4);\node (A) at (0.8,4.7){$\pi_k$\footnotesize{($0$)}};
    \node (D1) at (1.65, 4.7){\scriptsize{$\dots$}};
		\draw[rounded corners, very thin, fill=white] (1.8,4) rectangle (3.2,5.4);\node (B) at (2.5,4.7){$\pi_k$\footnotesize{($i$-$1$)}};

    \draw[pattern=north east lines, pattern color=gray!55, rounded corners, very thin] (3.3,3.5) rectangle (6.6,5.5);\node (B2) at (4.95,3.75){$B_2$};
    \draw[rounded corners, very thin, fill=white] (3.4,4) rectangle (4.8,5.4);\node (C) at (4.1,4.7){$\pi_k$\footnotesize{($i$)}};
    \node (D2) at (4.95, 4.7){\scriptsize{$\dots$}};
		\draw[rounded corners, very thin, fill=white] (5.1,4) rectangle (6.5,5.4);\node (D) at (5.8,4.7){$\pi_k$\footnotesize{($i$+$l$-$1$)}};
		
    \draw[rounded corners, very thin, color=black!50, fill=gray!25] (6.6,3.5) rectangle (9.9,5.5);\node (B3) at (8.25,3.75){$B_3$};
    \draw[rounded corners, very thin, fill=white] (6.7,4) rectangle (8.1,5.4);\node (E) at (7.4,4.7){$\pi_k$\footnotesize{($i$+$l$)}};
    \node (D3) at (8.25, 4.7){\scriptsize{$\dots$}};
		\draw[rounded corners, very thin, fill=white] (8.4,4) rectangle (9.8,5.4);\node (F) at (9.1,4.7){$\pi_k$\footnotesize{($j$)}};
		
    \draw[rounded corners, very thin, color=black!50, fill=gray!25] (9.9,3.5) rectangle (13.2,5.5);\node (B4) at (11.55,3.75){$B_4$};
    \draw[rounded corners, very thin, fill=white] (10,4) rectangle (11.4,5.4);\node (G) at (10.7,4.7){$\pi_k$\footnotesize{($j$+$1$)}};
    \node (D4) at (11.55, 4.7){\scriptsize{$\dots$}};
		\draw[rounded corners, very thin, fill=white] (11.7,4) rectangle (13.1,5.4);\node (H) at (12.4,4.7){$\pi_k$\footnotesize{($n_k$)}};

		\draw[dashed, rounded corners, thin, color=red] (3.2,3.4) rectangle (6.7,5.6);
		\draw[dashed, thin, color=red] (4.95,5.6) -- (4.95,5.9) -- (9.9,5.9) [->] -- (9.9, 5.6);
		\draw[->, very thick, color=black] (6.6,3.3) -- (6.6,2.7);
    \draw[rounded corners, very thin, color=black!50, fill=gray!25] (0,0.5) rectangle (3.3,2.5);\node (B1) at (1.65,0.75){$B_1$};
    \draw[rounded corners, very thin, fill=white] (0.1,1) rectangle (1.5,2.4);\node (A) at (0.8,1.7){$\pi_k$\footnotesize{($0$)}};
    \node (D1) at (1.65, 1.7){\scriptsize{$\dots$}};
		\draw[rounded corners, very thin, fill=white] (1.8,1) rectangle (3.2,2.4);\node (B) at (2.5,1.7){$\pi_k$\footnotesize{($i$-$1$)}};

    \draw[rounded corners, very thin, color=black!50, fill=gray!25] (3.3,0.5) rectangle (6.6,2.5);\node (B2) at (4.95,0.75){$B_3$};
    \draw[rounded corners, very thin, fill=white] (3.4,1) rectangle (4.8,2.4);\node (C) at (4.1,1.7){$\pi_k$\footnotesize{($i$+$l$)}};
    \node (D2) at (4.95, 1.7){\scriptsize{$\dots$}};
		\draw[rounded corners, very thin, fill=white] (5.1,1) rectangle (6.5,2.4);\node (D) at (5.8,1.7){$\pi_k$\footnotesize{($j$)}};
		
    \draw[pattern=north east lines, pattern color=gray!55, rounded corners, very thin] (6.6,0.5) rectangle (9.9,2.5);\node (B3) at (8.25,0.75){$B_2$};
    \draw[rounded corners, very thin, fill=white] (6.7,1) rectangle (8.1,2.4);\node (E) at (7.4,1.7){$\pi_k$\footnotesize{($i$)}};
    \node (D3) at (8.25, 1.7){\scriptsize{$\dots$}};
		\draw[rounded corners, very thin, fill=white] (8.4,1) rectangle (9.8,2.4);\node (F) at (9.1,1.7){$\pi_k$\footnotesize{($i$+$l$-$1$)}};
		
    \draw[rounded corners, very thin, color=black!50, fill=gray!25] (9.9,0.5) rectangle (13.2,2.5);\node (B4) at (11.55,0.75){$B_4$};
    \draw[rounded corners, very thin, fill=white] (10,1) rectangle (11.4,2.4);\node (G) at (10.7,1.7){$\pi_k$\footnotesize{($j$+$1$)}};
    \node (D4) at (11.55, 1.7){\scriptsize{$\dots$}};
		\draw[rounded corners, very thin, fill=white] (11.7,1) rectangle (13.1,2.4);\node (H) at (12.4,1.7){$\pi_k$\footnotesize{($n_k$)}};
  \end{tikzpicture}
	\caption{$l$-block insertion forward intra-machine}
	\label{fig:l-blockInsertionFwd}
\end{figure}
\begin{figure}[!ht]
	\centering
  \begin{tikzpicture}[thick, scale=0.72, every node/.style={transform shape}, >=stealth', dot/.style = {draw, fill = white, circle, inner sep = 0pt, minimum size = 4pt}]

  \draw[rounded corners, very thin, color=black!50, fill=gray!25] (0,3.5) rectangle (3.3,5.5);\node (B1) at (1.65,3.75){$B_1$};
    \draw[rounded corners, very thin, fill=white] (0.1,4) rectangle (1.5,5.4);\node (A) at (0.8,4.7){$\pi_k$\footnotesize{($0$)}};
    \node (D1) at (1.65, 4.7){\scriptsize{$\dots$}};
		\draw[rounded corners, very thin, fill=white] (1.8,4) rectangle (3.2,5.4);\node (B) at (2.5,4.7){$\pi_k$\footnotesize{($j$-$1$)}};

    \draw[rounded corners, very thin, color=black!50, fill=gray!25] (3.3,3.5) rectangle (6.6,5.5);\node (B2) at (4.95,3.75){$B_2$};
    \draw[rounded corners, very thin, fill=white] (3.4,4) rectangle (4.8,5.4);\node (C) at (4.1,4.7){$\pi_k$\footnotesize{($j$)}};
    \node (D2) at (4.95, 4.7){\scriptsize{$\dots$}};
		\draw[rounded corners, very thin, fill=white] (5.1,4) rectangle (6.5,5.4);\node (D) at (5.8,4.7){$\pi_k$\footnotesize{($i$-$1$)}};
		
    \draw[pattern=north east lines, pattern color=gray!55, rounded corners, very thin] (6.6,3.5) rectangle (9.9,5.5);\node (B3) at (8.25,3.75){$B_3$};
    \draw[rounded corners, very thin, fill=white] (6.7,4) rectangle (8.1,5.4);\node (E) at (7.4,4.7){$\pi_k$\footnotesize{($i$)}};
    \node (D3) at (8.25, 4.7){\scriptsize{$\dots$}};
		\draw[rounded corners, very thin, fill=white] (8.4,4) rectangle (9.8,5.4);\node (F) at (9.1,4.7){$\pi_k$\footnotesize{($i$+$l$-$1$)}};
		
    \draw[rounded corners, very thin, color=black!50, fill=gray!25] (9.9,3.5) rectangle (13.2,5.5);\node (B4) at (11.55,3.75){$B_4$};
    \draw[rounded corners, very thin, fill=white] (10,4) rectangle (11.4,5.4);\node (G) at (10.7,4.7){$\pi_k$\footnotesize{($i$+$l$)}};
    \node (D4) at (11.55, 4.7){\scriptsize{$\dots$}};
		\draw[rounded corners, very thin, fill=white] (11.7,4) rectangle (13.1,5.4);\node (H) at (12.4,4.7){$\pi_k$\footnotesize{($n_k$)}};

		\draw[dashed, rounded corners, thin, color=red] (6.5,3.4) rectangle (10,5.6);
		\draw[dashed, thin, color=red] (8.25,5.6) -- (8.25,5.9) -- (3.3,5.9) [->] -- (3.3, 5.6);
		\draw[->, very thick, color=black] (6.6,3.3) -- (6.6,2.7);
    \draw[rounded corners, very thin, color=black!50, fill=gray!25] (0,0.5) rectangle (3.3,2.5);\node (B1) at (1.65,0.75){$B_1$};
    \draw[rounded corners, very thin, fill=white] (0.1,1) rectangle (1.5,2.4);\node (A) at (0.8,1.7){$\pi_k$\footnotesize{($0$)}};
    \node (D1) at (1.65, 1.7){\scriptsize{$\dots$}};
		\draw[rounded corners, very thin, fill=white] (1.8,1) rectangle (3.2,2.4);\node (B) at (2.5,1.7){$\pi_k$\footnotesize{($j$-$1$)}};

    \draw[pattern=north east lines, pattern color=gray!55, rounded corners, very thin] (3.3,0.5) rectangle (6.6,2.5);\node (B2) at (4.95,0.75){$B_3$};
    \draw[rounded corners, very thin, fill=white] (3.4,1) rectangle (4.8,2.4);\node (C) at (4.1,1.7){$\pi_k$\footnotesize{($i$)}};
    \node (D2) at (4.95, 1.7){\scriptsize{$\dots$}};
		\draw[rounded corners, very thin, fill=white] (5.1,1) rectangle (6.5,2.4);\node (D) at (5.8,1.7){$\pi_k$\footnotesize{($i$+$l$-$1$)}};
		
    \draw[rounded corners, very thin, color=black!50, fill=gray!25] (6.6,0.5) rectangle (9.9,2.5);\node (B3) at (8.25,0.75){$B_2$};
    \draw[rounded corners, very thin, fill=white] (6.7,1) rectangle (8.1,2.4);\node (E) at (7.4,1.7){$\pi_k$\footnotesize{($j$)}};
    \node (D3) at (8.25, 1.7){\scriptsize{$\dots$}};
		\draw[rounded corners, very thin, fill=white] (8.4,1) rectangle (9.8,2.4);\node (F) at (9.1,1.7){$\pi_k$\footnotesize{($i$-$1$)}};
		
    \draw[rounded corners, very thin, color=black!50, fill=gray!25] (9.9,0.5) rectangle (13.2,2.5);\node (B4) at (11.55,0.75){$B_4$};
    \draw[rounded corners, very thin, fill=white] (10,1) rectangle (11.4,2.4);\node (G) at (10.7,1.7){$\pi_k$\footnotesize{($i$+$l$)}};
    \node (D4) at (11.55, 1.7){\scriptsize{$\dots$}};
		\draw[rounded corners, very thin, fill=white] (11.7,1) rectangle (13.1,2.4);\node (H) at (12.4,1.7){$\pi_k$\footnotesize{($n_k$)}};
  \end{tikzpicture}
	\caption{$l$-block insertion backward intra-machine}
	\label{fig:l-blockInsertionBack}
\end{figure}
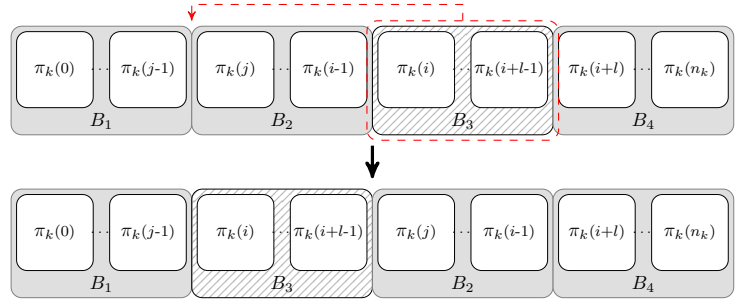

  \item $(l,l')$-block swap intra-machine --- consists of interchanging a block ($B_2$) of size $l$, starting from job $\pi_k(i)$, with another one belonging to the same machine ($B_4$) of size $l'$, starting from job $\pi_k(j)$, as shown in Fig. \ref{fig:swapIntra}.

\begin{figure}[!ht]
	\centering
  \begin{tikzpicture}[thick, scale=0.72, every node/.style={transform shape}, >=stealth', dot/.style = {draw, fill = white, circle, inner sep = 0pt, minimum size = 4pt}]
		\draw[rounded corners, very thin, color=black!50, fill=gray!25] (0,3.5) rectangle (3.3,5.5);\node (B1) at (1.65,3.75){$B_1$};
    \draw[rounded corners, very thin, fill=white] (0.1,4) rectangle (1.5,5.4);\node (A) at (0.8,4.7){$\pi_k$\footnotesize{($0$)}};
    \node (D1) at (1.65, 4.7){\scriptsize{$\dots$}};
		\draw[rounded corners, very thin, fill=white] (1.8,4) rectangle (3.2,5.4);\node (B) at (2.5,4.7){$\pi_k$\footnotesize{($i$-$1$)}};

    \draw[pattern=north east lines, pattern color=gray!55, rounded corners, very thin] (3.3,3.5) rectangle (6.6,5.5);\node (B2) at (4.95,3.75){$B_2$};
    \draw[rounded corners, very thin, fill=white] (3.4,4) rectangle (4.8,5.4);\node (C) at (4.1,4.7){$\pi_k$\footnotesize{($i$)}};
    \node (D2) at (4.95, 4.7){\scriptsize{$\dots$}};
		\draw[rounded corners, very thin, fill=white] (5.1,4) rectangle (6.5,5.4);\node (D) at (5.8,4.7){$\pi_k$\footnotesize{($i$+$l$-$1$)}};
		
    \draw[rounded corners, very thin, color=black!50, fill=gray!25] (6.6,3.5) rectangle (9.9,5.5);\node (B3) at (8.25,3.75){$B_3$};
    \draw[rounded corners, very thin, fill=white] (6.7,4) rectangle (8.1,5.4);\node (E) at (7.4,4.7){$\pi_k$\footnotesize{($i$+$l$)}};
    \node (D3) at (8.25, 4.7){\scriptsize{$\dots$}};
		\draw[rounded corners, very thin, fill=white] (8.4,4) rectangle (9.8,5.4);\node (F) at (9.1,4.7){$\pi_k$\footnotesize{($j$-$1$)}};
		
    \draw[pattern=north east lines, pattern color=gray!55, rounded corners, very thin] (9.9,3.5) rectangle (13.2,5.5);\node (B4) at (11.55,3.75){$B_4$};
    \draw[rounded corners, very thin, fill=white] (10,4) rectangle (11.4,5.4);\node (G) at (10.7,4.7){$\pi_k$\footnotesize{($j$)}};
    \node (D4) at (11.55, 4.7){\scriptsize{$\dots$}};
		\draw[rounded corners, very thin, fill=white] (11.7,4) rectangle (13.1,5.4);\node (H) at (12.4,4.7){$\pi_k$\footnotesize{($j$+$l^\prime$-$1$)}};
		
    \draw[rounded corners, very thin, color=black!50, fill=gray!25] (13.2,3.5) rectangle (16.5,5.5);\node (B5) at (14.85,3.75){$B_5$};
    \draw[rounded corners, very thin, fill=white] (13.3,4) rectangle (14.7,5.4);\node (I) at (14,4.7){$\pi_k$\footnotesize{($j$+$l^\prime$)}};
    \node (D5) at (14.85, 4.7){\scriptsize{$\dots$}};
		\draw[rounded corners, very thin, fill=white] (15,4) rectangle (16.4,5.4);\node (H) at (15.7,4.7){$\pi_k$\footnotesize{($n_k$)}};

		\draw[dashed, rounded corners, thin, color=red] (9.8,3.4) rectangle (13.3,5.6);
		\draw[dashed, rounded corners, thin, color=red] (3.2,3.4) rectangle (6.7,5.6);
		\draw[dashed, thin, color=red] (5.45,5.6) -- (5.45,5.9) -- (12.05,5.9) [->] -- (12.05, 5.6);
		\draw[dashed, thin, color=red] (11.05,3.4) -- (11.05,3.1) -- (4.45,3.1) [->] -- (4.45, 3.4);
		\draw[->, very thick, color=black] (8.25,3.3) -- (8.25,2.7);
    \draw[rounded corners, very thin, color=black!50, fill=gray!25] (0,0.5) rectangle (3.3,2.5);\node (B1) at (1.65,0.75){$B_1$};
    \draw[rounded corners, very thin, fill=white] (0.1,1) rectangle (1.5,2.4);\node (A) at (0.8,1.7){$\pi_k$\footnotesize{($0$)}};
    \node (D1) at (1.65, 1.7){\scriptsize{$\dots$}};
		\draw[rounded corners, very thin, fill=white] (1.8,1) rectangle (3.2,2.4);\node (B) at (2.5,1.7){$\pi_k$\footnotesize{($i$-$1$)}};

    \draw[pattern=north east lines, pattern color=gray!55, rounded corners, very thin] (3.3,0.5) rectangle (6.6,2.5);\node (B2) at (4.95,0.75){$B_4$};
    \draw[rounded corners, very thin, fill=white] (3.4,1) rectangle (4.8,2.4);\node (C) at (4.1,1.7){$\pi_k$\footnotesize{($j$)}};
    \node (D2) at (4.95, 1.7){\scriptsize{$\dots$}};
		\draw[rounded corners, very thin, fill=white] (5.1,1) rectangle (6.5,2.4);\node (D) at (5.8,1.7){$\pi_k$\footnotesize{($j$+$l^\prime$-$1$)}};
		
    \draw[rounded corners, very thin, color=black!50, fill=gray!25] (6.6,0.5) rectangle (9.9,2.5);\node (B3) at (8.25,0.75){$B_3$};
    \draw[rounded corners, very thin, fill=white] (6.7,1) rectangle (8.1,2.4);\node (E) at (7.4,1.7){$\pi_k$\footnotesize{($i$+$l$)}};
    \node (D3) at (8.25, 1.7){\scriptsize{$\dots$}};
		\draw[rounded corners, very thin, fill=white] (8.4,1) rectangle (9.8,2.4);\node (F) at (9.1,1.7){$\pi_k$\footnotesize{($j$-$1$)}};
		
    \draw[pattern=north east lines, pattern color=gray!55, rounded corners, very thin] (9.9,0.5) rectangle (13.2,2.5);\node (B4) at (11.55,0.75){$B_2$};
    \draw[rounded corners, very thin, fill=white] (10,1) rectangle (11.4,2.4);\node (G) at (10.7,1.7){$\pi_k$\footnotesize{($i$)}};
    \node (D4) at (11.55, 1.7){\scriptsize{$\dots$}};
		\draw[rounded corners, very thin, fill=white] (11.7,1) rectangle (13.1,2.4);\node (H) at (12.4,1.7){$\pi_k$\footnotesize{($i$+$l$-$1$)}};
		
    \draw[rounded corners, very thin, color=black!50, fill=gray!25] (13.2,0.5) rectangle (16.5,2.5);\node (B5) at (14.85,0.75){$B_5$};
    \draw[rounded corners, very thin, fill=white] (13.3,1) rectangle (14.7,2.4);\node (I) at (14,1.7){$\pi_k$\footnotesize{($j$+$l^\prime$)}};
    \node (D5) at (14.85, 1.7){\scriptsize{$\dots$}};
		\draw[rounded corners, very thin, fill=white] (15,1) rectangle (16.4,2.4);\node (H) at (15.7,1.7){$\pi_k$\footnotesize{($n_k$)}};
	\end{tikzpicture}
	\caption{$(l,l')$-block Swap intra-machine}
	\label{fig:swapIntra}
\end{figure}
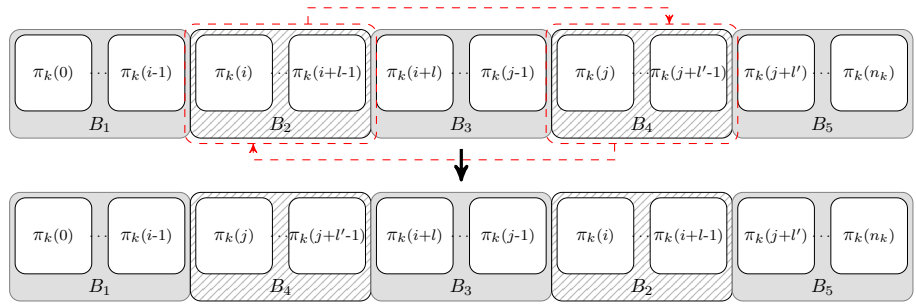
  \item $l$-block insertion inter-machine --- consists of removing a block ($B_2$) of size $l$ from a machine $k$, starting from $\pi_k(i)$, and inserting it in machine $k^{\prime}$ at position $j$, after block $B_4$, as illustrated in Fig. \ref{fig:l-blockInsertionInter}.

\begin{figure}[!ht]
	\centering
  \begin{tikzpicture}[thick, scale=0.75, every node/.style={transform shape}, >=stealth', dot/.style = {draw, fill = white, circle, inner sep = 0pt, minimum size = 4pt}]
		\draw[rounded corners, very thin, color=black!50, fill=gray!5] (-0.2,6) rectangle (10.1, 0.3);
		\node (M1) at (0.8, 5.8){Machine $k$};
  
		\draw[rounded corners, very thin, color=black!50, fill=gray!25] (0,3.5) rectangle (3.3,5.5);\node (B1) at (1.65,3.75){$B_1$};
    \draw[rounded corners, very thin, fill=white] (0.1,4) rectangle (1.5,5.4);\node (A) at (0.8,4.7){$\pi_k$\footnotesize{($0$)}};
    \node (D1) at (1.65, 4.7){\scriptsize{$\dots$}};
		\draw[rounded corners, very thin, fill=white] (1.8,4) rectangle (3.2,5.4);\node (B) at (2.5,4.7){$\pi_k$\footnotesize{($i$-$1$)}};

    \draw[pattern=north east lines, pattern color=gray!55, rounded corners, very thin] (3.3,3.5) rectangle (6.6,5.5);\node (B2) at (4.95,3.75){$B_2$};
    \draw[rounded corners, very thin, fill=white] (3.4,4) rectangle (4.8,5.4);\node (C) at (4.1,4.7){$\pi_k$\footnotesize{($i$)}};
    \node (D2) at (4.95, 4.7){\scriptsize{$\dots$}};
		\draw[rounded corners, very thin, fill=white] (5.1,4) rectangle (6.5,5.4);\node (D) at (5.8,4.7){$\pi_k$\footnotesize{($i$+$l$-$1$)}};
		
    \draw[rounded corners, very thin, color=black!50, fill=gray!25] (6.6,3.5) rectangle (9.9,5.5);\node (B3) at (8.25,3.75){$B_3$};
    \draw[rounded corners, very thin, fill=white] (6.7,4) rectangle (8.1,5.4);\node (E) at (7.4,4.7){$\pi_k$\footnotesize{($i$+$l$)}};
    \node (D3) at (8.25, 4.7){\scriptsize{$\dots$}};
		\draw[rounded corners, very thin, fill=white] (8.4,4) rectangle (9.8,5.4);\node (F) at (9.1,4.7){$\pi_k$\footnotesize{($n_k$)}};

		\draw[dashed, rounded corners, thin, color=red] (3.2,3.4) rectangle (6.7,5.6);
		\draw[dashed, thin, color=red] (4.95,3.4) [->]-- (4.95,2.6);
    \node (M2) at (2.45, 2.8){Machine $k^\prime$};
    
    \draw[rounded corners, very thin, color=black!50, fill=gray!25] (1.65,0.5) rectangle (4.95,2.5);\node (B4) at (3.3,0.75){$B_4$};
    \draw[rounded corners, very thin, fill=white] (1.75,1) rectangle (3.15,2.4);\node (A) at (2.45,1.7){$\pi_{k^\prime}$\footnotesize{($0$)}};
    \node (D1) at (3.3, 1.7){\scriptsize{$\dots$}};
		\draw[rounded corners, very thin, fill=white] (3.45,1) rectangle (4.85,2.4);\node (B) at (4.15,1.7){$\pi_{k^\prime}$\footnotesize{($j$-$1$)}};

    \draw[rounded corners, very thin, color=black!50, fill=gray!25] (4.95,0.5) rectangle (8.25,2.5);\node (B5) at (6.6,0.75){$B_5$};
    \draw[rounded corners, very thin, fill=white] (5.05,1) rectangle (6.45,2.4);\node (C) at (5.75,1.7){$\pi_{k^\prime}$\footnotesize{($j$)}};
    \node (D2) at (6.6, 1.7){\scriptsize{$\dots$}};
		\draw[rounded corners, very thin, fill=white] (6.75,1) rectangle (8.15,2.4);\node (D) at (7.45,1.7){$\pi_{k^\prime}$\footnotesize{($n_{k^\prime}$)}};
		\draw[->, very thick, color=black] (4.95,0.2) -- (4.95,-0.5);
    \draw[rounded corners, very thin, color=black!50, fill=gray!5] (-0.2,-0.55) rectangle (10.1,-6.25);
    \node (M1) at (2.45, -0.75){Machine $k$};
    
    \draw[rounded corners, very thin, color=black!50, fill=gray!25] (1.65,-3.05) rectangle (4.95,-1.05);\node (B4) at (3.3,-2.8){$B_1$};
    \draw[rounded corners, very thin, fill=white] (1.75,-2.55) rectangle (3.15,-1.15);\node (A) at (2.45,-1.85){$\pi_k$\footnotesize{($0$)}};
    \node (D1) at (3.3, -1.85){\scriptsize{$\dots$}};
		\draw[rounded corners, very thin, fill=white] (3.45,-2.55) rectangle (4.85,-1.15);\node (B) at (4.15,-1.85){$\pi_k$\footnotesize{($i$-$1$)}};

    \draw[rounded corners, very thin, color=black!50, fill=gray!25] (4.95,-3.05) rectangle (8.25,-1.05);\node (B5) at (6.6,-2.8){$B_3$};
    \draw[rounded corners, very thin, fill=white] (5.05,-2.55) rectangle (6.45,-1.15);\node (C) at (5.75,-1.85){$\pi_k$\footnotesize{($i$+$l$)}};
    \node (D2) at (6.6, -1.85){\scriptsize{$\dots$}};
		\draw[rounded corners, very thin, fill=white] (6.75,-2.55) rectangle (8.15,-1.15);\node (D) at (7.45,-1.85){$\pi_k$\footnotesize{($n_k$)}};
		\node (M2) at (0.8, -3.75){Machine $k^\prime$};
  
		\draw[rounded corners, very thin, color=black!50, fill=gray!25] (0,-6.05) rectangle (3.3,-4.05);\node (B1) at (1.65,-5.8){$B_4$};
    \draw[rounded corners, very thin, fill=white] (0.1,-5.55) rectangle (1.5,-4.15);\node (A) at (0.8,-4.85){$\pi_{k^\prime}$\footnotesize{($0$)}};
    \node (D1) at (1.65, -4.85){\scriptsize{$\dots$}};
		\draw[rounded corners, very thin, fill=white] (1.8,-5.55) rectangle (3.2,-4.15);\node (B) at (2.5,-4.85){$\pi_{k^\prime}$\footnotesize{($j$-$1$)}};

    \draw[pattern=north east lines, pattern color=gray!55, rounded corners, very thin] (3.3,-6.05) rectangle (6.6,-4.05);\node (B2) at (4.95,-5.8){$B_2$};
    \draw[rounded corners, very thin, fill=white] (3.4,-5.55) rectangle (4.8,-4.15);\node (C) at (4.1,-4.85){$\pi_k$\footnotesize{($i$)}};
    \node (D2) at (4.95, -4.85){\scriptsize{$\dots$}};
		\draw[rounded corners, very thin, fill=white] (5.1,-5.55) rectangle (6.5,-4.15);\node (D) at (5.8,-4.85){$\pi_k$\footnotesize{($i$+$l$-$1$)}};
		
    \draw[rounded corners, very thin, color=black!50, fill=gray!25] (6.6,-6.05) rectangle (9.9,-4.05);\node (B3) at (8.25,-5.8){$B_5$};
    \draw[rounded corners, very thin, fill=white] (6.7,-5.55) rectangle (8.1,-4.15);\node (E) at (7.4,-4.85){$\pi_{k^\prime}$\footnotesize{($j$)}};
    \node (D3) at (8.25, -4.85){\scriptsize{$\dots$}};
		\draw[rounded corners, very thin, fill=white] (8.4,-5.55) rectangle (9.8,-4.15);\node (F) at (9.1,-4.85){$\pi_{k^\prime}$\footnotesize{($n_{k^\prime}$)}};
  \end{tikzpicture}
	\caption{$l$-block insertion inter-machine}
	\label{fig:l-blockInsertionInter}
\end{figure}
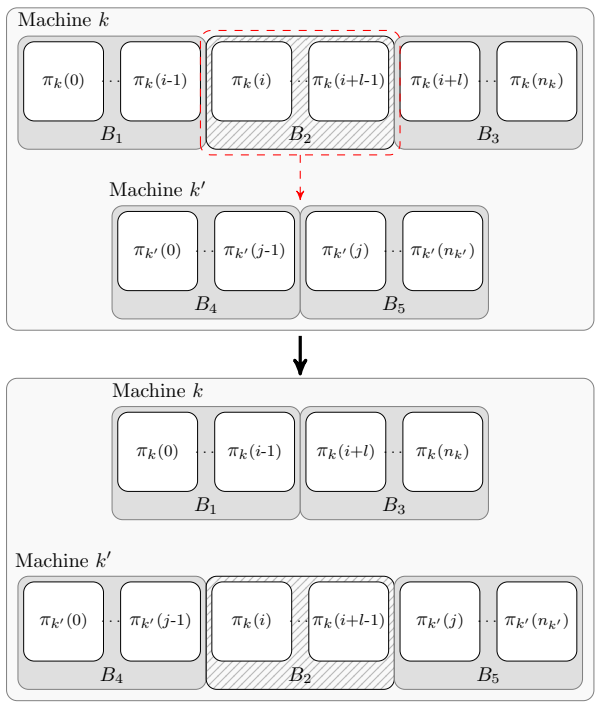  
  
  \item $(l,l')$-block swap inter-machine --- consists of interchanging a block ($B_2$) of size $l$ of a machine $k$, starting from job $\pi_k(i)$, with a block  ($B_5$) of size $l'$ of a machine $k^{\prime}$ that starts from job $\pi_{k^\prime}(j)$, as presented in Fig. \ref{fig:blockswapInter}.

\begin{figure}[!ht]
	\centering
  \begin{tikzpicture}[thick, scale=0.75, every node/.style={transform shape}, >=stealth', dot/.style = {draw, fill = white, circle, inner sep = 0pt, minimum size = 4pt}]
		\draw[rounded corners, very thin, color=black!50, fill=gray!5] (-0.2,6) rectangle (10.1, 0.3);
		\node (M1) at (0.8, 5.8){Machine $k$};
  
		\draw[rounded corners, very thin, color=black!50, fill=gray!25] (0,3.5) rectangle (3.3,5.5);\node (B1) at (1.65,3.75){$B_1$};
    \draw[rounded corners, very thin, fill=white] (0.1,4) rectangle (1.5,5.4);\node (A) at (0.8,4.7){$\pi_k$\footnotesize{($0$)}};
    \node (D1) at (1.65, 4.7){\scriptsize{$\dots$}};
		\draw[rounded corners, very thin, fill=white] (1.8,4) rectangle (3.2,5.4);\node (B) at (2.5,4.7){$\pi_k$\footnotesize{($i$-$1$)}};

    \draw[pattern=north east lines, pattern color=gray!55, rounded corners, very thin] (3.3,3.5) rectangle (6.6,5.5);\node (B2) at (4.95,3.75){$B_2$};
    \draw[rounded corners, very thin, fill=white] (3.4,4) rectangle (4.8,5.4);\node (C) at (4.1,4.7){$\pi_k$\footnotesize{($i$)}};
    \node (D2) at (4.95, 4.7){\scriptsize{$\dots$}};
		\draw[rounded corners, very thin, fill=white] (5.1,4) rectangle (6.5,5.4);\node (D) at (5.8,4.7){$\pi_k$\footnotesize{($i$+$l$-$1$)}};
		
    \draw[rounded corners, very thin, color=black!50, fill=gray!25] (6.6,3.5) rectangle (9.9,5.5);\node (B3) at (8.25,3.75){$B_3$};
    \draw[rounded corners, very thin, fill=white] (6.7,4) rectangle (8.1,5.4);\node (E) at (7.4,4.7){$\pi_k$\footnotesize{($i$+$l$)}};
    \node (D3) at (8.25, 4.7){\scriptsize{$\dots$}};
		\draw[rounded corners, very thin, fill=white] (8.4,4) rectangle (9.8,5.4);\node (F) at (9.1,4.7){$\pi_k$\footnotesize{($n_k$)}};

		\draw[dashed, rounded corners, thin, color=red] (3.2,3.4) rectangle (6.7,5.6);
		\draw[dashed, thin, color=red] (3.95,3.4) [->]-- (3.95,2.6);
		\node (M2) at (0.8, 2.8){Machine $k^\prime$};
  
		\draw[rounded corners, very thin, color=black!50, fill=gray!25] (0,0.5) rectangle (3.3,2.5);\node (B4) at (1.65,0.75){$B_4$};
    \draw[rounded corners, very thin, fill=white] (0.1,1) rectangle (1.5,2.4);\node (A) at (0.8,1.7){$\pi_k$\footnotesize{($0$)}};
    \node (D1) at (1.65, 1.7){\scriptsize{$\dots$}};
		\draw[rounded corners, very thin, fill=white] (1.8,1) rectangle (3.2,2.4);\node (B) at (2.5,1.7){$\pi_k$\footnotesize{($j$-$1$)}};

    \draw[pattern=north east lines, pattern color=gray!55, rounded corners, very thin] (3.3,0.5) rectangle (6.6,2.5);\node (B5) at (4.95,0.75){$B_5$};
    \draw[rounded corners, very thin, fill=white] (3.4,1) rectangle (4.8,2.4);\node (C) at (4.1,1.7){$\pi_k$\footnotesize{($j$)}};
    \node (D2) at (4.95, 1.7){\scriptsize{$\dots$}};
		\draw[rounded corners, very thin, fill=white] (5.1,1) rectangle (6.5,2.4);\node (D) at (5.8,1.7){$\pi_k$\footnotesize{($j$+$l^\prime$-$1$)}};
		
    \draw[rounded corners, very thin, color=black!50, fill=gray!25] (6.6,0.5) rectangle (9.9,2.5);\node (B6) at (8.25,0.75){$B_6$};
    \draw[rounded corners, very thin, fill=white] (6.7,1) rectangle (8.1,2.4);\node (E) at (7.4,1.7){$\pi_k$\footnotesize{($j$+$l^\prime$)}};
    \node (D3) at (8.25, 1.7){\scriptsize{$\dots$}};
		\draw[rounded corners, very thin, fill=white] (8.4,1) rectangle (9.8,2.4);\node (F) at (9.1,1.7){$\pi_k$\footnotesize{($n_{k^\prime}$)}};

		\draw[dashed, rounded corners, thin, color=red] (3.2,0.4) rectangle (6.7,2.6);
		\draw[dashed, thin, color=red] (5.95,2.6) [->]-- (5.95,3.4);
		\draw[->, very thick, color=black] (4.95,0.2) -- (4.95,-0.5);
		\draw[rounded corners, very thin, color=black!50, fill=gray!5] (-0.2,-0.55) rectangle (10.1, -6.25);
		\node (M1) at (0.8, -0.75){Machine $k$};
  
		\draw[rounded corners, very thin, color=black!50, fill=gray!25] (0,-3.05) rectangle (3.3,-1.05);\node (B1) at (1.65,-2.8){$B_1$};
    \draw[rounded corners, very thin, fill=white] (0.1,-2.55) rectangle (1.5,-1.15);\node (A) at (0.8,-1.85){$\pi_k$\footnotesize{($0$)}};
    \node (D1) at (1.65, -1.85){\scriptsize{$\dots$}};
		\draw[rounded corners, very thin, fill=white] (1.8,-2.55) rectangle (3.2,-1.15);\node (B) at (2.5,-1.85){$\pi_k$\footnotesize{($i$-$1$)}};

    \draw[pattern=north east lines, pattern color=gray!55, rounded corners, very thin] (3.3,-3.05) rectangle (6.6,-1.05);\node (B2) at (4.95,-2.8){$B_5$};
    \draw[rounded corners, very thin, fill=white] (3.4,-2.55) rectangle (4.8,-1.15);\node (C) at (4.1,-1.85){$\pi_k$\footnotesize{($j$)}};
    \node (D2) at (4.95, -1.85){\scriptsize{$\dots$}};
		\draw[rounded corners, very thin, fill=white] (5.1,-2.55) rectangle (6.5,-1.15);\node (D) at (5.8,-1.85){$\pi_k$\footnotesize{($j$+$l^\prime$-$1$)}};
		
    \draw[rounded corners, very thin, color=black!50, fill=gray!25] (6.6,-3.05) rectangle (9.9,-1.05);\node (B3) at (8.25,-2.8){$B_3$};
    \draw[rounded corners, very thin, fill=white] (6.7,-2.55) rectangle (8.1,-1.15);\node (E) at (7.4,-1.85){$\pi_k$\footnotesize{($i$+$l$)}};
    \node (D3) at (8.25, -1.85){\scriptsize{$\dots$}};
		\draw[rounded corners, very thin, fill=white] (8.4,-2.55) rectangle (9.8,-1.15);\node (F) at (9.1,-1.85){$\pi_k$\footnotesize{($n_k$)}};
		\node (M2) at (0.8, -3.75){Machine $k^\prime$};
  
		\draw[rounded corners, very thin, color=black!50, fill=gray!25] (0,-6.05) rectangle (3.3,-4.05);\node (B4) at (1.65,-5.8){$B_4$};
    \draw[rounded corners, very thin, fill=white] (0.1,-5.55) rectangle (1.5,-4.15);\node (A) at (0.8,-4.85){$\pi_k$\footnotesize{($0$)}};
    \node (D1) at (1.65, -4.85){\scriptsize{$\dots$}};
		\draw[rounded corners, very thin, fill=white] (1.8,-5.55) rectangle (3.2,-4.15);\node (B) at (2.5,-4.85){$\pi_k$\footnotesize{($j$-$1$)}};

    \draw[pattern=north east lines, pattern color=gray!55, rounded corners, very thin] (3.3,-6.05) rectangle (6.6,-4.05);\node (B5) at (4.95,-5.8){$B_2$};
    \draw[rounded corners, very thin, fill=white] (3.4,-5.55) rectangle (4.8,-4.15);\node (C) at (4.1,-4.85){$\pi_k$\footnotesize{($i$)}};
    \node (D2) at (4.95, -4.85){\scriptsize{$\dots$}};
		\draw[rounded corners, very thin, fill=white] (5.1,-5.55) rectangle (6.5,-4.15);\node (D) at (5.8,-4.85){$\pi_k$\footnotesize{($i$+$l$-$1$)}};
		
    \draw[rounded corners, very thin, color=black!50, fill=gray!25] (6.6,-6.05) rectangle (9.9,-4.05);\node (B6) at (8.25,-5.8){$B_6$};
    \draw[rounded corners, very thin, fill=white] (6.7,-5.55) rectangle (8.1,-4.15);\node (E) at (7.4,-4.85){$\pi_k$\footnotesize{($j$+$l^\prime$)}};
    \node (D3) at (8.25, -4.85){\scriptsize{$\dots$}};
		\draw[rounded corners, very thin, fill=white] (8.4,-5.55) rectangle (9.8,-4.15);\node (F) at (9.1,-4.85){$\pi_k$\footnotesize{($n_{k^\prime}$)}};
		\end{tikzpicture}
		\caption{$(l,l')$-block Swap inter-machine}
    \label{fig:blockswapInter}
\end{figure}
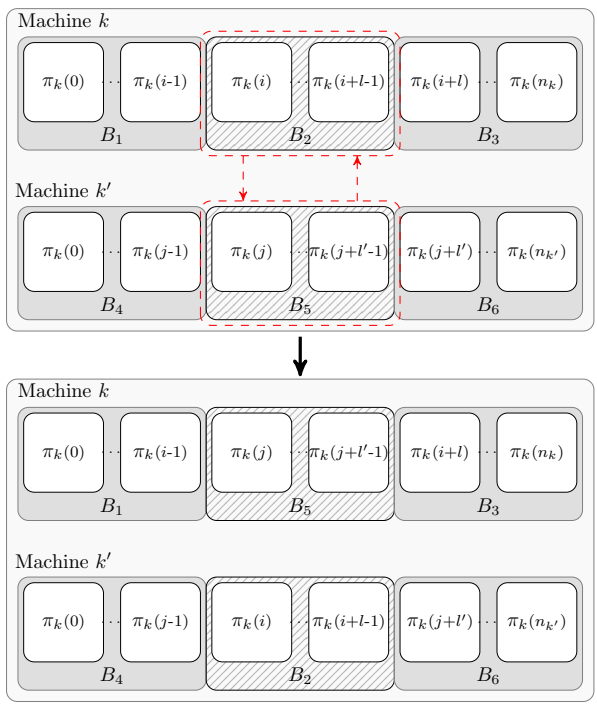  
    
\end{itemize}

The size(s) of the block(s) in each type of neighborhood is an input parameter. Let $L_{intra}$, $L_{inter}$, $L'_{intra}$ and $L'_{inter}$  be the set of possible block sizes for the neighborhoods $l$-block insert intra-machine, $l$-block insert inter-machine, $(l,l')$-block swap intra-machine and $(l,l')$-block swap inter-machine, respectively. For example, if $L_{intra} = \{1, 2,\dots,7\}$, we assume that there are 7 neighborhoods of the type $l$-block insert intra-machine to be considered in the RVND procedure, each with a distinct value for $l$. Similarly, if $L'_{inter} = \{(1,1), (2,1), (2,2), (1,3), (2,3), (3,3)\}$, then the RVND procedure will consider 6 neighborhoods of the type $(l,l')$-block swap inter-machine, each of them with a different setting for $(l, l')$. 

The size of each individual neighborhood of any of the four types mentioned above is $\mathcal{O}(n^2)$. For problems without idle time, a straightforward move evaluation can be performed in $\mathcal{O}(n)$ \citep{SubramanianFarias2015}, leading to an overall complexity of $\mathcal{O}(n^3)$ for enumerating and examining all possible moves from that neighborhood. However, when there are no sequence-dependent setup times, the move evaluation can be performed in amortized 
$\mathcal{O}(1)$ time by extending the ideas presented in \cite{ErgunOrlin2006}, yielding an overall complexity of $\mathcal{O}(n^2)$. 
Otherwise, it is known that the move evaluation can still be performed in amortized
$\mathcal{O}(1)$ time, but with an overall complexity of 
$\mathcal{O}(n^2logn)$,  by extending the approach proposed in \cite{Liaoetal2012}. Nevertheless, it appears that this approach is only worth to be implemented in practice for large sized instances/sequences. 

For problems with idle time, one should solve the timing problem, i.e., determine the optimal start of the jobs associated with the solution under evaluation, which is usually solved by dynamic programming \citep{Ibarakietal2005}, and the resulting overall complexity if often greater than the case without idle time. However, when earliness penalties are not considered and the need for idle time is due to the existence of release dates, the timing problem becomes trivial and one should only ensure that the job does not start before its release date. This can be done in $\mathcal{O}(n)$ time and therefore the overall complexity in this case is $\mathcal{O}(n^3)$. 

The next four sections present different ways of performing the move evaluation according to the characteristics of the problem.


%

\subsubsection{Straightforward move evaluation for problems without idle time}
\label{sec:MoveEval1}

Define $a$ and $b$ as the position associated with the first and last jobs of $u$-$th$ block $B_u$ in the sequence, respectively, and define $b'$ as the position associated with the last job of the preceding block $B_{u-1}$ in the sequence. Let $C\_b'$ be the completion time of the job in position $b'$. Alg. \ref{alg:CompCostBlock} shows how to compute the cost of block $B_u$ in $\mathcal{O}(|B_u|$) steps.

\begin{algorithm}
\caption{CompCostBlock}	 
\label{alg:CompCostBlock}
\footnotesize
\begin{algorithmic}[1]
\Procedure {CompCostBlock}{$b', a, b, \pi_k, C\_b'$}
\State $cost \leftarrow 0$
\State $Ctemp \leftarrow C\_b' + s^{k}_{\pi_k(b'),\pi_k(a)} + p^{k}_{\pi_{k}(a)}$ {\scriptsize \color{gray}\Comment Global variable that stores a temporary completion time}
\If {$Ctemp > d_{\pi_{k}(a)}$}
    \State $cost \leftarrow w_{\pi_{k}(a)} \times \Big(Ctemp - d_{\pi_{k}(a)}\Big)$
\Else
    \State $cost \leftarrow w'_{\pi_{k}(a)} \times \Big(d_{\pi_{k}(a)} - Ctemp\Big)$
\EndIf 
\For {$a' = a+1 \dots b$} \label{loop-start}
    \State $Ctemp \leftarrow Ctemp + s^{k}_{\pi_k(a'-1),\pi_k(a')} + p^{k}_{\pi_{k}(a')}$ 
    \If {$Ctemp > d_{\pi_{k}(a')}$}
	\State $cost \leftarrow cost + w_{\pi_{k}(a')} \times \Big(Ctemp - d_{\pi_{k}(a')}\Big)$
    \Else
	\State $cost \leftarrow cost + w'_{\pi_{k}(a')} \times \Big(d_{\pi_{k}(a')} - Ctemp\Big)$
    \EndIf 
\EndFor \label{loop-end}
\State \Return $cost$
\EndProcedure
\end{algorithmic}
\end{algorithm}

The cost of a neighbor solution can be computed by simply summing up the cost of each block involved in the move. For example, when evaluating the cost of a move related to the $(l,l')$-block swap intra-machine neighborhood, one can directly sum the costs of the blocks $B_1$, $B_4$, $B_3$, $B_2$, and $B_5$ (see Fig. \ref{fig:swapIntra})  in this particular order. Note that the cost of a block depends on the completion time of the last job of the previous sequence, as illustrated in Alg. \ref{alg:CompCostBlock}. In the worst case, the affected blocks may together contain all jobs of the instance, which imply in $\mathcal{O}(n)$ operations for computing the cost. 

In order to speed up the process of computing the cost one can keep track of the cumulated cost up to a particular position of the sequence. Hence, define $W^k_j$ as a data structure that stores the cumulated cost up to the position $j$ of a sequence $\pi_k$, more precisely:
\begin{align*}
W^k_j = \sum_{i=1}^{j}c_i^k, \qquad j = 1,\dots n_k
\end{align*}
where:
\begin{align}
	\label{eq:W}
	c_i^k = \left\{
	\begin{array}{l@{\qquad}l}
		w'_{\pi_{k}(i)}(d_{\pi_{k}(i)}-C_{\pi_k(i)}),\qquad \text{if}\quad C_{\pi_k(i)} \leq d_{\pi_{k}(i)}\\
		w_{\pi_{k}(i)}(C_{\pi_k(i)}-d_{\pi_{k}(i)}),\qquad \text{if}\quad C_{\pi_k(i)} \geq d_{\pi_{k}(i)}
	\end{array}
  \right.
\end{align}

For example, the cost of block $B_1 = (\pi_{k}(0),\dots,\pi_{k}(5))$ can be accessed in $\mathcal{O}(1)$ time by directly verifying the value of $W^k_5$.

We remark that the straightforward way of computing the cost of a move described in this section is not new and we refer to \cite{SubramanianFarias2015} for a detailed description of the move evaluation using this type of approach. It is worthy of note that the same authors developed a local search limitation strategy, based on the setup variation due to a move, for problem $1|s_{ij}|\sum w_jT_j$ that turned out to be very efficient in practice. They implemented a filtering mechanism that avoids unpromising moves to be evaluated. We believe that this idea can be extended for problems involving multiple machines and sequence-dependent setup times.

\subsubsection{Move evaluation for problems without both idle time and sequence-dependent setup times}
\label{sec:MoveEval2}

\citet{ErgunOrlin2006} presented ways of evaluating the moves of the neighborhoods $1$-block insertion intra-machine, $1$-block swap intra-machine and block reverse (a.k.a. twist) in amortized $\mathcal{O}(1)$ time for problem $1||\sum w_jT_j$. In this work we extend their approaches for the neighborhoods described in Section \ref{sec:Neighborhoods} to solve problem $R||\sum w'_jE_j + w_jT_j$ (without idle time) and its particular cases, that is, all problems without sequence-dependent setup times and release dates shown in Table \ref{tab:Problems}. 


Before describing the move evaluation schemes for all neighborhoods, we will introduce some useful auxiliary functions and data structures.

Let $\rho_{j}^{k}(t)$ be a non-negative, convex and piecewise linear function that represents the penalty of start processing job $j \in J$ in machine $k \in M$ at time $t$: 
\begin{align} 
\label{eq:funcPen}
\rho_{j}^{k}(t) =
\left\{
\begin{array}{l@{\qquad}l}
  M (r_j-t) + w^{\prime}_j (d_j-p_{j}^{k} - t),\qquad t \in [0, r_j]\\
  w^{\prime}_j (d_j-p_{j}^{k} - t) ,\qquad t \in [r_j, d_j-p_{j}^{k}]\\
  w_j (t - d_j + p_{j}^{k}) ,\qquad t \in [d_j-p_{j}^{k}, +\infty)\\
\end{array}
\right.
\end{align}
\noindent where $M$ is constant, positive and sufficiently large, used to penalize release date violation. For the ease of presentation we assume that $r_j \leq d_j - p_j^k$. If this condition is not satisfied in practice, i.e. $r_j > d_j - p_j^k$, then the functions above should be modified accordingly.

Each function $\rho^{k}_{j}(t)$ is composed of three segments, where the transition points, a.k.a. \emph{breakpoints}, are defined by $r_j$, $d_j$ and $p^{k}_{j}$. Release dates were considered for the sake of generality, but their existence typically enforce the presence of idle times. Since we are not considering idle times in this section, we can simply disregard one of the segments by setting $r_j = 0, \forall j \in J$, and the functions will remain valid, but with two segments each.


Now, for a given sequence $\pi_k$, let $g_{j}^{k}(t)$ be a piecewise linear function denoting the cost of a block composed of jobs  $\pi_{k}(j), \pi_{k}(j+1),\dots,\pi_{k}(n_{k})$ sequenced in machine $k$ in this order and that job $\pi_{k}(j)$ starts to be processed at time $t$. Note that $n_k$ is the index of the last job assigned to machine $k$. Each function  $g_{j}^{k}(t)$,  $\forall j \in J, \forall k \in M$, can be determined by the following recursion:
\begin{equation*}
\label{eq:funcE&O1}
g_{j}^{k}(t) = 
\left\{
\begin{array}{l@{\qquad}l}
	\rho_{j}^{k}(t),\qquad \qquad \qquad \qquad \qquad \qquad \substack{ j = n_{k},\\ t \in [0, +\infty)}\\
	g_{j+1}^{k}(t+p_{j}^{k} + s^k_{j,j+1}) + \rho_{j}^{k}(t),\qquad \substack{j \in (0,n_{k}) \\ t \in (0,+\infty)}\\
\end{array}
\right.
\end{equation*}

To better clarify the meaning of functions $\rho_{j}^{k}(t)$ and $g_{j}^{k}(t)$, consider the example given in Table \ref{tab:example}, but disregarding the release dates and setup times.
\begin{table}[!htbp]
  \footnotesize
  \centering
  \caption{Example involving 5 jobs on a single machine}
  \begin{tabular}{ccccc}
  \hline 
  $j$ &  $p_j^1$ & $d_j$  & $w'_j$ & $w_j$ \\ \hline
  $1$ &  $3$  & $8$ & $2$ & $4$ \\
  $2$ &  $2$ & $7$ & $1$ & $2$ \\
  $3$ &  $1$ & $4$ & $2$ & $4$\\
  $4$ &  $3$ & $3$ & $1$ & $3$\\
  $5$ &  $1$ & $13$ & $1$ & $1$\\ \hline
  \end{tabular}
  \label{tab:example}
\end{table}
Now assuming, for instance, that $\pi_1 = (0, 1, 2)$, we can write functions $\rho_{1}^{1}(t), \rho_{2}^{1}(t), g_{1}^{1}(t)$ and $g_{2}^{1}(t)$ as follows:
\begin{equation*}
\rho^{1}_{1}(t) =
\left\{
\begin{array}{l@{\qquad}l}
 10 -2t, \quad t\in[0,5]\\
 4(t-5), \quad t\in[5,+\infty)\\
\end{array}
\right.
\end{equation*} 
 
\begin{equation*}
\rho^{1}_{2}(t) =
\left\{
\begin{array}{l@{\qquad}l}
 5 -1t, \quad t\in[0,5]\\
 2(t-5), \quad t\in[5,+\infty)\\
\end{array}
\right.
\end{equation*}

\begin{equation*}
g^{1}_{2}(t+p^{1}_{1}) =
\left\{
\begin{array}{l@{\qquad}l}
 2 -1t, \quad t\in[0,2]\\
 2(t-2), \quad t\in[2,+\infty)\\
\end{array}
\right.
\end{equation*}

\begin{equation*}
g^{1}_{1}(t) =
\left\{
\begin{array}{l@{\qquad}l}
 12 -3t, \quad t\in[0,2]\\
 6, \quad t\in[2,5]\\
 6 + 6(t-5),\quad t\in[5,+\infty)\\
\end{array}
\right.
\end{equation*}

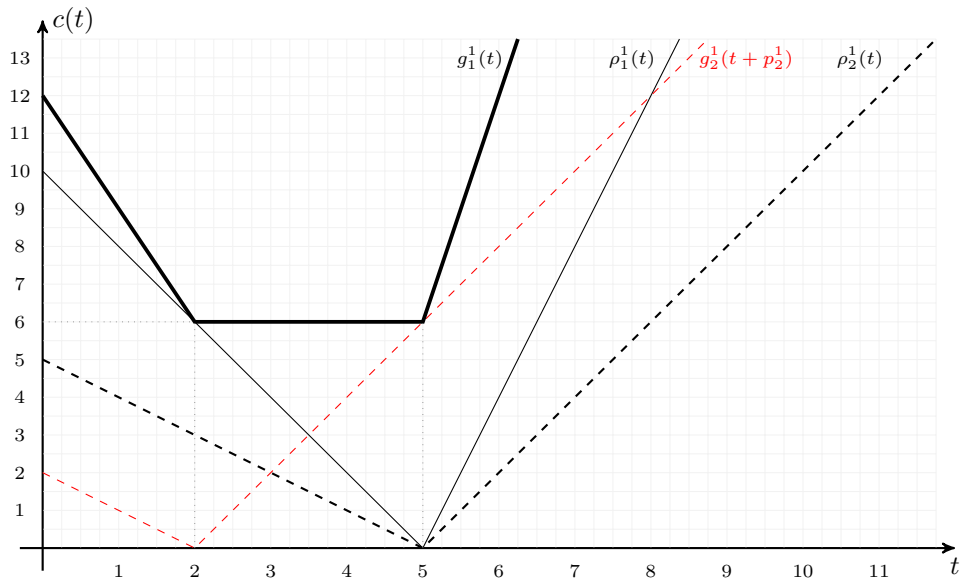
\begin{figure}[!ht]
\centering
  \begin{tikzpicture}[
      thick,
      scale=1.00, every node/.style={transform shape},
      >=stealth',
      dot/.style = {
			draw,
			fill = white,
			circle,
			inner sep = 0pt,
			minimum size = 4pt
      }
    ]
    \coordinate (O) at (0,0);
    \draw[->] (-0.3,0) -- (12,0) coordinate[label = {below:$t$}] (xmax);
    \draw[->] (0,-0.3) -- (0,7) coordinate[label = {right:$c(t)$}] (ymax);
    \draw[step=.25cm,gray!10,very thin] (0.01, 0.01) grid (11.75,6.75);

    \draw[-,thin] (0, 5) -- (5, 0) -- (8.375, 6.75); 
    \draw[-,dashed] (0, 2.5) -- (5, 0) -- (11.75, 6.75); 
    \draw[-,dashed,thin,red] (0, 1) -- (2, 0) -- (8.75, 6.75); 
    \draw[-,line width=0.5mm, black] (0,6) -- (2,3) -- (5,3) -- (6.25,6.75); 
    \scriptsize{
      \node (B1) at (1,-0.3) {$1$};
      \node (C1) at (2,-0.3) {$2$};
      \node (D1) at (3,-0.3) {$3$};
      \node (E1) at (4,-0.3) {$4$};
      \node (F1) at (5,-0.3) {$5$};
      \node (G1) at (6,-0.3) {$6$};
      \node (H1) at (7,-0.3) {$7$};
      \node (I1) at (8,-0.3) {$8$};
      \node (J1) at (9,-0.3) {$9$};
      \node (K1) at (10,-0.3) {$10$};
      \node (L1) at (11,-0.3) {$11$};     
      \node (B1) at (-0.3,0.5) {$1$};
      \node (D1) at (-0.3,1) {$2$};
      \node (F1) at (-0.3,1.5) {$3$};
      \node (H1) at (-0.3,2) {$4$};
      \node (J1) at (-0.3,2.5) {$5$};
      \node (L1) at (-0.3,3) {$6$};
      \node (L1) at (-0.3,3.5) {$7$};
      \node (L1) at (-0.3,4) {$8$};
      \node (L1) at (-0.3,4.5) {$9$};
      \node (L1) at (-0.3,5) {$10$};
      \node (L1) at (-0.3,5.5) {$11$};
      \node (L1) at (-0.3,6) {$12$};
      \node (L1) at (-0.3,6.5) {$13$};
      \draw[-, dotted, line width=0.1mm, black!50] (0, 3) -- (2, 3);
      \draw[-, dotted, line width=0.1mm, black!50] (2, 0) -- (2, 3);
      \draw[-, dotted, line width=0.1mm, black!50] (5, 0) -- (5, 3);
      \node(g) at (5.75, 6.5) {{$g_{1}^{1}(t)$}};
      \node(rho1) at (7.75, 6.5) {{$\rho_{1}^{1}(t)$}};
      \node(rho2_t+t1) at (9.25, 6.5) {\textcolor{red}{$g_{2}^{1}(t+p_{2}^{1})$}};
      \node(rho2) at (10.75, 6.5) {{$\rho_{2}^{1}(t)$}};
    } 
  \end{tikzpicture}
	\caption{Penalty functions}
	\label{fig:funcPen}
\end{figure}

Fig. \ref{fig:funcPen} illustrates a graphical representation of the piecewise linear functions.

\begin{figure}[!ht]
  \begin{center}
    \includegraphics[scale=0.45]{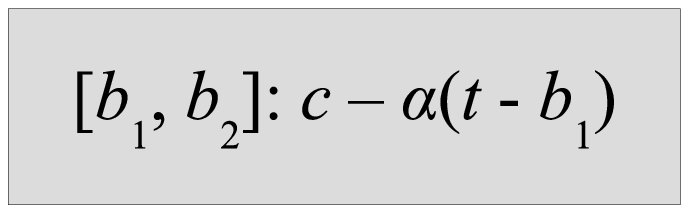}
    \caption{Representation of a segment}
     \label{fig:segment}
  \end{center}
\end{figure}

Functions $\rho^{k}_{j}(t)$ e $g^{k}_{j}(t)$ can be stored in memory by means of linked lists, here denoted as \texttt{L\_j\_k}, where each element of the list is associated with one of the segments of the piecewise linear function. We use a data structure called \texttt{seg}  to represent each segment of the function. Such data structure (see Fig. \ref{fig:segment}) stores the following piece of information:
\begin{itemize}
 \item $\texttt{b}_{\texttt{1}}$: beginning of the segment's interval;
 \item $\texttt{b}_{\texttt{2}}$: end of the segment's interval;
 \item $\texttt{c}$: value of the function at time $b_1$;
 \item $\alpha$: slope of the segment.
\end{itemize}

\begin{figure}[!ht]
  \begin{center}
    \includegraphics[scale=0.45]{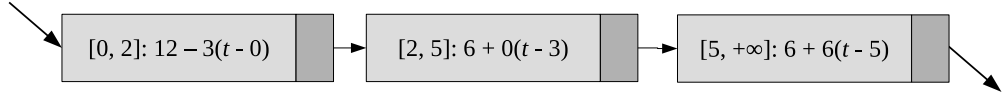}
    \caption{Representation of function  $g_1^1(t)$ using a linked list}
     \label{fig:list}
  \end{center}
\end{figure}

Fig. \ref{fig:list} shows an example of how function $g_1^1(t)$ of Fig. \ref{fig:funcPen} is stored in memory using the linked list \texttt{L\_1\_1}.

In order to compute the cost of a block $(\pi_{k}(j), \pi_{k}(j+1),\dots,\pi_{k}(n_{k}))$ starting at time $t$ by means of functions $g^{k}_{j}(t)$ using the information mentioned above, one needs to walk through the breakpoints of the corresponding function until the interval $b_1 \leq t \leq b_2$ is reached and then compute the cost which is given by $c + \alpha \times (t-b_1)$.

Finally, another important data structure is the so-called \texttt{ProcessingList} \citep{ErgunOrlin2006}, which in our case is defined for all blocks of size $l$ from a given sequence $\pi_{k}$. Each element of this list contains a pair of information, namely: \texttt{pos} and \texttt{p}, where the first stores the position of the first job of the block in the sequence, while the second stores the total processing time of the block in machine $k'$.  The \texttt{ProcessingList} must be sorted in 
descending order according to the value of \texttt{p}.  Alg. \ref{alg:ProcList} shows how the \texttt{ProcessingList} is created.

\begin{algorithm}[!ht]
\caption{CreateProcessingList}	 
\label{alg:ProcList}
\footnotesize
\begin{algorithmic}[1]
\Procedure {CreateProcessingList}{$\pi_k, k^{\prime}, l$}
  \State $p_{B} \leftarrow 0$ {\scriptsize \color{gray}\Comment $p_B$ is the total processing time of the block of size $l$}
  \For {$i^{\prime} = 1,\dots,l$}
    \State $p_{B} \leftarrow p_{B} + p^{k^{\prime}}_{\pi_{k}(i^{\prime})}$
  \EndFor
  \For {$i \leftarrow 1,\dots, n_{k}-l+1$}
    \State \texttt{ProcessingList.pos[i]}$\leftarrow i$
    \State \texttt{ProcessingList.p[i]} $\leftarrow p_{B} $
    \State $p_{B} \leftarrow p_{B} - p^{k^{\prime}}_{\pi_{k}(i)} + p^{k^{\prime}}_{\pi_{k}(i+l)}$
  \EndFor
  \State \texttt{sort(ProcessingList,p)}
\State \Return \texttt{ProcessingList}
\EndProcedure
\end{algorithmic}
\end{algorithm}

Once the auxiliary functions and data structures were defined, it is now possible to show how to compute the cost of a sequence $\pi_k^{\prime}$ generated after applying one of the neighborhood operators. 

For a given sequence $\pi_k$,  $g^{k}_{j}(t)$ can have at most $n_k$ breakpoints. Since $g^{k}_{j}(t)$ are piecewise linear functions, the complexity of obtaining the cost of a block $(\pi_{k}(j), \pi_{k}(j+1),\dots,\pi_{k}))$ starting at time $t$ is, in principle, $\mathcal{O}(n_k)$, because in the worst case $t$ can be greater than the last breakpoint of the function. Hence, the complexity of evaluating a move would be $\mathcal{O}(n_k)$. However, it is possible to evaluate a move of the neighborhoods in amortized $\mathcal{O}(1)$ time if the costs that depend on $g^{k}_{j}(t)$ are precomputed following a given order.

Table \ref{tab:moveEval} shows how to compute the cost of the modified solution in amortized $\mathcal{O}(1)$ time given that the costs that depend on $g^{k}_{j}(t)$ are precomputed in $\mathcal{O}(n_k^2)$.
The first column identifies the blocks related to the move using the same convention adopted in Figures \ref{fig:l-blockInsertionFwd}-\ref{fig:blockswapInter}, while the remaining ones show how to compute the cost of each particular block depending on the neighborhood and machines involved in the move. Also, as shown in the referred figures, we can observe that the number of resulting blocks of a move varies from 4 to 6, according to the neighborhood. For example, a move of the neighborhood $l$--block insertion inter considers blocks $B_1$ and $B_3$ on machine $k$, and blocks $B_2, B_4$ and $B_5$ on machine $k^\prime$. For a given solution, the indices $i$ and $j$ refer to the initial position of the blocks associated with the move (see Section \ref{sec:Neighborhoods} for more details). Hence, the cost of a neighbor solution can be computed by summing up the costs of the blocks involved in the move.

\begin{table*}[!htbp]
\centering
\caption{Move evaluation schemes}
\scriptsize
\renewcommand{\tabcolsep}{0.08cm}
\begin{tabular}{|c|c|c|cc|cc|}
\multicolumn{7}{c}{Complexity of performing a single move evaluation in amortized $\mathcal{O}(1)$ time}\\
\hline
 & \multicolumn{1}{c|}{$l$-block insertion intra} & \multicolumn{1}{c|}{($l$-$l'$)-block swap intra} & \multicolumn{2}{c|}{$l$-block insertion inter}  & \multicolumn{2}{c|}{($l$-$l'$)-block swap inter} \\ 
\cline{2-7}
\up{Block} & $k$  &  $k$  &  $k$  &  $k^\prime$  &  $k$  &  $k^\prime$  \\ \hline
$B_1$  & $W_{i-1}^{k}$  &  $W_{i-1}^{k}$  &  $W_{i-1}^{k}$ & - &  $W_{i-1}^{k}$  & - \\
$B_2$  &  $g_{i}^{k}(t_2) - g_{i+l}^{k}(t_3)$  & $g_{i}^{k}(t_5) - g_{i+l}^{k}(t_6)$ & - &  $g_{i}^{k^\prime}(t_{11}) - g_{i+l}^{k^\prime}(t_{10})$  & - &  $g_{i}^{k^\prime}(t_{11}) - g_{i+l}^{k^\prime}(t_{10})$\\
$B_3$  &  $g_{i+l}^{k}(t_1) - g_{j+1}^{k}(t_2)$  & $g_{i+l}^{k}(t_4) - g_{j}^{k}(t_5)$  &  $g_{i+l}^{k}(t_{9})$  & - &  $g_{i+l}^{k}(t_{12})$  & - \\
$B_4$  &  $W_{n_k}^{k} - W_{j}^{k}$  &  $g_{j}^{k}(t_7) - g_{j+l^\prime}^{k}(t_{8})$ & - &  $W_{j-1}^{k^\prime}$  & - &  $W_{j-1}^{k^\prime}$\\
$B_5$  & - &  $W_{n_k}^{k} - W_{j+l^\prime -1}^{k}$  & - &  $g_{j}^{k^\prime}(t_{10})$  &  $g_{j}^{k}(t_{9}) - g_{j+l^\prime}^{k}(t_{12})$ & - \\
$B_6$  &  - &  - &  -  &  - &  - &  $g_{j+l^\prime}^{k^\prime}(t_{10})$\\
\hline
\multicolumn{7}{c}{Complexity of performing a single move evaluation in amortized $\mathcal{O}(l)$ / $\mathcal{O}(\max\{l,l^\prime\})$ time}\\
\hline
& \multicolumn{1}{c|}{$l$-block insertion intra} & \multicolumn{1}{c|}{($l$-$l'$)-block swap intra} & \multicolumn{2}{c|}{$l$-block insertion inter}  & \multicolumn{2}{c|}{($l$-$l'$)-block swap inter} \\ 
\cline{2-7}
\up{Block} & $k$  &  $k$  &  $k$  &  $k^\prime$  &  $k$  &  $k^\prime$  \\ \hline
$B_1$  & $W_{i-1}^{k}$  &  $W_{i-1}^{k}$  &  $W_{i-1}^{k}$ & - &  $W_{i-1}^{k}$  & - \\ 
$B_2$  &  \texttt{CompCostBlock}$(B_2)$  &\texttt{CompCostBlock}$(B_2)$  & - &  \texttt{CompCostBlock}$(B_2)$  & - &  \texttt{CompCostBlock}$(B_2)$\\
$B_3$  &  $g_{i+l}^{k}(t_1) - g_{j+1}^{k}(t_2)$  & $g_{i+l}^{k}(t_4) - g_{j}^{k}(t_5)$  &  $g_{i+l}^{k}(t_{9})$  & - &  $g_{i+l}^{k}(t_{12})$  & - \\
$B_4$  &  $W_{n_k}^{k} - W_{j}^{k}$  &  \texttt{CompCostBlock}$(B_4)$ & - &  $W_{j-1}^{k^\prime}$  & - &  $W_{j-1}^{k^\prime}$\\
$B_5$  & - &  $W_{n_k}^{k} - W_{j+l^\prime -1}^{k}$  & - &  $g_{j}^{k^\prime}(t_{10})$  &  \texttt{CompCostBlock}$(B_5)$  & - \\
$B_6$  &  - &  - &  -  &  - &  - &  $g_{j+l^\prime}^{k^\prime}(t_{10})$\\
\hline
\multicolumn{7}{c}{Where:}\\
\hline
\multicolumn{7}{|l|}{  $t_{1} = C^{k}_{i-1}$; \quad $t_{2} = C^{k}_{j} - \displaystyle\sum_{a = i}^{i+l-1} p^{k}_{\pi_{k}(a)}$; \quad  $t_{3} = C^{k}_{j}$; \quad $t_{4} = C^{k}_{i-1} + \displaystyle\sum_{a = j}^{j+l^\prime-1} p^{k}_{\pi_{k}(a)}$;}\\
\multicolumn{7}{|l|}{$t_{5} = C^{k}_{j+l^\prime-1} - \displaystyle\sum_{a = i}^{i+l-1} p^{k}_{\pi_{k}(a)}$; \quad  $t_{6} = C^{k}_{j+l^\prime-1}$; \quad  $t_{7} = C^{k}_{i-1}$; \quad $t_{8} = C^{k}_{i-1} + \displaystyle\sum_{a = j}^{j+l^\prime-1} p^{k}_{\pi_{k}(a)}$;}\\
\multicolumn{7}{|l|}{$t_{9} = C^{k}_{i-1} $; \quad $t_{10} = C^{k^\prime}_{j-1} + \displaystyle\sum_{a = i}^{i+l-1} p^{k^\prime}_{\pi_{k}(a)}$; \quad $t_{11} = C^{k^\prime}_{j-1}$; \quad $t_{12} = C^{k}_{i-1} + \displaystyle\sum_{a = j}^{j+l^\prime-1} p^{k}_{\pi_{k^\prime}(a)}$.}\\
\hline
\end{tabular}
\label{tab:moveEval}
\end{table*}

Nevertheless, since we adopted very small values for $l$ and $l'$, as will be shown in Section \ref{sec:Results}, we decided to implement the move evaluation schemes with complexity  $\mathcal{O}(l_{\max})$, where $l_{\max} = \max \{l,l^\prime \}$, as also shown in Table \ref{tab:moveEval} rather than  $\mathcal{O}(1)$, which results in an overall complexity of $\mathcal{O}(l_{\max} n_k^2)$, but in practice it seems to offer a more interesting scalability, even for large size instances. This happens because the preprocessing phase
requires an additional overhead for dealing with blocks of size $l_{\max} > 1$ that in the end does not compensate the advantage of performing the move evaluations in $\mathcal{O}(1)$ steps.

The preprocessing consists of precomputing the costs of the blocks that depend on $g^k$ in a particular order. This enables the move evaluation procedure to access the cost of those blocks in constant time. For example, when performing the move evaluation of a neighbor solution after applying a move of the neighborhood $l$-block insertion intra, one can compute the associated cost in constant time by means of the following expression: $W_{i-1}^{k} + [g_i^k(t_2) - g_{i+l}^{k}(t_3) ] + [g_{i+l}^{k}(t_1) - g_{j+1}^{k}(t_2)] + [W_{n_k}^{k} - W_{j}^{k}]$ (see Table \ref{tab:moveEval}). The terms of this expression that depend on $W^k$ are precomputed in $\mathcal{O}(n_k)$ steps, as shown in \eqref{eq:W}, while the terms of this expression that depend on $g^k$ can be precomputed in $\mathcal{O}(n_k^2)$ steps, as explained next.


For example, in order to store all values of $g_i^k(t_2 = C_j^k - \sum_{a=i}^{i+l-1}p_{\pi_k}^{k}(a))$ in $\mathcal{O}(n_k^2)$ steps, it is first necessary to sort the values of $C_j^k - \sum_{a=i}^{i+l-1}p_{\pi_k}^{k}(a)$ in ascending order. Note that this is possible to be achived by sorting the values of  $C_j^k$ in ascending order, which can be done in $\mathcal{O}(n_k \log n_k)$ by using a standard sorting procedure. Next, for every $i$, it is now possible to compute the values of $g_i^k(t_2 = C_j^k - \sum_{a=i}^{i+l-1}p_{\pi_k}^{k}(a))$ in $\mathcal{O}(n_k)$ steps, because this function has at most $n$ segments, which leads to an overall complexity of $\mathcal{O}(n_k^2)$. A similar rationale can be applied to the other neighborhoods.

We will conclude this section by presenting alternative and simpler local search algorithms for the $l$-block insertion neighborhoods that do not rely on the auxiliary data structures and functions mentioned above. The overall complexity of the search using such algorithms is  $\mathcal{O}(l_{\max}n_k^2)$ and they are also based on the work of \cite{ErgunOrlin2006}.

In the case of the intra-machine neighborhood, two similar but distinct procedures are required to move a block forward and backward in the sequence, respectively. Alg. \ref{alg:l-blockInsertionIntraFwd} shows how the moves are evaluated when a block of size $l$ is removed and reinserted in a forward position in the sequence. It can be observed that moves are evaluated in $\mathcal{O}(l)$ steps (lines \ref{alg:l-blockInsertionIntraFwd:begMoveEval}-\ref{alg:l-blockInsertionIntraFwd:endMoveEval}) by making use of the move evaluation performed immediately before the current one. The same rationale can be applied for moving blocks of size $l$ to a backward position in the sequence. Therefore, it can be verified that the overall complexity of Alg. \ref{alg:l-blockInsertionIntraFwd} is $\mathcal{O}(ln_k^2)$.

\begin{algorithm*}[!ht]
\caption{$l$--block insertion intra forward}	 
\label{alg:l-blockInsertionIntraFwd}
\footnotesize
\begin{algorithmic}[1]
\Procedure {$l$-BlockInsertionIntraF}{$\pi_k, f_{\pi_{k}}, l, C$} 
\State $\pi_k^* \leftarrow \pi_{k}; f_k^*\leftarrow f_{\pi_{k}}$ {\scriptsize \color{gray}\Comment  $\pi_k$ is the current sequence of machine $k$ with associated cost $f_{\pi_k}$ and $\pi_k^*$ is the best improving neighbor sequence with associated cost $f_k^*$}
\State $p_{B}\leftarrow 0$ {\scriptsize \color{gray}\Comment $p_B$ is the total processing time of the block considered for reinsertion}
\For {$i'=1,\dots,l$}
  \State $p_{B}\leftarrow p_{B} + p^{k}_{\pi_{k}(i')}$
\EndFor
\For {$i=1,\dots,n_{k}-l$} {\scriptsize \color{gray}\Comment $n_k$ is the number of jobs in machine $k$} 
  \State $f_{\pi_k^\prime}\leftarrow f_{\pi_{k}}$; {\scriptsize \color{gray}\Comment $f_{\pi_k^\prime}$ is the cost of the neighbor sequence $\pi_k^\prime$ under evaluation}
  \For {$j=i+l,\dots,n_{k}-1$}
    \State ${lateness} \leftarrow C_{\pi_{k}(j)}-p_{B}-d_{\pi_{k}(j)}$ \label{alg:l-blockInsertionIntraFwd:begMoveEval}
    \State $f_{\pi_k^\prime} \leftarrow f_{\pi_k^\prime} + \max\{w_{\pi_{k}(j)} \times {lateness}, \text{ } w'_{\pi_{k}(j)} \times (-{lateness})\}$
    \State $f_{\pi_k^\prime} \leftarrow f_{\pi_k^\prime} - f_{\pi_{k}(j)}$ {\scriptsize{\color{gray}\Comment $f_{\pi_{k}(j)}$ is the cost of job $\pi_{k}(j)$ in the sequence}}
     \State $f_{\pi_k^\prime} \leftarrow f_{\pi_k^\prime} + \texttt{CompCostBlock}(j, i, {i+l-1}, \pi_k, \pi_k, C_{\pi_{k}(j)} - p_B)$ \label{alg:l-blockInsertionIntraFwd:endMoveEval} 
    \If {$f_{\pi_k^\prime} < f_k^*$}
      \State $\pi_k^* \leftarrow \pi_k^\prime; f_k^* \leftarrow f_{\pi_k^\prime}$ \label{alg:l-blockInsertionIntraFwd:endMoveEval2} 
    \EndIf
  \EndFor
  \State $p_{B}\leftarrow p_{B}-p^{k}_{\pi_{k}(i)} + p^{k}_{\pi_{k}(i+l)}$
\EndFor
\State \Return $f_k^*$
\EndProcedure
\end{algorithmic}
\end{algorithm*}

Alg. \ref{alg:l-blockInsertionInter} shows the pseudocode of the neighborhood $l$-block insertion inter-machine. Note that when a block of size $l$ is removed from machine $k$, one needs to compute the cost of the new sequence of $k$ only once and this can be done in $\mathcal{O}(n_k)$ steps as in Section \ref{sec:MoveEval1}. As for machine $k^{\prime}$, where the block is going to be inserted in several positions, the cost of each modified sequence can be computed in $\mathcal{O}(l)$ using the same idea presented in Alg. \ref{alg:l-blockInsertionIntraFwd} after computing the cost of inserting the block in the first position in $\mathcal{O}(n_{k^\prime})$ also as in Section \ref{sec:MoveEval1}. Hence, the overall complexity of Alg. \ref{alg:l-blockInsertionInter} is $\mathcal{O}(ln_k^2)$.

\begin{algorithm*}[!ht]
\caption{$l$--block insertion inter}        
\label{alg:l-blockInsertionInter}
\footnotesize
\begin{algorithmic}[1]
\Procedure {$l$-BlockInsertionInter}{$\pi, f, l$}
\State {$\pi^{*} \leftarrow \pi, f^{*} \leftarrow f_\pi$} {\scriptsize \color{gray}\Comment  $\pi$ is the current solution with associated cost $f_\pi$ and $\pi^*$ is the best improving neighbor with associated cost $f^*$}
\For {$k = 1,\dots m$}
   \For {$k^{\prime} = 1,\dots m$}
     \If{$k \neq k^{\prime}$}
       \For {$i=1,\dots,n_{k}-l+1$}
         \State $f_{\pi_{k}^{\prime}} \leftarrow W^{k}_{i-1} + \texttt{CompCostBlock}\big(i-1,i+l,n_{k},\pi_{k},\pi_k^{\prime} ,C_{\pi_{k}(i-1)}\big)$\ {\scriptsize \color{gray}\Comment  $f_{\pi_k^{\prime}}$ is the cost of the modified sequence $\pi_k^{\prime}$}
         \State $\tilde{\pi}_{k^{\prime}} \leftarrow \langle \pi_{k}(i),\dots,\pi_{k}(i+l-1) \rangle \oplus \langle \pi_{k^{\prime}}\rangle$ 
	  \State $f_{\pi_{k^{\prime}}^{\prime}} \leftarrow \texttt{l-blockInsertionIntraF}(\tilde{\pi}_{k^{\prime}}, f_{\tilde{\pi}_{k^{\prime}}}, l)$ \underline{only for $i=1$} (lines 
\ref{alg:l-blockInsertionIntraFwd:begMoveEval}-\ref{alg:l-blockInsertionIntraFwd:endMoveEval2} of Alg. \ref{alg:l-blockInsertionIntraFwd}) {\scriptsize \color{gray}\Comment  $f_{\pi_{k^{\prime}}^{\prime}}$ is the cost of the modified sequence $\pi_{k^{\prime}}^{\prime}$}
         \If {$f_\pi - f_{\pi_{k}} - f_{\pi_{k^{\prime}}} + f_{\pi_{k}^{\prime}} + f_{\pi_{k^{\prime}}^{\prime}} < f^{*}$}
           \State {$f^{*} \leftarrow f_\pi - f_{\pi_{k}} - f_{\pi_{k^{\prime}}} + f_{\pi_{k}^{\prime}} + f_{\pi_{k^{\prime}}^{\prime}}$}
           \State $\pi^{*} \leftarrow \pi^{\prime}$
         \EndIf
       \EndFor
     \EndIf
   \EndFor
 \EndFor 
\State \Return $f^*$
\EndProcedure
\end{algorithmic}
\end{algorithm*}

\subsubsection{Move evaluation for problems with sequence-dependent setup times but without idle time}
\label{sec:MoveEval3}

\cite{Liaoetal2012} showed, for problem $1|s_{ij}|w_jT_j$, that is possible to perform the move evaluations of the same neighborhoods considered by \cite{ErgunOrlin2006} in amortized $\mathcal{O}(1)$ time, but at the expense of preprocessing the required auxiliary data structures in $\mathcal{O}(n^2\log n)$ operations. Their method is based on the one of the latter authors and it can be extended to solve problem $R|s_{ij}^{k}|\sum{w_{j}^{\prime}E_{j} + w_{j}T_{j}}$ and its particular cases (only those without idle time) by using the same rationale presented in Section \ref{sec:MoveEval2}. 

The main difference between the approaches of \cite{Liaoetal2012} and \cite{ErgunOrlin2006} is the preprocessing phase, where the sorting procedure should be called for each $i$  before computing the values of $g_i^k(t)$ for each $j$ in the sequence, because $t$ now depends on the existing setup times, thus leading to a complexity of $\mathcal{O}(n^2logn)$.

Nevertheless, this approach seems to be indicated for large scale instances, as suggested by the experiments performed in \cite{Liaoetal2012}. For small-medium scale instances the procedure described in Section \ref{sec:MoveEval1} appears to be more suitable to address the problems with sequence-dependent setup times listed in Table \ref{tab:Problems}.

\subsubsection{Move evaluation for problems with idle time}
\label{sec:MoveEval4}

\cite{Ibarakietal2005} proposed an efficient approach for evaluating the moves during the local search when idle times are considered. In this case, the \emph{timing problem}, which consists of determining the optimal start time of each job in the sequence, should also be solved when evaluating the cost of a solution. The authors first present a dynamic programming based algorithm to solve this problem and then they show how to integrate such method within the local search.

As in \cite{ErgunOrlin2006}, the method of \cite{Ibarakietal2005} relies on auxiliary data structures and functions. The idea is to determine functions that return the minimum penalty of the subsequences (blocks) when starting at a particular time $t$, both forward and backward. The cost of the solution under evaluation is thus computed by using such information over the resulting sequence obtained after concatenating certain subsequences.

In \cite{Ibarakietal2008} the authors suggested a balanced binary tree based implementation that allows for solving the timing problem in $\mathcal{O}(n_k\log\delta_k)$ time, where  $\delta_k$ denotes the total number of segments of the penalty function for the jobs sequenced in machine $k$. As demonstrated by the authors, a move can be evaluated in amortized  $\mathcal{O}(\log\delta_{\max})$ time, where 
$\delta_{\max}$ is the largest $\delta_k$ among the machines involved. The necessary functions of each machine $k$ can be precomputed in $\mathcal{O}(\delta_k\log\delta_k)$ steps. However, for simplicity, we used a linked list based implementation \citep{Ibarakietal2005}, which yields an amortized complexity of $\mathcal{O}(\delta_{\max})$ where the necessary functions for each machine $k$ can be precomputed in $\mathcal{O}(n_k\delta_k)$ steps.

For the neighborhoods considered in this work, the modified solutions can be obtained by concatenating at most three subsequences, as described in detail in \cite{Ibarakietal2005, Ibarakietal2008}.

\subsection{Perturbation Mechanisms}
\label{sec:Perturbation}

The following three perturbation mechanisms were implemented:

\begin{itemize}
  \item ($l,l'$)-block swap intra-machine: one ($l,l'$)-block swap intra-machine move is performed at random with $l$ and $l' \in \{2,\dots,n/4\}$. 
  \item Multiple ($1,1$)-insertion inter-machine: a job $j$ from a machine $k$ is moved to a machine $k'$, while a job $j'$ from $k'$ is moved to $k$. The jobs, machines and positions to be inserted are all chosen at random and such procedure is repeated one, two or three consecutive times.
  \item Multiple ($l,l'$)-block insertion inter-machine: this perturbation generalizes the previous one but in this case blocks of jobs of sizes $l$ and $l'$, respectively, are involved in the move with $l\in \{1,2\}$ and $l' \in \{2,3\}$.
\end{itemize}

The first mechanism is only applied for single machine problems, while the second and third are only applied for parallel machine problems and they are chosen at random.


\section{Computational experiments}
\label{sec:Results}

The UILS algorithm was coded in C++ and the experiments were executed in an Intel Core i7-2600 with 3.40 GHz and 16 GB of RAM running under Ubuntu Linux 12.04. Only a single thread was used in our testing. The proposed algorithm was executed 10 times for each instance in the final experiments. Furthermore, with a view of better comparing the runtime performance between UILS and other methods from the literature that used machines with a quite inferior hardware performance than ours, we have scaled the CPU time values reported in their work to our machine. This was done by means of approximation factors that were computed using the single thread rating values reported in \url{https://www.cpubenchmark.net/}. 


Unfortunately, the vast majority of the problems listed in Table \ref{tab:Problems} does not have publicly available instances. Most authors usually generate the instances for a particular problem  themselves but they seldom make them easily accessible for the research community. Table \ref{tab:ProblemsConsidered} lists some problems in which we are aware that there are instances available online. We executed the UILS algorithm only for those with results reported in the literature.

We remark that we do not present the results obtained by our algorithm for the publicly available instances of problems $1|s_{ij}|\sum T_j$ and $1|s_{ij}|\sum w_jT_j$, because those found by a similar and simplified version of this algorithm were already reported in \cite{Subramanian2014}. UILS still finds slightly better results than the simplified algorithm, but for brevity we chose not report them here.  We also do not report the results found for the publicly available instances of single machine problems without sequence-dependent setup times, in particular, $1||\sum w_j T_j$, $1|r_j|\sum w_j T_j$, $1||\sum w'_j E_j + w_j T_j$, $1|r_j|\sum w'_j E_j + w_j T_j$, $1|r_j|\sum w_j C_j$, because they are well-solved by the highly sophisticated exact algorithms of \cite{Tanakaetal2009, TanakaFujikuma2012}. UILS even finds the optimal solutions of such instances but not as fast as the exact methods. Moreover, we are not aware of any heuristic method whose performance is comparable to the exact ones for these problems.

\begin{table}[!ht]
  \centering
  \onehalfspacing
  \scriptsize
  \caption{Selected problems with publicly available instances}
  \begin{tabular}{lccccc}
    \hline
    \multicolumn{1}{c}{Problem} & $\#$Instances & $|J|$ & $|M|$ & Related work \\
    \hline
    $P||\sum T_j$ & $2250^1$ & $20$ and $25$ & $2$ to $10$ & \cite{TanakaAraki2008}\\
    $P||\sum  w_jT_j$ &  $1125^2$ & $40$ to $100$ & $2, 4, 10,20$ and $30$ & \cite{Rodrigues2008, Pessoaetal2010} \\
    $P||\sum w'_jE_j+w_jT_j$ & $5000^3$ & $40$ to $500$ & $2, 4, 10,20$ and $30$ & \cite{Amorim2013, Amorim2013b}\\
    $R||\sum w_jT_j$ & $1440^4$ & $40$ to $200$ & $2$ to $5$ & \cite{SenBulbul2015}\\
    $R||\sum w'_jE_j+w_jT_j$ & $1440^4$ & $40$ to $200$ & $2$ to $5$ &  \cite{SenBulbul2015}\\ 
    \hline
    \multicolumn{5}{l}{$^1$ Available at: \url{https://sites.google.com/site/shunjitanaka/pmtt}}\\
    \multicolumn{5}{l}{$^2$ Available at: \url{http://algox.icomp.ufam.edu.br/index.php/benchmark-instances/weighted-tardiness-scheduling}}\\
    \multicolumn{5}{l}{$^3$ Available at: \url{http://algox.icomp.ufam.edu.br/index.php/benchmark-instances/weighted-earliness-tardiness-scheduling}}\\
    \multicolumn{5}{l}{$^4$ Available at: \url{http://people.sabanciuniv.edu/~bulbul/papers/Sen_Bulbul_Rm_TWT-TWET_source-data-results_IJOC_2015.rar}}

  \end{tabular}
%
  \label{tab:ProblemsConsidered}
\end{table}


\subsection{Parameter tuning}

In this section we explain how we tuned the main parameters of UILS, that is, $I_R$, $I_{ILS}$, and the neighborhood sets $L_{intra}$, $L'_{intra}$, $L_{inter}$ and $L'_{inter}$.

We start by describing how we selected the neighborhood sets. We used an incremental approach where we included one neighborhood at a time in ascending order of $l$ and $l'$, and then we evaluated the impact on the solution quality after such inclusion. More precisely, for every new neighborhood considered, we applied the RVND procedure over an initial solution and then we computed the value of the gap of the resulting local optimum with respect to the initial solution. 
Initially, only one neighborhood of a particular type was considered, namely 1-block insertion intra-machine, and then we kept increasing the set until no improvement was verified. In this case, we stopped adding the $l$-block insertion intra-machine neighborhoods and started including $(l,l')$-block swap intra-machine neighborhoods until no improvement was observed. Then we included the  $l$-block insertion inter-machine neighborhoods and, finally, the $(l,l')$-block swap intra-machine neighborhoods. 

We selected 24 challenging instances among those available for problem $P||\sum w'_jE_j+w_jT_j$ (without idle time) to determine the neighborhood sets. For each instance and for each combination of neighborhoods, we ran the RVND procedure 5 times over different initial solutions and we stored the average improvement obtained. Table \ref{tab:NeighborhoodsTuning} reports the average improvements obtained for each combination. Strikethrough entries correspond to those neighborhoods that have been disregarded due to lack of average improvement on the solution quality. The following sets were selected:  $L_{intra} = \{1,2\}$, $L'_{intra} = \{(1,1)\}$, $L_{inter} = \{1,2\}$ and $L'_{inter} = \{(1,1), (1,2), (1,3), (2,2), (2,3), (2,4), (3,3), (3,4), (4,4)\}$.

It is worth mentioning that for single machine problems we considered $L_{intra} = \{1,2, 3\}$ and $L'_{intra} = {(1,1)}$ as in \cite{Subramanian2014}.

\begin{table}[htbp]
\caption{Improvements found by RVND when including a new neighborhood}
\centering
\onehalfspacing
\footnotesize
\renewcommand{\tabcolsep}{0.08cm}
\begin{tabular}{cc}
\hline
\multirow{2}{*}{Neighborhoods} & Avg. Imp. \\ 
& (\%) \\ 

\hline
1-block insertion intra-machine  & 64.22 \\ 
+ 2-block insertion intra-machine  & 64.33 \\ 
+ \sout{3-block insertion intra-machine}  & 63.33 \\ 
+ (1.1)-swap intra-machine  & 65.75 \\ 
+ \sout{(1.2)-swap intra-machine}  & 65.63 \\ 
+ 1-block insertion inter-machine  & 67.50 \\ 
+ 2-block insertion inter-machine  & 68.01 \\ 
+ \sout{3-block insertion inter}  & 67.62 \\ 
+ (1.1)-swap inter-machine  & 68.04 \\ 
+ (1.2)-swap inter-machine  & 68.32 \\ 
+ (1.3)-swap inter-machine  & 69.02 \\ 
+ (2.2)-swap inter-machine  & 69.22 \\ 
+ (2.3)-swap inter-machine  & 69.47 \\ 
+ (2.4)-swap inter-machine  & 69.58 \\ 
+ (3.3)-swap inter-machine  & 69.63 \\ 
+ (3.4)-swap inter-machine  & 69.65 \\ 
+ (4.4)-swap inter-machine  & 69.79 \\ 
+ \sout{(4.5)-swap inter-machine}  & 69.63 \\ 
\hline
\end{tabular}
\label{tab:NeighborhoodsTuning}
\end{table}

In order to tune the parameter $I_{ILS}$ we performed a series of experiments for the same 24 instances previously selected. We considered different values, namely, $n/2$, $n$, $2 \times n$, $4 \times n$ and $6 \times n$. We noticed that a single start of the method, i.e. $I_R = 1$, led to inconclusive results. Hence, we decided to adopt $I_R = 10$ as in \cite{Subramanian2014} in order to better calibrate the value of $I_{ILS}$. Table \ref{tab:ParameterTuning_PWET} show the results obtained with different values of $I_{ILS}$ where the gap reported for every instance is between the average solution of 5 runs and the best known solution (BKS). We also report the average time of the 5 runs, as well as the number of times BKS was found or improved. 

\begin{table}[!ht]
  \caption{Results for different values of $I_{ILS}$ in selected instances of problem $P||\sum w'_jE_j+w_jT_j$ (without idle time)}
  \centering
  \onehalfspacing
  \footnotesize
  \renewcommand{\tabcolsep}{0.04cm}
    \begin{tabular}{ccccccccccccccccccccc}
    \hline
    \multirow{3}{*}{Instance}&  & \multicolumn{ 3}{c}{$I_{ILS}=n/2$} &  & \multicolumn{ 3}{c}{$I_{ILS}=n$} &  & \multicolumn{ 3}{c}{$I_{ILS}=2 \times n$} &  & \multicolumn{ 3}{c}{$I_{ILS}=4 \times n$} &  & \multicolumn{ 3}{c}{$I_{ILS}=6 \times n$}\\ 
    \cline{3-5}\cline{7-9}\cline{11-13}\cline{15-17}\cline{19-21}
    &  & Gap & Time & \multirow{2}{*}{\#BKS} & & Gap & Time & \multirow{2}{*}{\#BKS} & & Gap & Time & \multirow{2}{*}{\#BKS} & & Gap & Time & \multirow{2}{*}{\#BKS} & & Gap & Time & \multirow{2}{*}{\#BKS}\\
    & & (\%) & (s) &  &  & (\%) & (s) &  &  & (\%) & (s) &  &  & (\%) & (s) &  &  & (\%) & (s) & \\ 
    \hline
    wet100-10m-121  &    & 0.08 & 20.5 & 0 &    & 0.06 & 31.9 & 0 &    & 0.05 & 57.2 & 0 &    & 0.03 & 111.0 & 0 &    & 0.03 & 172.0 & 0 \\ 
    wet100-10m-31  &    & 0.78 & 29.2 & 0 &    & 0.63 & 52.7 & 0 &    & 0.47 & 96.6 & 0 &    & 0.37 & 173.5 & 0 &    & 0.29 & 264.6 & 0 \\ 
    wet100-10m-61  &    & 0.43 & 24.5 & 0 &    & 0.34 & 39.1 & 0 &    & 0.25 & 76.6 & 0 &    & 0.18 & 116.5 & 0 &    & 0.09 & 182.5 & 1 \\ 
    wet100-2m-1  &    & 0.02 & 34.2 & 0 &    & 0.01 & 64.2 & 0 &    & 0.01 & 148.9 & 0 &    & 0.00 & 262.4 & 0 &    & 0.00 & 411.1 & 1 \\ 
    wet100-2m-11  &    & 0.04 & 30.6 & 1 &    & 0.02 & 67.5 & 0 &    & 0.02 & 101.1 & 0 &    & 0.00 & 174.8 & 3 &    & 0.01 & 244.5 & 2 \\ 
    wet100-2m-111  &    & 0.08 & 34.0 & 0 &    & 0.05 & 67.6 & 0 &    & 0.00 & 136.5 & 2 &    & 0.01 & 219.3 & 2 &    & 0.00 & 307.8 & 1 \\ 
    wet100-2m-121  &    & 0.01 & 20.5 & 0 &    & 0.01 & 40.1 & 0 &    & 0.00 & 83.2 & 1 &    & 0.00 & 131.5 & 2 &    & 0.00 & 209.4 & 4 \\ 
    wet100-2m-31  &    & 0.07 & 34.0 & 0 &    & 0.02 & 64.0 & 0 &    & 0.02 & 121.4 & 0 &    & 0.00 & 227.0 & 1 &    & 0.00 & 347.4 & 2 \\ 
    wet100-2m-61  &    & 0.13 & 31.3 & 0 &    & 0.04 & 55.8 & 0 &    & 0.02 & 95.4 & 0 &    & 0.00 & 174.3 & 5 &    & 0.00 & 234.3 & 3 \\ 
    wet100-2m-71  &    & 0.00 & 17.0 & 0 &    & 0.00 & 32.7 & 1 &    & 0.00 & 58.5 & 2 &    & 0.00 & 99.7 & 2 &    & 0.00 & 140.7 & 3 \\ 
    wet100-2m-81  &    & 0.37 & 37.1 & 0 &    & 0.25 & 65.4 & 0 &    & 0.17 & 109.1 & 0 &    & 0.10 & 244.9 & 1 &    & 0.08 & 293.7 & 1 \\ 
    wet100-2m-91  &    & 0.02 & 25.4 & 1 &    & 0.01 & 47.8 & 2 &    & 0.00 & 81.4 & 4 &    & 0.00 & 125.8 & 4 &    & 0.00 & 182.0 & 5 \\ 
    wet100-4m-111  &    & 0.29 & 35.1 & 0 &    & 0.26 & 80.3 & 0 &    & 0.20 & 131.1 & 0 &    & 0.09 & 238.5 & 0 &    & 0.14 & 337.6 & 0 \\ 
    wet100-4m-31  &    & 0.44 & 33.8 & 0 &    & 0.23 & 72.6 & 0 &    & 0.26 & 134.5 & 0 &    & 0.19 & 274.1 & 0 &    & 0.17 & 374.5 & 0 \\ 
    wet100-4m-61  &    & 0.30 & 33.5 & 0 &    & 0.22 & 66.2 & 0 &    & 0.18 & 102.9 & 0 &    & 0.15 & 219.0 & 0 &    & 0.06 & 271.6 & 1 \\ 
    wet100-4m-81  &    & 2.06 & 38.6 & 0 &    & 1.89 & 75.9 & 0 &    & 1.56 & 143.8 & 0 &    & 0.91 & 248.7 & 0 &    & 0.85 & 379.6 & 0 \\ 
    wet150-2m-1  &    & 0.02 & 124.0 & 0 &    & 0.02 & 257.8 & 0 &    & 0.01 & 564.7 & 0 &    & 0.01 & 1044.1 & 0 &    & 0.00 & 1569.2 & 0 \\ 
    wet200-2m-1  &    & 0.02 & 385.9 & 0 &    & 0.02 & 825.3 & 0 &    & 0.01 & 1725.8 & 0 &    & 0.01 & 3603.8 & 0 &    & 0.01 & 5201.9 & 0 \\ 
    wet40-2m-1  &    & 0.02 & 1.4 & 0 &    & 0.00 & 2.5 & 3 &    & 0.00 & 5.8 & 5 &    & 0.00 & 9.1 & 5 &    & 0.00 & 11.8 & 5 \\ 
    wet40-4m-1  &    & 0.24 & 1.5 & 0 &    & 0.10 & 3.1 & 1 &    & 0.02 & 5.6 & 3 &    & 0.02 & 11.0 & 3 &    & 0.00 & 14.2 & 5 \\ 
    wet40-4m-111  &    & 0.01 & 1.3 & 4 &    & 0.00 & 2.8 & 5 &    & 0.00 & 4.3 & 5 &    & 0.00 & 7.6 & 5 &    & 0.00 & 11.4 & 5 \\ 
    wet40-4m-121  &    & 0.00 & 0.9 & 5 &    & 0.00 & 1.6 & 5 &    & 0.00 & 2.6 & 5 &    & 0.00 & 4.5 & 5 &    & 0.00 & 6.2 & 5 \\ 
    wet40-4m-91  &    & 0.03 & 1.1 & 4 &    & 0.00 & 1.8 & 5 &    & 0.00 & 3.4 & 5 &    & 0.00 & 6.7 & 5 &    & 0.00 & 9.3 & 5 \\ 
    wet50-2m-1  &    & 0.02 & 3.2 & 0 &    & 0.02 & 5.4 & 0 &    & 0.01 & 10.8 & 0 &    & 0.00 & 19.4 & 1 &    & 0.00 & 28.9 & 0 \\ 
    \hline
    Avg.  &    & 0.23 & 41.6 &  --  &    & 0.18 & 84.3 &  --  &    & 0.14 & 166.7 &  --  &    & 0.09 & 322.8 &  --  &    & 0.07 & 466.9 &  --  \\ 
    \hline
    Total  &    &  --  &  --  & 15 &    &  --  &  --  & 22 &    &  --  &  --  & 32 &    &  --  &  --  & 44 &    &  --  &  --  & 49 \\ 
    \hline
    \end{tabular}
\label{tab:ParameterTuning_PWET}
\end{table}

The results illustrated in Table \ref{tab:ParameterTuning_PWET} suggest that $4 \times n$ and $6 \times n$ seem to be superior in terms of solution quality than the other settings, but not very different from each other. We decided to adopt $I_{UILS} = 4 \times n$ because it seemed to provide an interesting compromise between solution quality and average CPU time. 

We performed similar experiments considering $I_R = 15$ and $I_R = 20$, but the increment in the CPU time due to more restarts do not seem to compensate the slight increase in the solution quality. Therefore, we chose to keep $I_R = 10$, but limiting the total execution time to $600$ seconds in order to avoid long runs for large instances.

While performing further experiments, we realized that the proposed algorithm performed quite slow for problems with idle time when adopting $I_{ILS} = 4 \times n$. This is because the local search phase is more expensive in terms of CPU time, as already explained. We thus performed a similar experiment as the one just described above, but for 10 selected instances of problem $R||\sum w'_jE_j+w_jT_j$. In this case we only considered the following values for $I_{ILS}$: $n/2$, $n$ and $2 \times n$. Furthermore, since UILS systematically found or improved the BKS, we decided to evaluate the gap with respect to the lower bound and also with respect to our best solution (OBS) so as to better appreciate the influence of the parameter. The results of these experiments are reported in Table \ref{tab:ParameterTuning_PWET_IT}. By following a similar criterion used in the previous case, we decided to adopt $I_{ILS} = n$ for problems with idle time.

\begin{table}[H]
\caption{Results for different values of $I_{ILS}$ in selected instances of problem $R||\sum w'_jE_j+w_jT_j$ (with idle time)}
\centering
\onehalfspacing
\footnotesize
\renewcommand{\tabcolsep}{0.03cm}
\begin{tabular}{cccccccccccccccc}
\hline
\multirow{3}{*}{Instance}&  & \multicolumn{ 4}{c}{$I_{ILS}=n/2$} &  & \multicolumn{ 4}{c}{$I_{ILS}=n$} &  & \multicolumn{ 4}{c}{$I_{ILS}=2 \times n$}\\
\cline{3-6}\cline{8-11}\cline{13-16}
&  & Gap$_{LB}$ & Gap$_{OBS}$ &  Time & \multirow{2}{*}{\#OBS} & & Gap$_{LB}$ & Gap$_{OBS}$ & Time & \multirow{2}{*}{\#OBS} & & Gap$_{LB}$ & Gap$_{OBS}$ & Time & \multirow{2}{*}{\#OBS}\\
& & (\%) & (\%) & (s) &  &  & (\%) & (\%) & (s) &  &  & (\%) & (\%) & (s) & \\ 
\hline
WET40\_R2\_25\_100\_021 &  & 1.87 & 0.01 & 18.3 & 4 &  & 1.86 & 0.00 & 31.7 & 5 &  & 1.86 & 0.00 & 59.5 & 5 \\ 
WET40\_R2\_25\_100\_101 &  & 0.00 & 0.00 & 5.8 & 5 &  & 0.00 & 0.00 & 10.2 & 5 &  & 0.00 & 0.00 & 18.7 & 5 \\ 
WET60\_R2\_25\_100\_021 &  & 1.25 & 0.07 & 131.2 & 1 &  & 1.18 & 0.00 & 245.3 & 5 &  & 1.19 & 0.02 & 391.3 & 4 \\ 
WET60\_R2\_25\_100\_101 &  & 0.08 & 0.00 & 33.3 & 5 &  & 0.08 & 0.00 & 60.4 & 5 &  & 0.08 & 0.00 & 102.9 & 5 \\ 
WET60\_R3\_25\_100\_021 &  & 0.05 & 0.04 & 102.7 & 3 &  & 0.01 & 0.00 & 204.2 & 4 &  & 0.01 & 0.00 & 363.1 & 5 \\ 
WET60\_R3\_25\_100\_101 &  & 0.01 & 0.00 & 31.6 & 5 &  & 0.01 & 0.00 & 53.5 & 5 &  & 0.01 & 0.00 & 95.6 & 5 \\ 
WET80\_R2\_25\_100\_021 &  & 1.24 & 0.13 & 547.5 & 1 &  & 1.14 & 0.03 & 951.4 & 1 &  & 1.12 & 0.01 & 1534.3 & 3 \\ 
WET80\_R2\_25\_100\_101 &  & 0.02 & 0.00 & 119.5 & 5 &  & 0.02 & 0.00 & 233.1 & 5 &  & 0.02 & 0.00 & 390.6 & 5 \\ 
WET90\_R3\_25\_100\_021 &  & 1.43 & 0.39 & 656.8 & 0 &  & 1.09 & 0.05 & 1307.7 & 1 &  & 1.09 & 0.05 & 2010.7 & 2 \\ 
WET90\_R3\_25\_100\_101 &  & 0.16 & 0.00 & 162.4 & 5 &  & 0.16 & 0.00 & 253.3 & 5 &  & 0.16 & 0.00 & 409.0 & 5 \\ 
\hline
Avg. &  & 0.61 & 0.06 & 180.9 & -- &  & 0.56 & 0.01 & 335.1 & -- &  & 0.55 & 0.01 & 537.6 & -- \\ 
\hline
Total &  & -- & -- & -- & 34 &  & -- & -- & -- & 41 &  & -- & -- & -- & 44 \\ 
\hline
\end{tabular}
\label{tab:ParameterTuning_PWET_IT}
\end{table}

\subsection{Results for problem $P||\sum T_j$}
\label{PTT}

 

The optimal solutions of all 2250 instances available for problem $P||\sum T_j$ were found by the exact method of \cite{TanakaAraki2008}. The UILS algorithm was capable of finding such optima in at least 9 of the 10 runs and the average computational time was smaller than one second, as can be observed in Table \ref{tab:PTT}. 




\begin{table}[htbp]
 \caption{Summary of the results for problem $P||\sum T_j$}
  \centering
  \onehalfspacing
  \footnotesize
  \renewcommand{\tabcolsep}{0.05cm}
	\begin{tabular}{cc>{\raggedright}p{0.15\columnwidth}cccccc}
		\hline
		\multirow{2}{*}{Instance} & \multirow{3}{*}{\#Inst} & \cite{TanakaAraki2008} & & \multicolumn{5}{c}{UILS}\\
		\cline{3-3} \cline{5-9}
		\multirow{2}{*}{group} &  & \multicolumn{1}{c}{Time$^1$} & & Gap$_{\text{best}}$ & Gap$_{\text{avg}}$ & Time$_{\text{avg.}}$ & \multirow{2}{*}{\#Best} & \multirow{2}{*}{\#Opt} \\ 
		& & \multicolumn{1}{c}{(s)} & & (\%) & (\%) & (s) & &\\
		\hline
		N20\_M2 & 125 & \multicolumn{1}{c}{0.2} &  & 0.00 & 0.00 & 0.3 & 1250 & 125 \\ 
		N20\_M3 & 125 & \multicolumn{1}{c}{0.1} &  & 0.00 & 0.00 & 0.3 & 1250 & 125 \\ 
		N20\_M4 & 125 & \multicolumn{1}{c}{0.1} &  & 0.00 & 0.00 & 0.3 & 1250 & 125 \\ 
		N20\_M5 & 125 & \multicolumn{1}{c}{$<0.1$} &  & 0.00 & 0.00 & 0.3 & 1250 & 125 \\ 
		N20\_M6 & 125 & \multicolumn{1}{c}{$<0.1$} &  & 0.00 & 0.00 & 0.3 & 1250 & 125 \\ 
		N20\_M7 & 125 & \multicolumn{1}{c}{$<0.1$} &  & 0.00 & 0.00 & 0.3 & 1250 & 125 \\ 
		N20\_M8 & 125 & \multicolumn{1}{c}{$<0.1$} &  & 0.00 & 0.00 & 0.3 & 1244 & 125 \\ 
		N20\_M9 & 125 & \multicolumn{1}{c}{$<0.1$} &  & 0.00 & 0.00 & 0.2 & 1247 & 125 \\ 
		N20\_M10 & 125 & \multicolumn{1}{c}{$<0.1$} &  & 0.00 & 0.00 & 0.2 & 1248 & 125 \\ 
		N25\_M2 & 125 & \multicolumn{1}{c}{0.5} &  & 0.00 & 0.00 & 0.6 & 1250 & 125 \\ 
		N25\_M3 & 125 & \multicolumn{1}{c}{6.3} &  & 0.00 & 0.00 & 0.6 & 1247 & 125 \\ 
		N25\_M4 & 125 & \multicolumn{1}{c}{15.5} &  & 0.00 & 0.00 & 0.6 & 1248 & 125 \\ 
		N25\_M5 & 125 & \multicolumn{1}{c}{4.7} &  & 0.00 & 0.00 & 0.6 & 1244 & 125 \\ 
		N25\_M6 & 125 & \multicolumn{1}{c}{0.1} &  & 0.00 & 0.00 & 0.6 & 1246 & 125 \\ 
		N25\_M7 & 125 & \multicolumn{1}{c}{0.1} &  & 0.00 & 0.00 & 0.5 & 1250 & 125 \\ 
		N25\_M8 & 125 & \multicolumn{1}{c}{$<0.1$} &  & 0.00 & 0.00 & 0.5 & 1248 & 125 \\ 
		N25\_M9 & 125 & \multicolumn{1}{c}{0.1} &  & 0.00 & 0.00 & 0.5 & 1243 & 125 \\ 
		N25\_M10 & 125 & \multicolumn{1}{c}{$<0.1$} &  & 0.00 & 0.00 & 0.4 & 1250 & 125 \\
		\hline
		Total & 2250 & \multicolumn{1}{c}{-} &  & - & - & - & 22465 & 2250 \\ 
		\hline
		Avg. & - & \multicolumn{1}{c}{1.5} &  & 0.00 & 0.00 & 0.4 & - & - \\ 
		\hline
	\multicolumn{8}{l}{\tiny$^1$ Intel Pentium 4 with 2.4 GHz scaled to our Intel i7 with 3.40 GHz.}\\
	\end{tabular}
 \label{tab:PTT}
\end{table}

\subsection{Results for problem $P||\sum w_jT_j$}
\label{PWT}

Tables \ref{tab:resPwt1} and \ref{tab:resPwt2} present the aggregated results found for the instances considered in \cite{Rodrigues2008} and \cite{Pessoaetal2010}, respectively. Each group contains 25 instances selected by the authors and has the format \texttt{wtX-Ym}, where \texttt{X} denotes the number of jobs and \texttt{Y} corresponds to the number of machines of the group. For example, \texttt{wt40-2m} indicates that the group is composed of instances containing 40 jobs and 2 machines. Detailed results can be found in Tables \ref{PwT40_2}-\ref{PwT100_4} in Appendix \ref{app1}. The optimal solutions of all instances of groups \texttt{wt40-2m}, \texttt{wt40-4m}, \texttt{wt50-2m} and \texttt{wt50-4m} were reported in \cite{Rodrigues2008}. The gaps shown in Table \ref{tab:resPwt1} are with respect to such optimal solutions, whereas those presented in Table \ref{tab:resPwt2} are with respect to the best known solutions. This is because some instances of groups \texttt{wt100-2m}, \texttt{wt100-4m} are still open. \cite{Pessoaetal2010} reported the best upper bounds for those cases where the optimal solutions were not found by their exact method.

\begin{table}[!htbp]
  \caption{Summary of results for problem $P||\sum w_jT_j$}
  \centering
  \onehalfspacing
  \footnotesize
  \renewcommand{\tabcolsep}{0.04cm}
  \begin{tabular}{ccccccccccccc}
    \hline
    \multicolumn{1}{c}{\multirow{3}{*}{Instance}} & \multicolumn{ 5}{c}{\cite{Rodrigues2008}} & & \multicolumn{ 6}{c}{UILS}\\ 
    \cline{2-13}
    \multicolumn{1}{c}{\multirow{3}{*}{group}} & \multicolumn{ 2}{c}{Best run} & & \multicolumn{ 2}{c}{Average} & & \multicolumn{ 3}{c}{Best run} & & \multicolumn{ 2}{c}{Average}\\
    \cline{2-3} \cline{5-6} \cline{8-10} \cline{12-13}
    & Gap & \multirow{2}{*}{\#BKS} & & Gap & Time$^1$ & & Gap & \#Equal & \multirow{2}{*}{\#Imp} & & Gap & Time \\
    \smallskip
    & (\%) & & & (\%) & (s) & & (\%) & to BKS & & & (\%) & (s) \\
    \hline
		wt40-2m & 0.00 & 24 &  & 0.87 & 6.0 &  & 0.00 & 25 & 0 &  & 0.01 & 3.1 \\ 
		wt40-4m & 0.00 & 25 &  & 3.29 & 20.0 &  & 0.00 & 25 & 0 &  & 0.00 & 4.0 \\ 
		wt50-2m & 0.00 & 25 &  & 1.85 & 14.8 &  & 0.00 & 25 & 0 &  & 0.00 & 6.5 \\ 
		wt50-4m & 0.00 & 25 &  & 3.08 & 50.6 &  & 0.00 & 25 & 0 &  & 0.00 & 8.2 \\ 
    \hline
    Total & -- & 99 & & -- & -- &  & -- & 100 & 0 &  & --& -- \\ 
    \hline
    Avg. & 0.00 & -- &  & 2.27 & 22.8 &  & 0.00 & -- & -- &  & 0.00 & 5.5 \\ 
    \hline
    \multicolumn{13}{l}{{\tiny$^1$ Average of $30 \times m \times n$ runs in an Intel Xeon 2.33 GHz scaled to our Intel i7}}\\
		\multicolumn{13}{l}{{\tiny with 3.40 GHz.}}
  \end{tabular}
  \label{tab:resPwt1}
\end{table}

\begin{table}[!htbp]
  \caption{Summary of results for problem $P||\sum w_jT_j$}
  \centering
  \onehalfspacing
  \footnotesize
  \renewcommand{\tabcolsep}{0.04cm}
  \begin{tabular}{cccccccccc}
    \hline
    \multicolumn{1}{c}{\multirow{3}{*}{Instance}} & \multicolumn{ 2}{c}{\cite{Pessoaetal2010}} & & \multicolumn{ 6}{c}{UILS}\\ 
    \cline{2-10}
    \multicolumn{1}{c}{\multirow{3}{*}{group}} & \multicolumn{ 2}{c}{Best run} & & \multicolumn{ 3}{c}{Best run} & & \multicolumn{ 2}{c}{Average}\\
    \cline{2-3} \cline{5-7} \cline{9-10}
    & Gap & \multirow{2}{*}{\#BKS} & & Gap & \#Equal & \multirow{2}{*}{\#Imp} & & Gap & Time \\
    \smallskip
    & (\%) & & & (\%) & to BKS & & & (\%) & (s) \\
    \hline
    wt100-2m & 0.00 & 21 &  &  0.00 & 22 & 2 &  & 0.00 & 71.1 \\ 
    wt100-4m & 0.00 & 16 &  &  -0.01 & 15 & 5 &  & 0.00 & 94.8 \\ 
    \hline
    Total & -- & 37 &  & -- & 37 & 7 &  &-- & -- \\ 
    \hline
    Avg. & 0.00 & -- &  & 0.00 & -- & -- &  & 0.00 & 83.0 \\ 
    \hline
  \end{tabular}
  \label{tab:resPwt2}
\end{table}


From the results obtained, it can be observed that UILS clearly outperforms the heuristic of \cite{Rodrigues2008} for the instances containing 40 and 50 jobs, respectively. High quality solutions were also obtained for 100-job instances and 7 improved solutions were found. 

\subsection{Results for problem $P||\sum w^\prime_jE_j + w_jT_j$ (without idle time)}
\label{PWET}

Table \ref{tab:resPwet1} shows a summary of the results found for problem $P||\sum w^\prime_jE_j + w_jT_j$, without idle time. Each group of instance has the format \texttt{wetX-Ym}, where \texttt{X} indicates the number of jobs and \texttt{Y} denotes the  number of machines of the group. The 2-machine and 4-machine groups have each of them 11 or 12 instances with available results for comparison, while the groups with 10 machines have 5 instances each. We compare our results with the best heuristics proposed in \cite{Amorim2013, Amorim2013b}, that is, ILS+PR, GA+LS+PR and ILS-M.  Detailed results are provided in Tables \ref{PWET40-2}-\ref{PWET100-10} in Appendix \ref{app2}. Note that for some instances our best solution has one machine less. 


For the instances of groups \texttt{wet40-2m} and \texttt{wet50-2m}, UILS found or improved all best known solutions. The same happened for group \texttt{wet100-2m}, except for a single instance where our best solution was only one unit above the BKS. For the 4-machine instances, our algorithm found or improved the BKS for all but five instances from the literature, 
Moreover, UILS improved one solution and equaled the results of the other ones for the instances of groups \texttt{wet40-10m} and \texttt{wet50-10m}. Finally, for the instances of group \texttt{wet100-10m}, UILS improved just one solution, but the average gap was only 0.24\%. As for the CPU time, it can be observed that UILS is considerably faster than the other algorithms and the machines used have comparable configurations.

\begin{table}[!htbp]
\caption{Summary of results for problem $P||\sum w^\prime_jE_j + w_jT_j$ -- without idle time}
  \centering
  \onehalfspacing
  \scriptsize
  \renewcommand{\tabcolsep}{0.04cm}
\begin{tabular}{ccccccccccccccccccccccccc}

\hline
\multirow{3}{*}{Instance} & \multicolumn{5}{c}{ILS+PR} & & \multicolumn{5}{c}{GA+LS+PR} & &  \multicolumn{5}{c}{ILS-M} & & \multicolumn{6}{c}{UILS}\\ 
\cline{2-25}
\multirow{3}{*}{Group} & \multicolumn{2}{c}{Best run} & & \multicolumn{2}{c}{Average} & & \multicolumn{2}{c}{Best run} & & \multicolumn{2}{c}{Average} & &  \multicolumn{2}{c}{Best run} & & \multicolumn{2}{c}{Average} & & \multicolumn{3}{c}{Best run} & & \multicolumn{2}{c}{Average}\\ \cline{2-3}\cline{5-6}\cline{8-9}\cline{11-12}\cline{14-15}\cline{17-18}\cline{20-22}\cline{24-25}

& Gap & \multirow{2}{*}{\#BKS} & & Gap & Time$^1$ &  & Gap & \multirow{2}{*}{\#BKS} &  & Gap & Time$^1$ &  & Gap & \multirow{2}{*}{\#BKS} &  & Gap & Time$^1$ &  & Gap & \#Equal & \multirow{2}{*}{\#Imp} &  & Gap & Time\\
& (\%) & & & (\%) & (s) &  & (\%) & & & (\%) & (s) &  & (\%) & & & (\%) & (s) &  & (\%) & to BKS & &  & (\%) & (s) \\ 
 \hline
wet40-2m & 0.00 & 12 &  & 0.00 & 15.4 &  & 0.00 & 12 &  & 0.01 & 33.6 &  & 0.00 & 12 &  & 0.00 & 13.6 &  & -0.58 & 11 & 1 &  & -0.58 & 5.6 \\ 
wet50-2m & 0.00 & 12 &  & 0.00 & 41.2 &  & 0.00 & 12 &  & 0.02 & 99.1 &  & 0.00 & 12 &  & 0.00 & 32.2 &  & -0.78 & 11 & 1 &  & -0.78 & 12.6 \\ 
wet100-2m & 0.02 & 5 &  & 0.02 & 588.0 &  & 0.00 & 10 &  & 0.01 & 3098.2 &  & 0.00 & 11 &  & 0.00 & 533.0 &  & -0.67 & 10 & 1 &  & -0.66 & 168.5 \\ 
wet40-4m & 0.74 & 11 &  & 0.74 & 68.4 &  & 0.74 & 11 &  & 0.75 & 201.1 &  & 0.74 & 11 &  & 0.74 & 37.0 &  & 0.00 & 12 & 0 &  & 0.00 & 6.3 \\ 
wet50-4m & 0.00 & 10 &  & 0.00 & 163.1 &  & 0.02 & 8 &  & 0.02 & 546.8 &  & 0.00 & 10 &  & 0.00 & 73.5 &  & -0.91 & 10 & 1 &  & -0.63 & 14.1 \\ 
wet100-4m & 0.14 & 2 &  & 0.16 & 3047.1 &  & 0.01 & 4 &  & 0.01 & 22689.9 &  & 0.01 & 8 &  & 0.03 & 1876.4 &  & 0.08 & 5 & 1 &  & 0.15 & 190.3 \\ 
wet40-10m & 0.00 & 5 &  & 0.00 & 325.6 &  & 0.00 & 5 &  & 0.00 & 4780.4 &  & 0.00 & 5 &  & 0.00 & 46.9 &  & 0.00 & 5 & 0 &  & 0.00 & 4.1 \\ 
wet50-10m & 0.03 & 4 &  & 0.03 & 852.7 &  & 0.00 & 5 &  & 0.01 & 10237.8 &  & 0.00 & 5 &  & 0.01 & 130.0 &  & -0.60 & 4 & 1 &  & -0.59 & 9.3 \\ 
wet100-10m & 1.43 & 0 &  & 1.45 & 16817.9 &  & 1.19 & 1 &  & 1.19 & 132735.9 &  & 1.19 & 0 &  & 1.23 & 8424.8 &  & -0.01 & 0 & 1 &  & 0.05 & 140.4 \\ 
\hline
Avg. & 0.26 & -- &  & 0.27 & 2435.5 &  & 0.22 & -- &  & 0.22 & 19380.3 &  & 0.22 & -- &  & 0.22 & 1240.8 &  & -0.39 & -- & -- &  & -0.34 & 61.3 \\ 
\hline
Total & -- & 61 &  & -- & -- &  & -- & 68 &  & -- & -- &  & -- & 74 &  & -- & -- &  & -- & 68 & 7 &  & -- & -- \\ 
\hline
\multicolumn{12}{l}{{\scriptsize$^1$ Average of 3 runs in an Intel i7-3770 3.40GHz with 12 GB of RAM.}}
\end{tabular}
\label{tab:resPwet1}
\end{table}

\subsection{Results for problem $R||\sum w_jT_j$}
\label{RWT}

There are 12 groups of instances for problem $R||\sum w_jT_j$, each of them containing 60 test-problems, with format \texttt{Nxx\_My}, where \texttt{xx} indicates the number of jobs and \texttt{y} the number of machines. We compare our aggregate results with those achieved by \cite{SenBulbul2015} in Table \ref{tab:resRwt}. For each instance group, we report the number of optimal solutions (\#Opt), improvements (\#Imp), and BKSs found (\#Equal to BKS), as well as the gaps with respect to the best lower bounds reported by \cite{SenBulbul2015} (Gap$_{\text{LB}}$), which were obtained either by a preemption based relaxation that was solved using  Benders' decomposition or CPLEX, or by the lower bound obtained through a time-indexed formulation that was solved using CPLEX. The upper bounds were found by applying the  single machine solver SiPS/SiPSi \citep{Tanakaetal2009, TanakaFujikuma2012} over each machine after running the Benders' decomposition. Finally, we also report the total average CPU time (Time (s)) spent by both algorithms, and the average CPU time required by UILS  to find or improve the BKS (Time to BKS (s)). Tables \ref{RwT40_2}-\ref{RwT200_5} in Appendix \ref{app3} contain detailed results for all instances considered. 


We can observe that the proposed unified heuristic was capable of finding or improving the upper bound of all open instances but one, and also to obtain all known optimal solutions. 
More precisely, UILS was found capable of improving the result of 591 open instances and to equal the result of another 128, where 64 of them are proven optimal solutions.
The total average CPU time spent by UILS ranged, on average, from approximately 3 seconds, for the 40-job instances, to roughly 561 seconds, for the 200-job instances. When compared to the gaps found by the mathematical programming based heuristic (PR+Benders+SiPS) proposed by \cite{SenBulbul2015}, it can be observed that the UILS clearly outperforms this method in terms of solution quality, but with higher total runtimes. 
However, UILS required, on average, only $3.6$ seconds to find or improve the best solution of 719 (out of 720) instances. 
\begin{table}[!htbp]
  \caption{Summary of results for problem $R||\sum w_jT_j$}
  \centering
  \onehalfspacing
  \footnotesize
  \renewcommand{\tabcolsep}{0.04cm}
  \begin{tabular}{ccccccccccccccccc}
  \hline
    \multirow{3}{*}{Instance} &\multicolumn{7}{c}{\cite{SenBulbul2015}} & & \multicolumn{8}{c}{UILS} \\
    \cline{2-8}\cline{10-17}
    \multirow{3}{*}{group} & \multicolumn{2}{c}{Best} & & \multicolumn{4}{c}{PR+Benders+SiPS} & & \multicolumn{4}{c}{Best run} & & \multicolumn{3}{c}{Average} \\ 
    \cline{2-3}\cline{5-8}\cline{10-13}\cline{15-17}
    & \multirow{2}{*}{\#Opt} & Gap$_{\text{LB}}$  & & \multirow{2}{*}{\#Opt} & \#Equal & Gap$_{\text{LB}}$ & Time$^1$ & & \multirow{2}{*}{\#Opt} & \multirow{2}{*}{\#Imp} & \#Equal & Gap$_{\text{LB}}$ & & Gap$_{\text{LB}}$ & Time to &  Time \\ 
    & & (\%) & & & to BKS & (\%) & (s) & & & & to BKS & (\%) & & (\%) & BKS (s) & (s) \\ 
    \hline
		N40\_M2 & 17 & 1.24 &  & 0 & 13 & 1.76 & 1.9 &  & 17 & 20 & 40 & 0.76 &  & 0.76 & $<$0.1 & 3.5 \\ 
		N60\_M2 & 8 & 1.20 &  & 0 & 40 & 1.46 & 6.4 &  & 8 & 41 & 19 & 0.63 &  & 0.63 & $<$0.1 & 11.9 \\ 
		N60\_M3 & 6 & 4.27 &  & 0 & 35 & 4.62 & 3.0 &  & 6 & 37 & 23 & 3.12 &  & 3.13 & 0.4 & 15.6 \\ 
		N80\_M2 & 3 & 1.43 &  & 0 & 55 & 1.52 & 13.2 &  & 3 & 55 & 5 & 0.66 &  & 0.66 & 0.3 & 29.2 \\ 
		N80\_M4 & 5 & 4.80 &  & 3 & 48 & 4.93 & 43.1 &  & 5 & 48 & 12 & 3.77 &  & 4.05 & 3.4 & 43.5 \\ 
		N90\_M3 & 1 & 3.05 &  & 1 & 56 & 3.11 & 8.0 &  & 1 & 56 & 4 & 1.18 &  & 1.23 & 2.0 &  58.5 \\ 
		N100\_M5 & 4 & 6.49 &  & 4 & 60 & 6.49$^2$ & 91.6 &  & 4 & 55 & 4 & 5.88 &  & 9.12$^3$ & 1.6 &  96.3 \\ 
		N120\_M3 & 1 & 5.94 &  & 1 & 60 & 5.94 & 19.8 &  & 1 & 58 & 2 & 3.99 &  & 4.79 & 5.8 & 149.9 \\ 
		N120\_M4 & 4 & 10.12 &  & 4 & 60 & 10.12 & 37.8 &  & 4 & 56 & 4 & 5.43 &  & 5.61 & 5.8 & 168.8 \\ 
		N150\_M5 & 5 & 3.20 &  & 5 & 60 & 3.20 & 51.5 &  & 5 & 55 & 5 & 1.99 &  & 2.09 & 8.5 & 367.6 \\ 
		N160\_M4 & 5 & 2.55 &  & 5 & 60 & 2.55 & 27.5 &  & 5 & 55 & 5 & 1.41 &  & 1.46 & 2.6 & 401.8 \\ 
		N200\_M5 & 5 & 3.13 &  & 5 & 60 & 3.13 & 88.3 &  & 5 & 55 & 5 & 1.89 &  & 2.03 & 13.2 & 560.6 \\ 
    \hline
		Total & 64 & - &  & 28 & 607 & - & - &  & 64 & 591 & 128 & - &  & - & - & - \\ 
    \hline
		Avg. & - & 3.95 &  & - & - & 4.07 & 32.67 &  & - & - & - & 2.56 & & 2.96 & 3.6 & 158.9 \\ 
		\hline
	\multicolumn{16}{l}{{\scriptsize$^1$ Single run on an Intel 3.80 GHz Core i7 920 with hyperthreading and 24 GB.}}\\
	\multicolumn{16}{l}{{\scriptsize$^2$ This value drops down to 4.48\% when instance 27 is disregarded.}}\\
	\multicolumn{16}{l}{{\scriptsize$^3$ This value drops down to 2.99\% when instance 27 is disregarded.}}
  \end{tabular}
\label{tab:resRwt}
\end{table}

\subsection{Results for problem $R||\sum w^\prime_jE_j + w_jT_j$}
\label{RWET}

We ran UILS for the same set of instances considered in Section \ref{RWT}, but now including earliness penalties. Idle times are also permitted. We performed similar comparisons and analyses as for problem $R||\sum w_jT_j$.
Aggregated results are provided in Table \ref{tab:resRwet}, while detailed results are presented in Tables \ref{RwET40_2}-\ref{RwET200_5} (see Appendix \ref{app4}).

From Table \ref{tab:resRwet}, we can observe that the proposed algorithm managed to improve the best known upper bounds of 634 instances and to equal the best results of another 84, including all known 33 optimal solutions. As expected, the CPU times are much higher than problem $R||\sum w_jT_j$, even for $I_{ILS} = n$ rather than $I_{ILS} = 4 \times n$, because of the existence of idle times, which make the local search procedure more time consuming. Moreover, UILS spent, on average, more CPU time than PR+Benders+SiPSi, but at the same time we can observe that the average gaps found by our heuristic was always smaller than those obtained by PR+Benders+SiPSi. Finally, despite the higher total runtimes, UILS required, on average, $48.8$ seconds to find or improve the best solutions of 718 instances.
\begin{table}[!htbp]
  \caption{Summary of results for problem $R||\sum w_j^\prime E_j + w_jT_j$}
  \centering
  \onehalfspacing
  \footnotesize
  \renewcommand{\tabcolsep}{0.04cm}
  \begin{tabular}{ccccccccccccccccc}
  \hline
    \multirow{3}{*}{Instance} &\multicolumn{7}{c}{\cite{SenBulbul2015}} & & \multicolumn{8}{c}{UILS} \\
    \cline{2-8}\cline{10-17}
    \multirow{3}{*}{group} & \multicolumn{2}{c}{Best} & & \multicolumn{4}{c}{PR+Benders+SiPSi} & & \multicolumn{4}{c}{Best run} & & \multicolumn{3}{c}{Average} \\ 
    \cline{2-3}\cline{5-8}\cline{10-13}\cline{15-17}
    & \multirow{2}{*}{\#Opt} & Gap$_{\text{LB}}$  & & \multirow{2}{*}{\#Opt} & \#Equal & Gap$_{\text{LB}}$ & Time$^1$ & & \multirow{2}{*}{\#Opt} & \multirow{2}{*}{\#Imp} & \#Equal & Gap$_{\text{LB}}$ & & Gap$_{\text{LB}}$ & Time to & Time \\ 
    & & (\%) & & & to BKS & (\%) & (s) & & & & to BKS & (\%) & & (\%) & BKS (s) & (s) \\ 
    \hline
		N40\_M2 & 22 & 0.16 &  & 1 & 1 & 0.96 & 52.9 &  & 22 & 13 & 47 & 0.13 &  & 0.13 & 2.6 & 22.1 \\ 
		N60\_M2 & 5 & 0.89 &  & 0 & 42 & 0.98 & 109.7 &  & 5 & 50 & 10 & 0.43 &  & 0.46 & 8.7 & 134.4 \\ 
		N60\_M3 & 4 & 0.82 &  & 0 & 20 & 1.58 & 120.2 &  & 4 & 37 & 23 & 0.39 &  & 0.44 & 18.4 & 109.3 \\ 
		N80\_M2 & 2 & 0.90 &  & 0 & 56 & 0.92 & 134.1 &  & 2 & 57 & 3 & 0.36 &  & 0.39 & 16.9 & 437.2 \\ 
		N80\_M4 & 0 & 4.54 &  & 0 & 58 & 4.57 & 228.3 &  & 0 & 60 & 0 & 2.15 &  & 2.30 & 7.9 & 317.5 \\ 
		N90\_M3 & 0 & 2.52 &  & 0 & 58 & 2.55 & 153.1 &  & 0 & 59 & 1 & 1.27 &  & 1.40 & 29.8 & 462.9 \\ 
		N100\_M5 & 0 & 8.83 &  & 0 & 60 & 8.83 & 297.7 &  & 0 & 60 & 0 & 6.03 &  & 6.33 & 17.0 & 519.2 \\ 
		N120\_M3 & 0 & 4.12 &  & 0 & 60 & 4.12 & 165.9 &  & 0 & 60 & 0 & 3.21 &  & 3.39 & 72.0 & 595.4 \\ 
		N120\_M4 & 0 & 6.98 &  & 0 & 60 & 6.98 & 217.1 &  & 0 & 60 & 0 & 5.20 &  & 5.50 & 45.2 & 593.9 \\ 
		N150\_M5 & 0 & 13.90 &  & 0 & 60 & 13.90 & 279.4 &  & 0 & 59 & 0 & 11.73 &  & 12.28 & 75.1 & 600.0 \\ 
		N160\_M4 & 0 & 8.60 &  & 0 & 60 & 8.60 & 202.5 &  & 0 & 60 & 0 & 7.44 &  & 7.87 & 131.2 & 600.0 \\ 
		N200\_M5 & 0 & 11.70 &  & 0 & 60 & 11.70 & 257.0 &  & 0 & 59 & 0 & 10.16 &  & 10.74 & 160.5 & 600.0 \\ 
    \hline
		Total & 33 & - &  & 1 & 595 & - & - &  & 33 & 634 & 84 & - &  & - & - & - \\ 
    \hline
		Avg. & - & 5.33 &  & - & - & 5.47 & 184.83 &  & - & - & - & 4.04 &  & 4.27 & 48.8 & 416.0 \\ 
		\hline
	\multicolumn{16}{l}{{\scriptsize$^1$ Single run on an Intel 3.80 GHz Core i7 920 with hyperthreading and 24 GB.}}
  \end{tabular}
\label{tab:resRwet}
\end{table}

\subsection{Impact of the developed move evaluation scheme on parallel machine problems without both idle times and sequence-dependent setup times}
\label{subsec:Speedup}

In this section we are interested in evaluating the performance of the move evaluation scheme that was developed for parallel machine problems that do not consider idle times as well as sequence-dependent setup times. We carried out some experiments in order to compare the effect of adopting the move evaluation approach that is performed in amortized $\mathcal{O}(\max\{l,l'\})$ time, as described in Section \ref{sec:MoveEval2}, rather than in $\mathcal{O}(n)$ time, as explained in Section \ref{sec:MoveEval1}. We consider the instances involving up to 200 jobs of problems  $P||\sum w_j T_j$ and 
$P||\sum w^\prime_jE_j + w_j T_j$. We executed a single iteration ($I_R = 1$) of UILS 5 times for each instance and Tables \ref{tab:speedup_wt} and \ref{tab:speedup_wet} report, for each of group of instances, the average CPU time spent by the algorithm  when using the traditional way of evaluating a move (UILS$_\text{Trad}$) and when using the enhanced way (UILS$_{\text{Fast}}$). In addition, we provide the speedup achieved by UILS$_\text{Fast}$, which is given by the ratio between the average CPU time spent by UILS$_\text{Trad}$ and the one spent by UILS$_{\text{Fast}}$.

\begin{table}[!ht]
    \centering
    \onehalfspacing
    \caption{CPU time spent for problem $P||\sum w_j T_j$ considering a single iteration of UILS$_\text{Trad}$ and UILS$_\text{Fast}$}
    \footnotesize
    \renewcommand{\tabcolsep}{0.06cm}
    \begin{tabular}{cccc}
      \hline
      Instance & {UILS\_Trad} & {UILS\_Fast} & \multirow{2}{*}{Speedup} \\ 
      group & $t_{\text{Avg.}}$(s) & $t_{\text{Avg.}}$(s) & \\ 
      \hline
      wt40-2m & 0.48 & 0.39 & \textbf{1.23} \\ 
      wt40-4m & 0.48 & 0.48 & 1.00 \\ 
      wt40-10m & 0.17 & 0.35 & 0.48 \\ 
      wt50-2m & 0.99 & 0.68 & \textbf{1.45} \\ 
      wt50-4m & 1.03 & 0.84 & \textbf{1.23} \\ 
      wt50-10m & 0.47 & 0.78 & 0.60 \\ 
      wt100-2m & 19.71 & 7.78 & \textbf{2.53} \\ 
      wt100-4m & 18.73 & 9.20 & \textbf{2.04} \\ 
      wt100-10m & 9.20 & 7.62 & \textbf{1.21} \\ 
      wt200-2m & 453.69 & 113.83 & \textbf{3.98} \\ 
      wt200-4m & 450.08 & 142.68 & \textbf{3.15} \\ 
      wt200-10m & 227.54 & 116.49 & \textbf{1.95} \\ 
      \hline
    \end{tabular}
    \label{tab:speedup_wt}
\end{table}

\begin{table}[!ht]
    \centering
    \onehalfspacing
    \caption{CPU time spent for problem $P||\sum w^\prime_jE_j + w_j T_j$  considering a single iteration of UILS$_\text{Trad}$ and UILS$_\text{Fast}$}
    \footnotesize
    \renewcommand{\tabcolsep}{0.06cm}
    \begin{tabular}{cccc}
      \hline
      Instance & {UILS\_Trad} & {UILS\_Fast} & \multirow{2}{*}{Speedup} \\ 
      group & $t_{\text{Avg.}}$(s) & $t_{\text{Avg.}}$(s) & \\ 
      \hline
      wet40-2m & 0.57 & 0.48 & \textbf{1.19} \\ 
      wet40-4m & 0.56 & 0.57 & 0.98 \\ 
      wet40-10m & 0.18 & 0.38 & 0.47 \\ 
      wet50-2m & 1.72 & 1.22 & \textbf{1.41} \\ 
      wet50-4m & 1.71 & 1.41 & \textbf{1.21} \\ 
      wet50-10m & 0.45 & 0.76 & 0.59 \\ 
      wet100-2m & 35.37 & 15.55 & \textbf{2.27} \\ 
      wet100-4m & 36.02 & 18.24 & \textbf{1.97} \\ 
      wet100-10m & 13.75 & 12.27 & \textbf{1.12} \\ 
      wet200-2m & 1047.07 & 260.40 & \textbf{4.02} \\ 
      wet200-4m & 842.24 & 267.81 & \textbf{3.14} \\ 
      wet200-10m & 363.45 & 188.43 & \textbf{1.93} \\  
      \hline
    \end{tabular}
    \label{tab:speedup_wet}
\end{table}

The results presented in Tables \ref{tab:speedup_wt} and \ref{tab:speedup_wet} demonstrate the benefits of using the enhanced move evaluation scheme, especially for large size instances containing a relative small number of machines, thus showing that the overhead of preprocessing the required data structures makes it worth to perform the move evaluation in amortized $\mathcal{O}(\max\{l,l'\})$, rather than in $O(n)$. In some cases, UILS$_\text{Fast}$ is around 4 times faster than UILS$_\text{Trad}$. However, the gains are less visible when the instances have a relative large number of machines. In fact, UILS$_\text{Fast}$ even performs slower than UILS$_\text{Trad}$ when the average number of jobs per machine is very small, meaning that the overhead of preprocessing the data structures necessary for evaluating a move in amortized $\mathcal{O}(\max\{l,l'\})$ steps for such instances does not always pay off.

\section{Concluding Remarks}
\label{sec:Conclusions}

This paper proposed a unified Iterated Local Search based heuristic algorithm, denoted as UILS, for a large class of earliness-tardiness scheduling problems. There is a huge number of scheduling problems in  the literature and they are usually solved by specific resolution methods. Because each problem has its own particularities, the development of a more general approach becomes a highly challenging task. This also partially explains the lack of unified algorithms for scheduling problems, especially when it comes to heuristics, which often have to be considerably modified to efficiently cope with different characteristics such as the inclusion of sequence-dependent setup and/or idle times. Therefore, we strongly believe that the introduction of an efficient unified heuristic approach capable of solving a number of scheduling problems can be seen as a relevant and important contribution. 

Due to the large number of works related to the variants considered in the present paper, performing a complete literature review can be very tough. Hence, another contribution of this work was to present an annotated bibliography containing more than 130 related works published in the last 25 years.

In addition to dealing with a significant variety of problems, the proposed unified algorithm also takes into account the particularities of each problem by efficiently performing the move evaluation during the local search. UILS automatically detects the most efficient method to be used according to some characteristics of the problem such as the existence of  sequence-dependent setup times, release dates, idle times and so on. The efficient move evaluation schemes used in this work were mostly based on those presented in \cite{Ibarakietal2005}, \cite{Ibarakietal2008}, \cite{ErgunOrlin2006} and \cite{Liaoetal2012}. While we directly applied the method of the first two works for problems with idle times, we had to generalize the ideas of the latter two in order to deal with parallel machines problems without idle times. This generalization was also a crucial contribution of the paper, since it improved the performance of UILS in terms of CPU time in some problems, more especially, those without sequence-dependent setup times. 

UILS was tested in hundreds of instances from the literature and the results obtained were compared with the best ones available. The computational experiments revealed that the unified method was capable of finding high quality solutions in competitive CPU times.


Despite the remarkable results found by the unified algorithm for the variants with available test-problems and results, the lack of existing benchmark instances prevented UILS to have its robustness better evaluated. Future avenues of research include the generation of new set of instances for the problems where it was not possible to test the proposed algorithm, as well as the development of a simple and general exact method capable of tackling the variants considered in this work, so as to provide lower bounds for the instances to be created. 


\section*{Acknowledgments}

We would like to thank Dr. Halil {\c S}en and Dr. Kerem B{\"u}lb{\"u}l for the valuable comments and suggestions. This work was partially supported by the Brazilian research agency CNPq, grants 305223/2015-1 and 131192/2013-2.

 \bibliographystyle{mmsbib}
\bibliography{ref}










\clearpage
\begin{appendices}

\section{Detailed results}
\label{app}

\subsection{$P||\sum w_jT_j$}
\label{app1}

Detailed results obtained for the instances of \cite{Rodrigues2008} (see Tables \ref{PwT40_2}-\ref{PwT100_4}).


}
\end{center}

\subsection{$P||\sum w^\prime_jE_j + w_jT_j$}
\label{app2}

Detailed results obtained for the instances of \cite{Amorim2013b}. Optimal solutions are highlighted in boldface (see Tables \ref{PWET40-2}-\ref{PWET100-10}).

\begin{table}[!htbp]
\caption{Results for $P||\sum w^\prime_jE_j + w_jT_j$ with $40$ jobs and $2$ machines}
\centering
\onehalfspacing
\scriptsize
\renewcommand{\tabcolsep}{0.04cm}
\begin{tabular}{cccccccccccccccccccccccccc}
\hline
\multirow{4}{*}{Inst} & \multirow{4}{*}{BKS }&  & \multicolumn{5}{c}{ILS+PR} &  & \multicolumn{5}{c}{GA+LS+PR} &  & \multicolumn{5}{c}{ILS-M} &  & \multicolumn{5}{c}{UILS} \\
\cline{3-26}
& &  & \multicolumn{1}{c}{Best} &  & \multicolumn{3}{c}{Average} &  & \multicolumn{1}{c}{Best} &  & \multicolumn{3}{c}{Average} &  & \multicolumn{1}{c}{Best} &  & \multicolumn{3}{c}{Average} &  & \multicolumn{1}{c}{Best} &  & \multicolumn{3}{c}{Average} \\
\cline{4-4}\cline{6-8}\cline{10-10}\cline{12-14}\cline{16-16}\cline{18-20}\cline{22-22}\cline{24-26}
& & & \multirow{2}{*}{Cost} &  & \multirow{2}{*}{Cost} & Gap & Time &  & \multirow{2}{*}{Cost} & & \multirow{2}{*}{Cost} & Gap & Time & & \multirow{2}{*}{Cost} & & \multirow{2}{*}{Cost} & Gap & Time & & \multirow{2}{*}{Cost} & & \multirow{2}{*}{Cost} & Gap & Time \\
 & & & & & & (\%) & (s) & & & & & (\%) & (s) & & & & & (\%) & (s) & & & & & (\%) & (s) \\
\hline
1 & \textbf{26063} &  & 26063 &  & 26063.0 & 0.00 & 15.4 &    & 26063 &  & 26064.0 & 0.00 & 33.3 &    & 26063 &  & 26063.0 & 0.00 & 13.4 &    & 26063 &  & 26063.1 & 0.00 & 8.8 \\ 
11 & \textbf{15451} &  & 15451 &  & 15451.0 & 0.00 & 13.9 &    & 15451 &  & 15451.0 & 0.00 & 31.5 &    & 15451 &  & 15451.0 & 0.00 & 13.3 &    & 15451 &  & 15451.0 & 0.00 & 5.4 \\ 
21 & \textbf{41054} &  & 41054 &  & 41054.0 & 0.00 & 13.1 &    & 41054 &  & 41054.0 & 0.00 & 33.6 &    & 41054 &  & 41054.0 & 0.00 & 13.2 &    & 41054 &  & 41054.0 & 0.00 & 2.4 \\ 
31 & \textbf{11679} &  & 11679 &  & 11679.0 & 0.00 & 14.9 &    & 11679 &  & 11679.0 & 0.00 & 32.8 &    & 11679 &  & 11679.0 & 0.00 & 13.1 &    & 11679 &  & 11679.0 & 0.00 & 6.3 \\
\hline
\end{tabular}
\label{PWET40-2}
\end{table}

\begin{table}[!htbp]
\caption{Results for $P||\sum w^\prime_jE_j + w_jT_j$ with $50$ jobs and $2$ machines}
\centering
\onehalfspacing
\scriptsize
\renewcommand{\tabcolsep}{0.0375cm}
\begin{tabular}{cccccccccccccccccccccccccc}
\hline
\multirow{4}{*}{Inst} & \multirow{4}{*}{BKS }&  & \multicolumn{5}{c}{ILS+PR} &  & \multicolumn{5}{c}{GA+LS+PR} &  & \multicolumn{5}{c}{ILS-M} &  & \multicolumn{5}{c}{UILS} \\
\cline{3-26}
& &  & \multicolumn{1}{c}{Best} &  & \multicolumn{3}{c}{Average} &  & \multicolumn{1}{c}{Best} &  & \multicolumn{3}{c}{Average} &  & \multicolumn{1}{c}{Best} &  & \multicolumn{3}{c}{Average} &  & \multicolumn{1}{c}{Best} &  & \multicolumn{3}{c}{Average} \\
\cline{4-4}\cline{6-8}\cline{10-10}\cline{12-14}\cline{16-16}\cline{18-20}\cline{22-22}\cline{24-26}
& & & \multirow{2}{*}{Cost} &  & \multirow{2}{*}{Cost} & Gap & Time &  & \multirow{2}{*}{Cost} & & \multirow{2}{*}{Cost} & Gap & Time & & \multirow{2}{*}{Cost} & & \multirow{2}{*}{Cost} & Gap & Time & & \multirow{2}{*}{Cost} & & \multirow{2}{*}{Cost} & Gap & Time \\
 & & & & & & (\%) & (s) & & & & & (\%) & (s) & & & & & (\%) & (s) & & & & & (\%) & (s) \\
\hline
1 & 62985 &  & 62985 &  & 62985.0 & 0.00 & 37.0 &    & 62985 &  & 62989.7 & 0.01 & 94.3 &    & 62985 &  & 62985.3 & 0.00 & 32.7 &    & 62985 &  & 62988.0 & 0.00 & 19.1 \\ 
11 & \textbf{27871} &  & 27871 &  & 27871.0 & 0.00 & 36.6 &    & 27871 &  & 27871.0 & 0.00 & 93.3 &    & 27871 &  & 27871.0 & 0.00 & 30.4 &    & 27871 &  & 27871.0 & 0.00 & 11.6 \\ 
21 & 111069 &  & 111069 &  & 111069.0 & 0.00 & 30.2 &    & 111069 &  & 111069.0 & 0.00 & 87.0 &    & 111069 &  & 111069.0 & 0.00 & 27.2 &    & 111069 &  & 111069.0 & 0.00 & 5.4 \\ 
31 & \textbf{18266} &  & 18266 &  & 18266.0 & 0.00 & 57.1 &    & 18266 &  & 18266.0 & 0.00 & 99.9 &    & 18266 &  & 18266.0 & 0.00 & 40.2 &    & 18266 &  & 18266.0 & 0.00 & 15.4 \\
\hline
\end{tabular}
\label{PWET50-2}
\end{table}

\begin{table}[!htbp]
\caption{Results for $P||\sum w^\prime_jE_j + w_jT_j$ with $100$ jobs and $2$ machines}
\centering
\onehalfspacing
\scriptsize
\renewcommand{\tabcolsep}{0.0375cm}
\begin{tabular}{cccccccccccccccccccccccccc}
\hline
\multirow{4}{*}{Inst} & \multirow{4}{*}{BKS }&  & \multicolumn{5}{c}{ILS+PR} &  & \multicolumn{5}{c}{GA+LS+PR} &  & \multicolumn{5}{c}{ILS-M} &  & \multicolumn{5}{c}{UILS} \\
\cline{3-26}
& &  & \multicolumn{1}{c}{Best} &  & \multicolumn{3}{c}{Average} &  & \multicolumn{1}{c}{Best} &  & \multicolumn{3}{c}{Average} &  & \multicolumn{1}{c}{Best} &  & \multicolumn{3}{c}{Average} &  & \multicolumn{1}{c}{Best} &  & \multicolumn{3}{c}{Average} \\
\cline{4-4}\cline{6-8}\cline{10-10}\cline{12-14}\cline{16-16}\cline{18-20}\cline{22-22}\cline{24-26}
& & & \multirow{2}{*}{Cost} &  & \multirow{2}{*}{Cost} & Gap & Time &  & \multirow{2}{*}{Cost} & & \multirow{2}{*}{Cost} & Gap & Time & & \multirow{2}{*}{Cost} & & \multirow{2}{*}{Cost} & Gap & Time & & \multirow{2}{*}{Cost} & & \multirow{2}{*}{Cost} & Gap & Time \\
 & & & & & & (\%) & (s) & & & & & (\%) & (s) & & & & & (\%) & (s) & & & & & (\%) & (s) \\
\hline
1 & 198281 &  & 198291 &  & 198292.3 & 0.01 & 575.7 &    & 198282 &  & 198302.7 & 0.01 & 2959.7 &    & 198281 &  & 198281.7 & 0.00 & 658.1 &    & 198282 &  & 198287.4 & 0.00 & 252.3 \\ 
11 & 127666 &  & 127676 &  & 127678.3 & 0.01 & 562.1 &    & 127666 &  & 127685.0 & 0.01 & 2957.8 &    & 127666 &  & 127666.0 & 0.00 & 464.9 &    & 127666 &  & 127671.3 & 0.00 & 171.6 \\ 
21 & 457865 &  & 457865 &  & 457865.0 & 0.00 & 417.9 &    & 457865 &  & 457865.0 & 0.00 & 2592.1 &    & 457865 &  & 457865.0 & 0.00 & 397.8 &    & 457865 &  & 457865.0 & 0.00 & 59.6 \\ 
31 & 95038 &  & 95051 &  & 95051.0 & 0.01 & 632.0 &    & 95038 &  & 95076.3 & 0.04 & 3811.4 &    & 95038 &  & 95038.0 & 0.00 & 636.6 &    & 95038 &  & 95049.3 & 0.01 & 218.0 \\
\hline
\end{tabular}
\label{PWET100-2}
\end{table}

\begin{table}[!htbp]
\caption{Results for $P||\sum w^\prime_jE_j + w_jT_j$ with $40$ jobs and $4$ machines}
\centering
\onehalfspacing
\scriptsize
\renewcommand{\tabcolsep}{0.04cm}
\begin{tabular}{cccccccccccccccccccccccccc}
\hline
\multirow{4}{*}{Inst} & \multirow{4}{*}{BKS }&  & \multicolumn{5}{c}{ILS+PR} &  & \multicolumn{5}{c}{GA+LS+PR} &  & \multicolumn{5}{c}{ILS-M} &  & \multicolumn{5}{c}{UILS} \\
\cline{3-26}
& &  & \multicolumn{1}{c}{Best} &  & \multicolumn{3}{c}{Average} &  & \multicolumn{1}{c}{Best} &  & \multicolumn{3}{c}{Average} &  & \multicolumn{1}{c}{Best} &  & \multicolumn{3}{c}{Average} &  & \multicolumn{1}{c}{Best} &  & \multicolumn{3}{c}{Average} \\
\cline{4-4}\cline{6-8}\cline{10-10}\cline{12-14}\cline{16-16}\cline{18-20}\cline{22-22}\cline{24-26}
& & & \multirow{2}{*}{Cost} &  & \multirow{2}{*}{Cost} & Gap & Time &  & \multirow{2}{*}{Cost} & & \multirow{2}{*}{Cost} & Gap & Time & & \multirow{2}{*}{Cost} & & \multirow{2}{*}{Cost} & Gap & Time & & \multirow{2}{*}{Cost} & & \multirow{2}{*}{Cost} & Gap & Time \\
 & & & & & & (\%) & (s) & & & & & (\%) & (s) & & & & & (\%) & (s) & & & & & (\%) & (s) \\
\hline
1 & \textbf{11985} &  & 11985 &  & 11985.0 & 0.00 & 97.4 &    & 11985 &  & 11986.7 & 0.01 & 187.6 &    & 11985 &  & 11986.7 & 0.01 & 63.7 &    & 11985 &  & 11986.3 & 0.01 & 9.8 \\ 
11 & \textbf{8206} &  & 8206 &  & 8206.0 & 0.00 & 61.3 &    & 8206 &  & 8206.0 & 0.00 & 212.3 &    & 8206 &  & 8206.0 & 0.00 & 40.1 &    & 8206 &  & 8206.0 & 0.00 & 4.8 \\ 
21 & \textbf{22793} &  & 22793 &  & 22793.0 & 0.00 & 62.5 &    & 22793 &  & 22793.0 & 0.00 & 183.8 &    & 22793 &  & 22793.0 & 0.00 & 42.9 &    & 22793 &  & 22793.0 & 0.00 & 3.5 \\ 
31 & \textbf{5950} &  & 5950 &  & 5950.0 & 0.00 & 64.0 &    & 5950 &  & 5950.0 & 0.00 & 233.4 &    & 5950 &  & 5950.0 & 0.00 & 42.3 &    & 5950 &  & 5950.0 & 0.00 & 6.5 \\
\hline
\end{tabular}
\label{PWET40-4}
\end{table}

\begin{table}[!htbp]
\caption{Results for $P||\sum w^\prime_jE_j + w_jT_j$ with $50$ jobs and $4$ machines}
\centering
\onehalfspacing
\scriptsize
\renewcommand{\tabcolsep}{0.04cm}
\begin{tabular}{cccccccccccccccccccccccccc}
\hline
\multirow{4}{*}{Inst} & \multirow{4}{*}{BKS }&  & \multicolumn{5}{c}{ILS+PR} &  & \multicolumn{5}{c}{GA+LS+PR} &  & \multicolumn{5}{c}{ILS-M} &  & \multicolumn{5}{c}{UILS} \\
\cline{3-26}
& &  & \multicolumn{1}{c}{Best} &  & \multicolumn{3}{c}{Average} &  & \multicolumn{1}{c}{Best} &  & \multicolumn{3}{c}{Average} &  & \multicolumn{1}{c}{Best} &  & \multicolumn{3}{c}{Average} &  & \multicolumn{1}{c}{Best} &  & \multicolumn{3}{c}{Average} \\
\cline{4-4}\cline{6-8}\cline{10-10}\cline{12-14}\cline{16-16}\cline{18-20}\cline{22-22}\cline{24-26}
& & & \multirow{2}{*}{Cost} &  & \multirow{2}{*}{Cost} & Gap & Time &  & \multirow{2}{*}{Cost} & & \multirow{2}{*}{Cost} & Gap & Time & & \multirow{2}{*}{Cost} & & \multirow{2}{*}{Cost} & Gap & Time & & \multirow{2}{*}{Cost} & & \multirow{2}{*}{Cost} & Gap & Time \\
 & & & & & & (\%) & (s) & & & & & (\%) & (s) & & & & & (\%) & (s) & & & & & (\%) & (s) \\
\hline
1  & 29229 &  & 29241 &  & 29243.7 & 0.05 & 170.4 &    & 29229 &  & 29230.7 & 0.01 & 548.8 &    & 29229 &  & 29230.0 & 0.00 & 81.4 &    & 26311 &  & 27190.7 & -6.97 & 23.1 \\ 
11 & \textbf{14828} &  & 14828 &  & 14828.0 & 0.00 & 151.3 &    & 14828 &  & 14828.0 & 0.00 & 474.1 &    & 14828 &  & 14828.0 & 0.00 & 67.5 &    & 14828 &  & 14828.0 & 0.00 & 13.9 \\ 
21 & \textbf{59441} &  & 59441 &  & 59441.0 & 0.00 & 145.1 &    & 59441 &  & 59441.0 & 0.00 & 419.5 &    & 59441 &  & 59441.0 & 0.00 & 62.0 &    & 59441 &  & 59441.0 & 0.00 & 7.6 \\ 
31 & \textbf{8709} &  & 8709 &  & 8709.0 & 0.00 & 179.1 &    & 8709 &  & 8709.0 & 0.00 & 506.8 &    & 8709 &  & 8709.0 & 0.00 & 76.7 &    & 8709 &  & 8709.0 & 0.00 & 16.1 \\
\hline
\end{tabular}
\label{PWET50-4}
\end{table}

\begin{table}[!htbp]
\caption{Results for $P||\sum w^\prime_jE_j + w_jT_j$ with $100$ jobs and $4$ machines}
\centering
\onehalfspacing
\scriptsize
\renewcommand{\tabcolsep}{0.0285cm}
\begin{tabular}{cccccccccccccccccccccccccc}
\hline
\multirow{4}{*}{Inst} & \multirow{4}{*}{BKS }&  & \multicolumn{5}{c}{ILS+PR} &  & \multicolumn{5}{c}{GA+LS+PR} &  & \multicolumn{5}{c}{ILS-M} &  & \multicolumn{5}{c}{UILS} \\
\cline{3-26}
& &  & \multicolumn{1}{c}{Best} &  & \multicolumn{3}{c}{Average} &  & \multicolumn{1}{c}{Best} &  & \multicolumn{3}{c}{Average} &  & \multicolumn{1}{c}{Best} &  & \multicolumn{3}{c}{Average} &  & \multicolumn{1}{c}{Best} &  & \multicolumn{3}{c}{Average} \\
\cline{4-4}\cline{6-8}\cline{10-10}\cline{12-14}\cline{16-16}\cline{18-20}\cline{22-22}\cline{24-26}
& & & \multirow{2}{*}{Cost} &  & \multirow{2}{*}{Cost} & Gap & Time &  & \multirow{2}{*}{Cost} & & \multirow{2}{*}{Cost} & Gap & Time & & \multirow{2}{*}{Cost} & & \multirow{2}{*}{Cost} & Gap & Time & & \multirow{2}{*}{Cost} & & \multirow{2}{*}{Cost} & Gap & Time \\
 & & & & & & (\%) & (s) & & & & & (\%) & (s) & & & & & (\%) & (s) & & & & & (\%) & (s) \\
\hline
1 & 94951 &  & 95027 &  & 95027.0 & 0.08 & 3407.5 &    & 94963 &  & 94969.7 & 0.02 & 20336.3 &    & 94951 &  & 94961.0 & 0.01 & 4685.8 &    & 94986 &  & 95002.3 & 0.05 & 283.3 \\ 
11 & 65719 &  & 65756 &  & 65771.3 & 0.08 & 2956.3 &    & 65719 &  & 65719.0 & 0.00 & 16975.2 &    & 65719 &  & 65719.0 & 0.00 & 1282.1 &    & 65719 &  & 65743.3 & 0.04 & 202.9 \\ 
21 & 237398 &  & 237398 &  & 237398.3 & 0.00 & 2296.5 &    & 237399 &  & 237399.0 & 0.00 & 19024.0 &    & 237399 &  & 237399.0 & 0.00 & 1115.4 &    & 237398 &  & 237398.8 & 0.00 & 91.7 \\ 
31 & 45644 &  & 45738 &  & 45738.0 & 0.21 & 3396.2 &    & 45656 &  & 45656.0 & 0.03 & 19761.5 &    & 45644 &  & 45692.3 & 0.11 & 1974.6 &    & 45664 &  & 45720.2 & 0.17 & 270.4 \\
\hline
\end{tabular}
\label{PWET100-4}
\end{table}

\begin{table}[!htbp]
\caption{Results for $P||\sum w^\prime_jE_j + w_jT_j$ with $40$ jobs and $10$ machines}
\centering
\onehalfspacing
\scriptsize
\renewcommand{\tabcolsep}{0.04cm}
\begin{tabular}{cccccccccccccccccccccccccc}
\hline
\multirow{4}{*}{Inst} & \multirow{4}{*}{BKS }&  & \multicolumn{5}{c}{ILS+PR} &  & \multicolumn{5}{c}{GA+LS+PR} &  & \multicolumn{5}{c}{ILS-M} &  & \multicolumn{5}{c}{UILS} \\
\cline{3-26}
& &  & \multicolumn{1}{c}{Best} &  & \multicolumn{3}{c}{Average} &  & \multicolumn{1}{c}{Best} &  & \multicolumn{3}{c}{Average} &  & \multicolumn{1}{c}{Best} &  & \multicolumn{3}{c}{Average} &  & \multicolumn{1}{c}{Best} &  & \multicolumn{3}{c}{Average} \\
\cline{4-4}\cline{6-8}\cline{10-10}\cline{12-14}\cline{16-16}\cline{18-20}\cline{22-22}\cline{24-26}
& & & \multirow{2}{*}{Cost} &  & \multirow{2}{*}{Cost} & Gap & Time &  & \multirow{2}{*}{Cost} & & \multirow{2}{*}{Cost} & Gap & Time & & \multirow{2}{*}{Cost} & & \multirow{2}{*}{Cost} & Gap & Time & & \multirow{2}{*}{Cost} & & \multirow{2}{*}{Cost} & Gap & Time \\
 & & & & & & (\%) & (s) & & & & & (\%) & (s) & & & & & (\%) & (s) & & & & & (\%) & (s) \\
\hline
1 & \textbf{3988} &  & 3988 &  & 3988.0 & 0.00 & 333.0 &    & 3988 &  & 3988.0 & 0.00 & 4229.0 &    & 3988 &  & 3988.0 & 0.00 & 47.5 &    & 3988 &  & 3988.0 & 0.00 & 4.4 \\ 
31 & \textbf{3558} &  & 3558 &  & 3558.0 & 0.00 & 337.4 &    & 3558 &  & 3558.0 & 0.00 & 5082.1 &    & 3558 &  & 3558.0 & 0.00 & 47.3 &    & 3558 &  & 3558.0 & 0.00 & 4.4 \\ 
61 & \textbf{5985} &  & 5985 &  & 5985.0 & 0.00 & 325.1 &    & 5985 &  & 5985.0 & 0.00 & 4749.9 &    & 5985 &  & 5985.0 & 0.00 & 47.5 &    & 5985 &  & 5985.0 & 0.00 & 4.0 \\ 
91 & \textbf{9818} &  & 9818 &  & 9818.0 & 0.00 & 320.4 &    & 9818 &  & 9818.0 & 0.00 & 4849.7 &    & 9818 &  & 9818.0 & 0.00 & 47.5 &    & 9818 &  & 9818.0 & 0.00 & 3.8 \\
\hline
\end{tabular}
\label{PWET40-10}
\end{table}

\begin{table}[!htbp]
\caption{Results for $P||\sum w^\prime_jE_j + w_jT_j$ with $50$ jobs and $10$ machines}
\centering
\onehalfspacing
\scriptsize
\renewcommand{\tabcolsep}{0.04cm}
\begin{tabular}{cccccccccccccccccccccccccc}
\hline
\multirow{4}{*}{Inst} & \multirow{4}{*}{BKS }&  & \multicolumn{5}{c}{ILS+PR} &  & \multicolumn{5}{c}{GA+LS+PR} &  & \multicolumn{5}{c}{ILS-M} &  & \multicolumn{5}{c}{UILS} \\
\cline{3-26}
& &  & \multicolumn{1}{c}{Best} &  & \multicolumn{3}{c}{Average} &  & \multicolumn{1}{c}{Best} &  & \multicolumn{3}{c}{Average} &  & \multicolumn{1}{c}{Best} &  & \multicolumn{3}{c}{Average} &  & \multicolumn{1}{c}{Best} &  & \multicolumn{3}{c}{Average} \\
\cline{4-4}\cline{6-8}\cline{10-10}\cline{12-14}\cline{16-16}\cline{18-20}\cline{22-22}\cline{24-26}
& & & \multirow{2}{*}{Cost} &  & \multirow{2}{*}{Cost} & Gap & Time &  & \multirow{2}{*}{Cost} & & \multirow{2}{*}{Cost} & Gap & Time & & \multirow{2}{*}{Cost} & & \multirow{2}{*}{Cost} & Gap & Time & & \multirow{2}{*}{Cost} & & \multirow{2}{*}{Cost} & Gap & Time \\
 & & & & & & (\%) & (s) & & & & & (\%) & (s) & & & & & (\%) & (s) & & & & & (\%) & (s) \\
\hline
1  & 9154 &  & 9168 &  & 9168.0 & 0.15 & 894.7 &    & 9154 &  & 9157.3 & 0.04 & 15211.1 &    & 9154 &  & 9157.3 & 0.04 & 145.9 &    & 8879 &  & 8885.2 & -2.94 & 14.3 \\ 
31 & \textbf{3839} &  & 3839 &  & 3839.0 & 0.00 & 881.7 &    & 3839 &  & 3839.0 & 0.00 & 11387.9 &    & 3839 &  & 3839.0 & 0.00 & 125.6 &    & 3839 &  & 3839.0 & 0.00 & 8.8 \\ 
61 & \textbf{5856} &  & 5856 &  & 5856.0 & 0.00 & 800.3 &    & 5856 &  & 5856.0 & 0.00 & 7480.1 &    & 5856 &  & 5856.0 & 0.00 & 115.1 &    & 5856 &  & 5856.0 & 0.00 & 7.4 \\ 
91 & \textbf{14527} &  & 14527 &  & 14527.0 & 0.00 & 857.2 &    & 14527 &  & 14527.0 & 0.00 & 9463.2 &    & 14527 &  & 14527.0 & 0.00 & 144.0 &    & 14527 &  & 14527.0 & 0.00 & 8.1 \\
\hline
\end{tabular}
\label{PWET50-10}
\end{table}

\begin{table}[H]
\caption{Results for $P||\sum w^\prime_jE_j + w_jT_j$ with $100$ jobs and $10$ machines}
\centering
\onehalfspacing
\scriptsize
\renewcommand{\tabcolsep}{0.04cm}
\begin{tabular}{cccccccccccccccccccccccccc}
\hline
\multirow{4}{*}{Inst} & \multirow{4}{*}{BKS }&  & \multicolumn{5}{c}{ILS+PR} &  & \multicolumn{5}{c}{GA+LS+PR} &  & \multicolumn{5}{c}{ILS-M} &  & \multicolumn{5}{c}{UILS} \\
\cline{3-26}
& &  & \multicolumn{1}{c}{Best} &  & \multicolumn{3}{c}{Average} &  & \multicolumn{1}{c}{Best} &  & \multicolumn{3}{c}{Average} &  & \multicolumn{1}{c}{Best} &  & \multicolumn{3}{c}{Average} &  & \multicolumn{1}{c}{Best} &  & \multicolumn{3}{c}{Average} \\
\cline{4-4}\cline{6-8}\cline{10-10}\cline{12-14}\cline{16-16}\cline{18-20}\cline{22-22}\cline{24-26}
& & & \multirow{2}{*}{Cost} &  & \multirow{2}{*}{Cost} & Gap & Time &  & \multirow{2}{*}{Cost} & & \multirow{2}{*}{Cost} & Gap & Time & & \multirow{2}{*}{Cost} & & \multirow{2}{*}{Cost} & Gap & Time & & \multirow{2}{*}{Cost} & & \multirow{2}{*}{Cost} & Gap & Time \\
 & & & & & & (\%) & (s) & & & & & (\%) & (s) & & & & & (\%) & (s) & & & & & (\%) & (s) \\
\hline
1 & 33278 &  & 33344 &  & 33354.0 & 5.92 & 17909.0 &  & 33278 &  & 33278.0 & 0.00 & 123686.2 &    & 33280 &  & 33289.3 & 0.03 & 19265.8 &    & 31251 &  & 31268.2 & -6.04 & 216.3 \\
31 & \textbf{17012} &  & 17150 &  & 17154.3 & 0.84 & 17047.1 &    & 17022 &  & 17022.0 & 0.06 & 128010.8 &    & 17021 &  & 17043.7 & 0.19 & 9060.8 &    & 17072 &  & 17091.3 & 0.47 & 171.3 \\ 
61 & \textbf{14634} &  & 14684 &  & 14684.0 & 0.34 & 16924.9 &    & 14661 &  & 14661.0 & 0.18 & 128529.3 &    & 14663 &  & 14664.0 & 0.21 & 3529.7 &    & 14677 &  & 14696.9 & 0.43 & 114.4 \\ 
91 & \textbf{35784} &  & 35800 &  & 35805.7 & 0.06 & 16591.3 &    & 35788 &  & 35788.0 & 0.01 & 151380.8 &    & 35791 &  & 35791.0 & 0.02 & 3303.5 &    & 35788 &  & 35792.7 & 0.02 & 97.5 \\
\hline
\end{tabular}
\label{PWET100-10}
\end{table}


\subsection{$R||\sum w_jT_j$}
\label{app3}

Detailed results obtained for the instances of \cite{SenBulbul2015} (see Tables \ref{RwT40_2}-\ref{RwT200_5}).

\begin{center}
{\scriptsize
\onehalfspacing
\setlength{\tabcolsep}{0.5mm}
\setlength{\LTcapwidth}{10in}
\begin{longtable}{ccccccccccccccc}
  \caption{Results for $R||\sum w_jT_j$ with $40$ jobs and $2$ machines}\\
	\hline
	\multirow{4}{*}{Inst}& \multicolumn{7}{c}{\cite{SenBulbul2015}} &  & \multicolumn{6}{c}{UILS}\\
	\cline{2-8}\cline{10-15}
	& \multicolumn{3}{c}{Best} &  & \multicolumn{3}{c}{PR+Benders+SiPS} & & Best & & \multicolumn{4}{c}{Average} \\
	\cline{2-4} \cline{6-8} \cline{10-10} \cline{12-15}
	& \multirow{2}{*}{LB} & \multirow{2}{*}{UB} & Gap$_{\text{LB}}$ & & \multirow{2}{*}{UB} & Gap$_{\text{LB}}$ & Time & & \multirow{2}{*}{UB} &  & \multirow{2}{*}{UB} & Gap$_{\text{LB}}$ & Time to & Time \\ 
	& & & ($\%$) & & & ($\%$) & (s) & & & & & ($\%$) & BKS (s) & (s) \\
	\hline
	\endfirsthead
	\caption[]{(\textit{Continued})}\\
	\hline
	\multirow{4}{*}{Inst}& \multicolumn{7}{c}{\cite{SenBulbul2015}} &  & \multicolumn{6}{c}{UILS}\\
	\cline{2-8}\cline{10-15}
	& \multicolumn{3}{c}{Best} &  & \multicolumn{3}{c}{PR+Benders+SiPS} & & Best & & \multicolumn{4}{c}{Average} \\
	\cline{2-4} \cline{6-8} \cline{10-10} \cline{12-15}
	& \multirow{2}{*}{LB} & \multirow{2}{*}{UB} & Gap$_{\text{LB}}$ & & \multirow{2}{*}{UB} & Gap$_{\text{LB}}$ & Time & & \multirow{2}{*}{UB} &  & \multirow{2}{*}{UB} & Gap$_{\text{LB}}$ & Time to & Time \\ 
	& & & ($\%$) & & & ($\%$) & (s) & & & & & ($\%$) & BKS (s) & (s) \\
	\hline
	\endhead
	\hline
	\multicolumn{15}{r}{\textit{Continued on next page}} \\
	\endfoot
	\endlastfoot	
		1 & 10375.5 & 10463 & 0.84 &  & 10555 & 1.73 & 0.3 &  & 10463 &  & 10463.0 & 0.84 & $<$0.1 & 3.4 \\ 
		3 & 9092.6 & 9234 & 1.56 &  & 9332 & 2.63 & 1.9 &  & 9174 &  & 9174.0 & 0.90 & $<$0.1 & 4.2 \\ 
		5 & 11559.2 & 11614 & 0.47 &  & 11627 & 0.59 & 1.9 &  & 11588 &  & 11588.0 & 0.25 & $<$0.1 & 3.7 \\ 
		7 & 12951.3 & 13071 & 0.92 &  & 13157 & 1.59 & 1.7 &  & 13071 &  & 13071.0 & 0.92 & $<$0.1 & 3.5 \\ 
		9 & 7816.9 & 7864 & 0.60 &  & 7912 & 1.22 & 2.7 &  & 7864 &  & 7864.0 & 0.60 & $<$0.1 & 3.6 \\ 
		11 & 5801.2 & 5858 & 0.98 &  & 5858 & 0.98 & 1.7 &  & 5858 &  & 5858.0 & 0.98 & $<$0.1 & 3.7 \\ 
		13 & 5509.2 & 5694 & 3.35 &  & 5694 & 3.35 & 1.8 &  & 5656 &  & 5656.0 & 2.66 & $<$0.1 & 4.4 \\ 
		15 & 6728.6 & 6995 & 3.96 &  & 6995 & 3.96 & 1.6 &  & 6860 &  & 6860.0 & 1.95 & $<$0.1 & 4.5 \\
\end{longtable}
}
\end{center}

\begin{center}
{\scriptsize
\onehalfspacing
\setlength{\tabcolsep}{0.5mm}
\setlength{\LTcapwidth}{10in}
\begin{longtable}{ccccccccccccccc}
  \caption{Results for $R||\sum w_jT_j$ with $60$ jobs and $2$ machines}\\
	\hline
	\multirow{4}{*}{Inst}& \multicolumn{7}{c}{\cite{SenBulbul2015}} &  & \multicolumn{6}{c}{UILS}\\
	\cline{2-8}\cline{10-15}
	& \multicolumn{3}{c}{Best} &  & \multicolumn{3}{c}{PR+Benders+SiPS} & & Best & & \multicolumn{4}{c}{Average} \\
	\cline{2-4} \cline{6-8} \cline{10-10} \cline{12-15}
	& \multirow{2}{*}{LB} & \multirow{2}{*}{UB} & Gap$_{\text{LB}}$ & & \multirow{2}{*}{UB} & Gap$_{\text{LB}}$ & Time & & \multirow{2}{*}{UB} &  & \multirow{2}{*}{UB} & Gap$_{\text{LB}}$ & Time to & Time \\ 
	& & & ($\%$) & & & ($\%$) & (s) & & & & & ($\%$) & BKS (s) & (s) \\
	\hline
	\endfirsthead
	\caption[]{(\textit{Continued})}\\
	\hline
	\multirow{4}{*}{Inst}& \multicolumn{7}{c}{\cite{SenBulbul2015}} &  & \multicolumn{6}{c}{UILS}\\
	\cline{2-8}\cline{10-15}
	& \multicolumn{3}{c}{Best} &  & \multicolumn{3}{c}{PR+Benders+SiPS} & & Best & & \multicolumn{4}{c}{Average} \\
	\cline{2-4} \cline{6-8} \cline{10-10} \cline{12-15}
	& \multirow{2}{*}{LB} & \multirow{2}{*}{UB} & Gap$_{\text{LB}}$ & & \multirow{2}{*}{UB} & Gap$_{\text{LB}}$ & Time & & \multirow{2}{*}{UB} &  & \multirow{2}{*}{UB} & Gap$_{\text{LB}}$ & Time to & Time \\ 
	& & & ($\%$) & & & ($\%$) & (s) & & & & & ($\%$) & BKS (s) & (s) \\
	\hline
	\endhead
	\hline
	\multicolumn{15}{r}{\textit{Continued on next page}} \\
	\endfoot
	\endlastfoot
		1 & 24313.6 & 24532 & 0.90 &  & 24532 & 0.90 & 8.5 &  & 24397 &  & 24397.0 & 0.34 & $<$0.1 & 11.9 \\ 
		3 & 16503.5 & 16563 & 0.36 &  & 16563 & 0.36 & 2.7 &  & 16524 &  & 16524.0 & 0.12 & $<$0.1 & 9.9 \\ 
		5 & 16903.4 & 17270 & 2.17 &  & 17270 & 2.17 & 3.1 &  & 17019 &  & 17019.0 & 0.68 & $<$0.1 & 11.3 \\ 
		7 & 23400.6 & 23503 & 0.44 &  & 23503 & 0.44 & 5.3 &  & 23503 &  & 23503.0 & 0.44 & 0.2 & 12.1 \\ 
		9 & 18236.3 & 18357 & 0.66 &  & 18357 & 0.66 & 5.4 &  & 18301 &  & 18301.0 & 0.35 & $<$0.1 & 11.7 \\ 
		11 & 9990.2 & 10054 & 0.64 &  & 10054 & 0.64 & 6.5 &  & 10054 &  & 10054.0 & 0.64 & 0.1 & 13.0 \\ 
		13 & 6258.9 & 6491 & 3.71 &  & 6491 & 3.71 & 4.4 &  & 6403 &  & 6403.0 & 2.30 & 0.3 & 16.0 \\ 
		15 & 12067.1 & 12303 & 1.95 &  & 12303 & 1.95 & 6.0 &  & 12169 &  & 12169.0 & 0.84 & 0.2 & 17.0 \\
\end{longtable}
}
\end{center}

\begin{center}
{\scriptsize
\onehalfspacing
\setlength{\tabcolsep}{0.5mm}
\setlength{\LTcapwidth}{10in}
\begin{longtable}{ccccccccccccccc}
  \caption{Results for $R||\sum w_jT_j$ with $60$ jobs and $3$ machines}\\
	\hline
	\multirow{4}{*}{Inst}& \multicolumn{7}{c}{\cite{SenBulbul2015}} &  & \multicolumn{6}{c}{UILS}\\
	\cline{2-8}\cline{10-15}
	& \multicolumn{3}{c}{Best} &  & \multicolumn{3}{c}{PR+Benders+SiPS} & & Best & & \multicolumn{4}{c}{Average} \\
	\cline{2-4} \cline{6-8} \cline{10-10} \cline{12-15}
	& \multirow{2}{*}{LB} & \multirow{2}{*}{UB} & Gap$_{\text{LB}}$ & & \multirow{2}{*}{UB} & Gap$_{\text{LB}}$ & Time & & \multirow{2}{*}{UB} &  & \multirow{2}{*}{UB} & Gap$_{\text{LB}}$ & Time to & Time \\ 
	& & & ($\%$) & & & ($\%$) & (s) & & & & & ($\%$) & BKS (s) & (s) \\
	\hline
	\endfirsthead
	\caption[]{(\textit{Continued})}\\
	\hline
	\multirow{4}{*}{Inst}& \multicolumn{7}{c}{\cite{SenBulbul2015}} &  & \multicolumn{6}{c}{UILS}\\
	\cline{2-8}\cline{10-15}
	& \multicolumn{3}{c}{Best} &  & \multicolumn{3}{c}{PR+Benders+SiPS} & & Best & & \multicolumn{4}{c}{Average} \\
	\cline{2-4} \cline{6-8} \cline{10-10} \cline{12-15}
	& \multirow{2}{*}{LB} & \multirow{2}{*}{UB} & Gap$_{\text{LB}}$ & & \multirow{2}{*}{UB} & Gap$_{\text{LB}}$ & Time & & \multirow{2}{*}{UB} &  & \multirow{2}{*}{UB} & Gap$_{\text{LB}}$ & Time to & Time \\ 
	& & & ($\%$) & & & ($\%$) & (s) & & & & & ($\%$) & BKS (s) & (s) \\
	\hline
	\endhead
	\hline
	\multicolumn{15}{r}{\textit{Continued on next page}} \\
	\endfoot
	\endlastfoot
		1 & 13667.0 & 13805 & 1.01 &  & 13805 & 1.01 & 3.3 &  & 13758 &  & 13758.0 & 0.67 & 0.1 & 16.3 \\ 
		3 & 12693.0 & 12971 & 2.19 &  & 12971 & 2.19 & 1.3 &  & 12787 &  & 12787.0 & 0.74 & $<$0.1 & 14.9 \\ 
		5 & 10842.5 & 11117 & 2.53 &  & 11117 & 2.53 & 2.5 &  & 11031 &  & 11031.0 & 1.74 & $<$0.1 & 19.6 \\ 
		7 & 13132.2 & 13484 & 2.68 &  & 13484 & 2.68 & 2.3 &  & 13331 &  & 13331.0 & 1.51 & $<$0.1 & 17.5 \\ 
		9 & 12715.6 & 12883 & 1.32 &  & 12883 & 1.32 & 1.5 &  & 12883 &  & 12883.0 & 1.32 & $<$0.1 & 11.7 \\ 
		11 & 7370.9 & 7596 & 3.05 &  & 7596 & 3.05 & 4.3 &  & 7526 &  & 7526.0 & 2.10 & 1.2 & 22.6 \\ 
		13 & 7764.8 & 7983 & 2.81 &  & 7983 & 2.81 & 6.1 &  & 7892 &  & 7896.4 & 1.69 & 0.6 & 19.3 \\ 
		15 & 4677.9 & 4880 & 4.32 &  & 4880 & 4.32 & 3.8 &  & 4786 &  & 4786.0 & 2.31 & 0.3 & 18.0 \\
\end{longtable}
}
\end{center}

\begin{center}
{\scriptsize
\onehalfspacing
\setlength{\tabcolsep}{0.5mm}
\setlength{\LTcapwidth}{10in}
\begin{longtable}{ccccccccccccccc}
  \caption{Results for $R||\sum w_jT_j$ with $80$ jobs and $2$ machines}\\
	\hline
	\multirow{4}{*}{Inst}& \multicolumn{7}{c}{\cite{SenBulbul2015}} &  & \multicolumn{6}{c}{UILS}\\
	\cline{2-8}\cline{10-15}
	& \multicolumn{3}{c}{Best} &  & \multicolumn{3}{c}{PR+Benders+SiPS} & & Best & & \multicolumn{4}{c}{Average} \\
	\cline{2-4} \cline{6-8} \cline{10-10} \cline{12-15}
	& \multirow{2}{*}{LB} & \multirow{2}{*}{UB} & Gap$_{\text{LB}}$ & & \multirow{2}{*}{UB} & Gap$_{\text{LB}}$ & Time & & \multirow{2}{*}{UB} &  & \multirow{2}{*}{UB} & Gap$_{\text{LB}}$ & Time to & Time \\ 
	& & & ($\%$) & & & ($\%$) & (s) & & & & & ($\%$) & BKS (s) & (s) \\
	\hline
	\endfirsthead
	\caption[]{(\textit{Continued})}\\
	\hline
	\multirow{4}{*}{Inst}& \multicolumn{7}{c}{\cite{SenBulbul2015}} &  & \multicolumn{6}{c}{UILS}\\
	\cline{2-8}\cline{10-15}
	& \multicolumn{3}{c}{Best} &  & \multicolumn{3}{c}{PR+Benders+SiPS} & & Best & & \multicolumn{4}{c}{Average} \\
	\cline{2-4} \cline{6-8} \cline{10-10} \cline{12-15}
	& \multirow{2}{*}{LB} & \multirow{2}{*}{UB} & Gap$_{\text{LB}}$ & & \multirow{2}{*}{UB} & Gap$_{\text{LB}}$ & Time & & \multirow{2}{*}{UB} &  & \multirow{2}{*}{UB} & Gap$_{\text{LB}}$ & Time to & Time \\ 
	& & & ($\%$) & & & ($\%$) & (s) & & & & & ($\%$) & BKS (s) & (s) \\
	\hline
	\endhead
	\hline
	\multicolumn{15}{r}{\textit{Continued on next page}} \\
	\endfoot
	\endlastfoot
		1 & 26573.0 & 26915 & 1.29 &  & 26915 & 1.29 & 12.9 &  & 26637 &  & 26637.0 & 0.24 & $<$0.1 & 24.2 \\ 
		3 & 34560.7 & 34695 & 0.39 &  & 34695 & 0.39 & 9.6 &  & 34652 &  & 34652.0 & 0.26 & 0.1 & 26.7 \\ 
		5 & 47022.7 & 47338 & 0.67 &  & 47338 & 0.67 & 2.1 &  & 47269 &  & 47269.0 & 0.52 & $<$0.1 & 25.2 \\ 
		7 & 37472.8 & 37959 & 1.30 &  & 37959 & 1.30 & 15.4 &  & 37563 &  & 37563.0 & 0.24 & $<$0.1 & 28.7 \\ 
		9 & 39576.8 & 40261 & 1.73 &  & 40261 & 1.73 & 2.8 &  & 39673 &  & 39673.0 & 0.24 & $<$0.1 & 27.5 \\ 
		11 & 15348.4 & 15609 & 1.70 &  & 15609 & 1.70 & 19.5 &  & 15467 &  & 15467.0 & 0.77 & 0.1 & 43.2 \\ 
		13 & 23546.0 & 24027 & 2.04 &  & 24027 & 2.04 & 34.7 &  & 23829 &  & 23829.0 & 1.20 & 0.5 & 39.8 \\ 
		15 & 20669.3 & 20974 & 1.47 &  & 20974 & 1.47 & 9.6 &  & 20791 &  & 20791.0 & 0.59 & $<$0.1 & 28.8 \\
\end{longtable}
}
\end{center}

\begin{center}
{\scriptsize
\onehalfspacing
\setlength{\tabcolsep}{0.5mm}
\setlength{\LTcapwidth}{10in}
\begin{longtable}{ccccccccccccccc}
  \caption{Results for $R||\sum w_jT_j$ with $80$ jobs and $4$ machines}\\
	\hline
	\multirow{4}{*}{Inst}& \multicolumn{7}{c}{\cite{SenBulbul2015}} &  & \multicolumn{6}{c}{UILS}\\
	\cline{2-8}\cline{10-15}
	& \multicolumn{3}{c}{Best} &  & \multicolumn{3}{c}{PR+Benders+SiPS} & & Best & & \multicolumn{4}{c}{Average} \\
	\cline{2-4} \cline{6-8} \cline{10-10} \cline{12-15}
	& \multirow{2}{*}{LB} & \multirow{2}{*}{UB} & Gap$_{\text{LB}}$ & & \multirow{2}{*}{UB} & Gap$_{\text{LB}}$ & Time & & \multirow{2}{*}{UB} &  & \multirow{2}{*}{UB} & Gap$_{\text{LB}}$ & Time to & Time \\ 
	& & & ($\%$) & & & ($\%$) & (s) & & & & & ($\%$) & BKS (s) & (s) \\
	\hline
	\endfirsthead
	\caption[]{(\textit{Continued})}\\
	\hline
	\multirow{4}{*}{Inst}& \multicolumn{7}{c}{\cite{SenBulbul2015}} &  & \multicolumn{6}{c}{UILS}\\
	\cline{2-8}\cline{10-15}
	& \multicolumn{3}{c}{Best} &  & \multicolumn{3}{c}{PR+Benders+SiPS} & & Best & & \multicolumn{4}{c}{Average} \\
	\cline{2-4} \cline{6-8} \cline{10-10} \cline{12-15}
	& \multirow{2}{*}{LB} & \multirow{2}{*}{UB} & Gap$_{\text{LB}}$ & & \multirow{2}{*}{UB} & Gap$_{\text{LB}}$ & Time & & \multirow{2}{*}{UB} &  & \multirow{2}{*}{UB} & Gap$_{\text{LB}}$ & Time to & Time \\ 
	& & & ($\%$) & & & ($\%$) & (s) & & & & & ($\%$) & BKS (s) & (s) \\
	\hline
	\endhead
	\hline
	\multicolumn{15}{r}{\textit{Continued on next page}} \\
	\endfoot
	\endlastfoot
		1 & 10907.8 & 11194 & 2.62 &  & 11194 & 2.62 & 6.1 &  & 11096 &  & 11096.0 & 1.73 & 0.2 & 40.0 \\ 
		3 & 11945.0 & 12645 & 5.86 &  & 12645 & 5.86 & 6.3 &  & 12157 &  & 12157.0 & 1.77 & $<$0.1 & 39.4 \\ 
		5 & 12227.2 & 12594 & 3.00 &  & 12594 & 3.00 & 3.4 &  & 12369 &  & 12369.0 & 1.16 & 0.2 & 41.7 \\ 
		7 & 10923.1 & 11337 & 3.79 &  & 11337 & 3.79 & 3.4 &  & 11062 &  & 11062.0 & 1.27 & 0.2 & 49.0 \\ 
		9 & 13165.4 & 13602 & 3.32 &  & 13602 & 3.32 & 6.4 &  & 13334 &  & 13334.0 & 1.28 & 0.4 & 48.8 \\ 
		11 & 7468.8 & 7876 & 5.45 &  & 7876 & 5.45 & 5.4 &  & 7533 &  & 7533.0 & 0.86 & 0.3 & 49.2 \\ 
		13 & 2845.5 & 3003 & 5.54 &  & 3003 & 5.54 & 4.6 &  & 2961 &  & 2970.3 & 4.39 & 9.4 & 54.4 \\ 
		15 & 3841.4 & 3982 & 3.66 &  & 3982 & 3.66 & 13.1 &  & 3914 &  & 3923.1 & 2.13 & 5.7 & 73.1 \\
\end{longtable}
}
\end{center}

\begin{center}
{\scriptsize
\onehalfspacing
\setlength{\tabcolsep}{0.5mm}
\setlength{\LTcapwidth}{10in}
\begin{longtable}{ccccccccccccccc}
  \caption{Results for $R||\sum w_jT_j$ with $90$ jobs and $3$ machines}\\
	\hline
	\multirow{4}{*}{Inst}& \multicolumn{7}{c}{\cite{SenBulbul2015}} &  & \multicolumn{6}{c}{UILS}\\
	\cline{2-8}\cline{10-15}
	& \multicolumn{3}{c}{Best} &  & \multicolumn{3}{c}{PR+Benders+SiPS} & & Best & & \multicolumn{4}{c}{Average} \\
	\cline{2-4} \cline{6-8} \cline{10-10} \cline{12-15}
	& \multirow{2}{*}{LB} & \multirow{2}{*}{UB} & Gap$_{\text{LB}}$ & & \multirow{2}{*}{UB} & Gap$_{\text{LB}}$ & Time & & \multirow{2}{*}{UB} &  & \multirow{2}{*}{UB} & Gap$_{\text{LB}}$ & Time to & Time \\ 
	& & & ($\%$) & & & ($\%$) & (s) & & & & & ($\%$) & BKS (s) & (s) \\
	\hline
	\endfirsthead
	\caption[]{(\textit{Continued})}\\
	\hline
	\multirow{4}{*}{Inst}& \multicolumn{7}{c}{\cite{SenBulbul2015}} &  & \multicolumn{6}{c}{UILS}\\
	\cline{2-8}\cline{10-15}
	& \multicolumn{3}{c}{Best} &  & \multicolumn{3}{c}{PR+Benders+SiPS} & & Best & & \multicolumn{4}{c}{Average} \\
	\cline{2-4} \cline{6-8} \cline{10-10} \cline{12-15}
	& \multirow{2}{*}{LB} & \multirow{2}{*}{UB} & Gap$_{\text{LB}}$ & & \multirow{2}{*}{UB} & Gap$_{\text{LB}}$ & Time & & \multirow{2}{*}{UB} &  & \multirow{2}{*}{UB} & Gap$_{\text{LB}}$ & Time to & Time \\ 
	& & & ($\%$) & & & ($\%$) & (s) & & & & & ($\%$) & BKS (s) & (s) \\
	\hline
	\endhead
	\hline
	\multicolumn{15}{r}{\textit{Continued on next page}} \\
	\endfoot
	\endlastfoot
		1 & 28789.8 & 29385 & 2.07 &  & 29385 & 2.07 & 3.5 &  & 29000 &  & 29000.0 & 0.73 & 0.1 & 66.6 \\ 
		3 & 26042.2 & 26348 & 1.17 &  & 26348 & 1.17 & 9.3 &  & 26244 &  & 26251.2 & 0.80 & 0.9 & 69.1 \\ 
		5 & 24344.3 & 24751 & 1.67 &  & 24751 & 1.67 & 10.4 &  & 24519 &  & 24519.0 & 0.72 & 0.2 & 55.8 \\ 
		7 & 22145.4 & 22622 & 2.15 &  & 22622 & 2.15 & 3.5 &  & 22262 &  & 22262.0 & 0.53 & 0.1 & 70.5 \\ 
		9 & 32342.0 & 32656 & 0.97 &  & 32656 & 0.97 & 2.7 &  & 32471 &  & 32471.0 & 0.40 & 0.6 & 58.3 \\ 
		11 & 8748.2 & 8920 & 1.96 &  & 8920 & 1.96 & 14.2 &  & 8830 &  & 8830.0 & 0.93 & 0.4 & 62.2 \\ 
		13 & 7877.6 & 8007 & 1.64 &  & 8007 & 1.64 & 10.2 &  & 8007 &  & 8007.0 & 1.64 & 22.4 & 77.0 \\ 
		15 & 12258.6 & 12895 & 5.19 &  & 12895 & 5.19 & 12.2 &  & 12350 &  & 12350.0 & 0.75 & 0.2 & 73.3 \\
\end{longtable}
}
\end{center}

\begin{center}
{\scriptsize
\onehalfspacing
\setlength{\tabcolsep}{0.5mm}
\setlength{\LTcapwidth}{10in}
\begin{longtable}{ccccccccccccccc}
  \caption{Results for $R||\sum w_jT_j$ with $100$ jobs and $5$ machines}\\
	\hline
	\multirow{4}{*}{Inst}& \multicolumn{7}{c}{\cite{SenBulbul2015}} &  & \multicolumn{6}{c}{UILS}\\
	\cline{2-8}\cline{10-15}
	& \multicolumn{3}{c}{Best} &  & \multicolumn{3}{c}{PR+Benders+SiPS} & & Best & & \multicolumn{4}{c}{Average} \\
	\cline{2-4} \cline{6-8} \cline{10-10} \cline{12-15}
	& \multirow{2}{*}{LB} & \multirow{2}{*}{UB} & Gap$_{\text{LB}}$ & & \multirow{2}{*}{UB} & Gap$_{\text{LB}}$ & Time & & \multirow{2}{*}{UB} &  & \multirow{2}{*}{UB} & Gap$_{\text{LB}}$ & Time to & Time \\ 
	& & & ($\%$) & & & ($\%$) & (s) & & & & & ($\%$) & BKS (s) & (s) \\
	\hline
	\endfirsthead
	\caption[]{(\textit{Continued})}\\
	\hline
	\multirow{4}{*}{Inst}& \multicolumn{7}{c}{\cite{SenBulbul2015}} &  & \multicolumn{6}{c}{UILS}\\
	\cline{2-8}\cline{10-15}
	& \multicolumn{3}{c}{Best} &  & \multicolumn{3}{c}{PR+Benders+SiPS} & & Best & & \multicolumn{4}{c}{Average} \\
	\cline{2-4} \cline{6-8} \cline{10-10} \cline{12-15}
	& \multirow{2}{*}{LB} & \multirow{2}{*}{UB} & Gap$_{\text{LB}}$ & & \multirow{2}{*}{UB} & Gap$_{\text{LB}}$ & Time & & \multirow{2}{*}{UB} &  & \multirow{2}{*}{UB} & Gap$_{\text{LB}}$ & Time to & Time \\ 
	& & & ($\%$) & & & ($\%$) & (s) & & & & & ($\%$) & BKS (s) & (s) \\
	\hline
	\endhead
	\hline
	\multicolumn{15}{r}{\textit{Continued on next page}} \\
	\endfoot
	\endlastfoot
		1 & 12461.6 & 13304 & 6.76 &  & 13304 & 6.76 & 7.5 &  & 12809 &  & 12809.0 & 2.79 & 0.4 & 90.5 \\ 
		3 & 9514.1 & 10185 & 7.05 &  & 10185 & 7.05 & 5.5 &  & 9880 &  & 9896.1 & 4.02 & 0.3 & 106.3 \\ 
		5 & 8023.9 & 8584 & 6.98 &  & 8584 & 6.98 & 6.6 &  & 8332 &  & 8332.4 & 3.85 & 0.3 & 94.0 \\ 
		7 & 12327.0 & 12873 & 4.43 &  & 12873 & 4.43 & 7.7 &  & 12652 &  & 12652.0 & 2.64 & 1.0 & 100.3 \\ 
		9 & 14028.0 & 14846 & 5.83 &  & 14846 & 5.83 & 6.5 &  & 14400 &  & 14411.7 & 2.74 & 0.2 & 103.6 \\ 
		11 & 3861.3 & 4408 & 14.16 &  & 4408 & 14.16 & 7.5 &  & 4181 &  & 4198.1 & 8.72 & 3.2 & 131.2 \\ 
		13 & 3096.6 & 3615 & 16.74 &  & 3615 & 16.74 & 23.3 &  & 3405 &  & 3424.0 & 10.57 & 1.6 & 127.8 \\ 
		15 & 3280.8 & 3669 & 11.83 &  & 3669 & 11.83 & 13.5 &  & 3562 &  & 3586.2 & 9.31 & 4.4 & 143.3 \\
\end{longtable}
}
\end{center}

\begin{center}
{\scriptsize
\onehalfspacing
\setlength{\tabcolsep}{0.5mm}
\setlength{\LTcapwidth}{10in}
\begin{longtable}{ccccccccccccccc}
  \caption{Results for $R||\sum w_jT_j$ with $120$ jobs and $3$ machines}\\
	\hline
	\multirow{4}{*}{Inst}& \multicolumn{7}{c}{\cite{SenBulbul2015}} &  & \multicolumn{6}{c}{UILS}\\
	\cline{2-8}\cline{10-15}
	& \multicolumn{3}{c}{Best} &  & \multicolumn{3}{c}{PR+Benders+SiPS} & & Best & & \multicolumn{4}{c}{Average} \\
	\cline{2-4} \cline{6-8} \cline{10-10} \cline{12-15}
	& \multirow{2}{*}{LB} & \multirow{2}{*}{UB} & Gap$_{\text{LB}}$ & & \multirow{2}{*}{UB} & Gap$_{\text{LB}}$ & Time & & \multirow{2}{*}{UB} &  & \multirow{2}{*}{UB} & Gap$_{\text{LB}}$ & Time to & Time \\ 
	& & & ($\%$) & & & ($\%$) & (s) & & & & & ($\%$) & BKS (s) & (s) \\
	\hline
	\endfirsthead
	\caption[]{(\textit{Continued})}\\
	\hline
	\multirow{4}{*}{Inst}& \multicolumn{7}{c}{\cite{SenBulbul2015}} &  & \multicolumn{6}{c}{UILS}\\
	\cline{2-8}\cline{10-15}
	& \multicolumn{3}{c}{Best} &  & \multicolumn{3}{c}{PR+Benders+SiPS} & & Best & & \multicolumn{4}{c}{Average} \\
	\cline{2-4} \cline{6-8} \cline{10-10} \cline{12-15}
	& \multirow{2}{*}{LB} & \multirow{2}{*}{UB} & Gap$_{\text{LB}}$ & & \multirow{2}{*}{UB} & Gap$_{\text{LB}}$ & Time & & \multirow{2}{*}{UB} &  & \multirow{2}{*}{UB} & Gap$_{\text{LB}}$ & Time to & Time \\ 
	& & & ($\%$) & & & ($\%$) & (s) & & & & & ($\%$) & BKS (s) & (s) \\
	\hline
	\endhead
	\hline
	\multicolumn{15}{r}{\textit{Continued on next page}} \\
	\endfoot
	\endlastfoot
		1 & 38987.5 & 39759 & 1.98 &  & 39759 & 1.98 & 7.5 &  & 39556 &  & 39561.0 & 1.47 & 0.7 & 150.1 \\ 
		3 & 39039.9 & 40141 & 2.82 &  & 40141 & 2.82 & 7.7 &  & 39614 &  & 39614.0 & 1.47 & 0.2 & 107.9 \\ 
		5 & 37753.8 & 38690 & 2.48 &  & 38690 & 2.48 & 8.1 &  & 38021 &  & 38021.0 & 0.71 & 0.2 & 154.3 \\ 
		7 & 34455.4 & 35204 & 2.17 &  & 35204 & 2.17 & 9.7 &  & 34919 &  & 34919.0 & 1.35 & 0.3 & 131.4 \\ 
		9 & 51858.5 & 52976 & 2.15 &  & 52976 & 2.15 & 7.2 &  & 52560 &  & 52560.0 & 1.35 & 0.3 & 153.8 \\ 
		11 & 19289.5 & 19734 & 2.30 &  & 19734 & 2.30 & 28.1 &  & 19679 &  & 19679.0 & 2.02 & 33.7 & 184.0 \\ 
		13 & 13448.5 & 14083 & 4.72 &  & 14083 & 4.72 & 28.3 &  & 13913 &  & 13923.5 & 3.53 & 0.7 & 166.8 \\ 
		15 & 11317.0 & 12832 & 13.39 &  & 12832 & 13.39 & 18.5 &  & 11957 &  & 11957.4 & 5.66 & 0.2 & 190.3 \\
\end{longtable}
}
\end{center}

\begin{center}
{\scriptsize
\onehalfspacing
\setlength{\tabcolsep}{0.5mm}
\setlength{\LTcapwidth}{10in}
\begin{longtable}{ccccccccccccccc}
  \caption{Results for $R||\sum w_jT_j$ with $120$ jobs and $4$ machines}\\
	\hline
	\multirow{4}{*}{Inst}& \multicolumn{7}{c}{\cite{SenBulbul2015}} &  & \multicolumn{6}{c}{UILS}\\
	\cline{2-8}\cline{10-15}
	& \multicolumn{3}{c}{Best} &  & \multicolumn{3}{c}{PR+Benders+SiPS} & & Best & & \multicolumn{4}{c}{Average} \\
	\cline{2-4} \cline{6-8} \cline{10-10} \cline{12-15}
	& \multirow{2}{*}{LB} & \multirow{2}{*}{UB} & Gap$_{\text{LB}}$ & & \multirow{2}{*}{UB} & Gap$_{\text{LB}}$ & Time & & \multirow{2}{*}{UB} &  & \multirow{2}{*}{UB} & Gap$_{\text{LB}}$ & Time to & Time \\ 
	& & & ($\%$) & & & ($\%$) & (s) & & & & & ($\%$) & BKS (s) & (s) \\
	\hline
	\endfirsthead
	\caption[]{(\textit{Continued})}\\
	\hline
	\multirow{4}{*}{Inst}& \multicolumn{7}{c}{\cite{SenBulbul2015}} &  & \multicolumn{6}{c}{UILS}\\
	\cline{2-8}\cline{10-15}
	& \multicolumn{3}{c}{Best} &  & \multicolumn{3}{c}{PR+Benders+SiPS} & & Best & & \multicolumn{4}{c}{Average} \\
	\cline{2-4} \cline{6-8} \cline{10-10} \cline{12-15}
	& \multirow{2}{*}{LB} & \multirow{2}{*}{UB} & Gap$_{\text{LB}}$ & & \multirow{2}{*}{UB} & Gap$_{\text{LB}}$ & Time & & \multirow{2}{*}{UB} &  & \multirow{2}{*}{UB} & Gap$_{\text{LB}}$ & Time to & Time \\ 
	& & & ($\%$) & & & ($\%$) & (s) & & & & & ($\%$) & BKS (s) & (s) \\
	\hline
	\endhead
	\hline
	\multicolumn{15}{r}{\textit{Continued on next page}} \\
	\endfoot
	\endlastfoot
		1 & 16974.4 & 17556 & 3.43 &  & 17556 & 3.43 & 5.6 &  & 17292 &  & 17295.0 & 1.89 & 0.4 & 163.3 \\ 
		3 & 25327.5 & 26224 & 3.54 &  & 26224 & 3.54 & 15.4 &  & 25811 &  & 25812.3 & 1.91 & 0.3 & 170.7 \\ 
		5 & 18694.4 & 19174 & 2.57 &  & 19174 & 2.57 & 15.8 &  & 19069 &  & 19069.0 & 2.00 & 1.3 & 176.0 \\ 
		7 & 23203.5 & 23845 & 2.76 &  & 23845 & 2.76 & 16.1 &  & 23580 &  & 23580.4 & 1.62 & 10.8 & 240.2 \\ 
		9 & 24103.9 & 24733 & 2.61 &  & 24733 & 2.61 & 14.9 &  & 24511 &  & 24511.0 & 1.69 & 1.6 & 196.0 \\ 
		11 & 11597.8 & 12584 & 8.50 &  & 12584 & 8.50 & 22.4 &  & 12248 &  & 12256.2 & 5.68 & 3.6 & 209.6 \\ 
		13 & 5985.5 & 6620 & 10.60 &  & 6620 & 10.60 & 21.6 &  & 6465 &  & 6467.8 & 8.06 & 2.7 & 237.4 \\ 
		15 & 6831.6 & 7229 & 5.82 &  & 7229 & 5.82 & 17.2 &  & 7131 &  & 7150.4 & 4.67 & 6.4 & 237.3 \\
\end{longtable}
}
\end{center}

\begin{center}
{\scriptsize
\onehalfspacing
\setlength{\tabcolsep}{0.5mm}
\setlength{\LTcapwidth}{10in}
\begin{longtable}{ccccccccccccccc}
  \caption{Results for $R||\sum w_jT_j$ with $150$ jobs and $5$ machines}\\
	\hline
	\multirow{4}{*}{Inst}& \multicolumn{7}{c}{\cite{SenBulbul2015}} &  & \multicolumn{6}{c}{UILS}\\
	\cline{2-8}\cline{10-15}
	& \multicolumn{3}{c}{Best} &  & \multicolumn{3}{c}{PR+Benders+SiPS} & & Best & & \multicolumn{4}{c}{Average} \\
	\cline{2-4} \cline{6-8} \cline{10-10} \cline{12-15}
	& \multirow{2}{*}{LB} & \multirow{2}{*}{UB} & Gap$_{\text{LB}}$ & & \multirow{2}{*}{UB} & Gap$_{\text{LB}}$ & Time & & \multirow{2}{*}{UB} &  & \multirow{2}{*}{UB} & Gap$_{\text{LB}}$ & Time to & Time \\ 
	& & & ($\%$) & & & ($\%$) & (s) & & & & & ($\%$) & BKS (s) & (s) \\
	\hline
	\endfirsthead
	\caption[]{(\textit{Continued})}\\
	\hline
	\multirow{4}{*}{Inst}& \multicolumn{7}{c}{\cite{SenBulbul2015}} &  & \multicolumn{6}{c}{UILS}\\
	\cline{2-8}\cline{10-15}
	& \multicolumn{3}{c}{Best} &  & \multicolumn{3}{c}{PR+Benders+SiPS} & & Best & & \multicolumn{4}{c}{Average} \\
	\cline{2-4} \cline{6-8} \cline{10-10} \cline{12-15}
	& \multirow{2}{*}{LB} & \multirow{2}{*}{UB} & Gap$_{\text{LB}}$ & & \multirow{2}{*}{UB} & Gap$_{\text{LB}}$ & Time & & \multirow{2}{*}{UB} &  & \multirow{2}{*}{UB} & Gap$_{\text{LB}}$ & Time to & Time \\ 
	& & & ($\%$) & & & ($\%$) & (s) & & & & & ($\%$) & BKS (s) & (s) \\
	\hline
	\endhead
	\hline
	\multicolumn{15}{r}{\textit{Continued on next page}} \\
	\endfoot
	\endlastfoot
		1 & 28038.8 & 29495 & 5.19 &  & 29495 & 5.19 & 17.9 &  & 28872 &  & 28876.5 & 2.99 & 0.5 & 401.3 \\ 
		3 & 22152.2 & 23107 & 4.31 &  & 23107 & 4.31 & 21.7 &  & 22564 &  & 22597.8 & 2.01 & 1.0 & 424.6 \\ 
		5 & 22212.3 & 22978 & 3.45 &  & 22978 & 3.45 & 19.8 &  & 22671 &  & 22676.0 & 2.09 & 2.3 & 391.2 \\ 
		7 & 30640.9 & 31516 & 2.86 &  & 31516 & 2.86 & 18.8 &  & 31213 &  & 31231.3 & 1.93 & 8.1 & 493.0 \\ 
		9 & 25077.3 & 26089 & 4.03 &  & 26089 & 4.03 & 23.8 &  & 25638 &  & 25663.4 & 2.34 & 2.1 & 418.7 \\ 
		11 & 7365.9 & 7719 & 4.79 &  & 7719 & 4.79 & 37.5 &  & 7665 &  & 7685.5 & 4.34 & 179.8 & 552.3 \\ 
		13 & 4849.6 & 5421 & 11.78 &  & 5421 & 11.78 & 37.5 &  & 5068 &  & 5094.4 & 5.05 & 3.9 & 557.0 \\ 
		15 & 5076.4 & 5681 & 11.91 &  & 5681 & 11.91 & 32.7 &  & 5455 &  & 5538.3 & 9.10 & 18.3 & 567.7 \\
\end{longtable}
}
\end{center}

\begin{center}
{\scriptsize
\onehalfspacing
\setlength{\tabcolsep}{0.5mm}
\setlength{\LTcapwidth}{10in}
\begin{longtable}{ccccccccccccccc}
  \caption{Results for $R||\sum w_jT_j$ with $160$ jobs and $4$ machines}\\
	\hline
	\multirow{4}{*}{Inst}& \multicolumn{7}{c}{\cite{SenBulbul2015}} &  & \multicolumn{6}{c}{UILS}\\
	\cline{2-8}\cline{10-15}
	& \multicolumn{3}{c}{Best} &  & \multicolumn{3}{c}{PR+Benders+SiPS} & & Best & & \multicolumn{4}{c}{Average} \\
	\cline{2-4} \cline{6-8} \cline{10-10} \cline{12-15}
	& \multirow{2}{*}{LB} & \multirow{2}{*}{UB} & Gap$_{\text{LB}}$ & & \multirow{2}{*}{UB} & Gap$_{\text{LB}}$ & Time & & \multirow{2}{*}{UB} &  & \multirow{2}{*}{UB} & Gap$_{\text{LB}}$ & Time to & Time \\ 
	& & & ($\%$) & & & ($\%$) & (s) & & & & & ($\%$) & BKS (s) & (s) \\
	\hline
	\endfirsthead
	\caption[]{(\textit{Continued})}\\
	\hline
	\multirow{4}{*}{Inst}& \multicolumn{7}{c}{\cite{SenBulbul2015}} &  & \multicolumn{6}{c}{UILS}\\
	\cline{2-8}\cline{10-15}
	& \multicolumn{3}{c}{Best} &  & \multicolumn{3}{c}{PR+Benders+SiPS} & & Best & & \multicolumn{4}{c}{Average} \\
	\cline{2-4} \cline{6-8} \cline{10-10} \cline{12-15}
	& \multirow{2}{*}{LB} & \multirow{2}{*}{UB} & Gap$_{\text{LB}}$ & & \multirow{2}{*}{UB} & Gap$_{\text{LB}}$ & Time & & \multirow{2}{*}{UB} &  & \multirow{2}{*}{UB} & Gap$_{\text{LB}}$ & Time to & Time \\ 
	& & & ($\%$) & & & ($\%$) & (s) & & & & & ($\%$) & BKS (s) & (s) \\
	\hline
	\endhead
	\hline
	\multicolumn{15}{r}{\textit{Continued on next page}} \\
	\endfoot
	\endlastfoot
		1 & 36992.8 & 38087 & 2.96 &  & 38087 & 2.96 & 9.5 &  & 37544 &  & 37544.0 & 1.49 & 0.5 & 441.1 \\ 
		3 & 40327.7 & 41424 & 2.72 &  & 41424 & 2.72 & 10.0 &  & 40851 &  & 40852.2 & 1.30 & 1.2 & 479.6 \\ 
		5 & 43588.6 & 45002 & 3.24 &  & 45002 & 3.24 & 10.0 &  & 44641 &  & 44641.5 & 2.42 & 1.2 & 389.9 \\ 
		7 & 30606.1 & 31741 & 3.71 &  & 31741 & 3.71 & 9.9 &  & 31250 &  & 31254.7 & 2.12 & 1.1 & 486.3 \\ 
		9 & 44563.7 & 45818 & 2.81 &  & 45818 & 2.81 & 11.1 &  & 45333 &  & 45337.9 & 1.74 & 1.5 & 412.2 \\ 
		11 & 14171.4 & 14670 & 3.52 &  & 14670 & 3.52 & 43.2 &  & 14513 &  & 14523.0 & 2.48 & 8.9 & 486.1 \\ 
		13 & 11753.1 & 12783 & 8.76 &  & 12783 & 8.76 & 44.9 &  & 12259 &  & 12347.2 & 5.05 & 4.4 & 569.7 \\ 
		15 & 10650.7 & 11368 & 6.73 &  & 11368 & 6.73 & 35.3 &  & 11157 &  & 11159.9 & 4.78 & 13.6 & 592.4 \\
\end{longtable}
}
\end{center}

\begin{center}
{\scriptsize
\onehalfspacing
\setlength{\tabcolsep}{0.5mm}
\setlength{\LTcapwidth}{10in}
\begin{longtable}{ccccccccccccccc}
  \caption{Results for $R||\sum w_jT_j$ with $200$ jobs and $5$ machines}\\
	\hline
	\multirow{4}{*}{Inst}& \multicolumn{7}{c}{\cite{SenBulbul2015}} &  & \multicolumn{6}{c}{UILS}\\
	\cline{2-8}\cline{10-15}
	& \multicolumn{3}{c}{Best} &  & \multicolumn{3}{c}{PR+Benders+SiPS} & & Best & & \multicolumn{4}{c}{Average} \\
	\cline{2-4} \cline{6-8} \cline{10-10} \cline{12-15}
	& \multirow{2}{*}{LB} & \multirow{2}{*}{UB} & Gap$_{\text{LB}}$ & & \multirow{2}{*}{UB} & Gap$_{\text{LB}}$ & Time & & \multirow{2}{*}{UB} &  & \multirow{2}{*}{UB} & Gap$_{\text{LB}}$ & Time to & Time \\ 
	& & & ($\%$) & & & ($\%$) & (s) & & & & & ($\%$) & BKS (s) & (s) \\
	\hline
	\endfirsthead
	\caption[]{(\textit{Continued})}\\
	\hline
	\multirow{4}{*}{Inst}& \multicolumn{7}{c}{\cite{SenBulbul2015}} &  & \multicolumn{6}{c}{UILS}\\
	\cline{2-8}\cline{10-15}
	& \multicolumn{3}{c}{Best} &  & \multicolumn{3}{c}{PR+Benders+SiPS} & & Best & & \multicolumn{4}{c}{Average} \\
	\cline{2-4} \cline{6-8} \cline{10-10} \cline{12-15}
	& \multirow{2}{*}{LB} & \multirow{2}{*}{UB} & Gap$_{\text{LB}}$ & & \multirow{2}{*}{UB} & Gap$_{\text{LB}}$ & Time & & \multirow{2}{*}{UB} &  & \multirow{2}{*}{UB} & Gap$_{\text{LB}}$ & Time to & Time \\ 
	& & & ($\%$) & & & ($\%$) & (s) & & & & & ($\%$) & BKS (s) & (s) \\
	\hline
	\endhead
	\hline
	\multicolumn{15}{r}{\textit{Continued on next page}} \\
	\endfoot
	\endlastfoot
		1 & 38452.8 & 39775 & 3.44 &  & 39775 & 3.44 & 37.0 &  & 39209 &  & 39217.5 & 1.99 & 3.8 & 600.0 \\ 
		3 & 31481.6 & 32492 & 3.21 &  & 32492 & 3.21 & 12.3 &  & 31884 &  & 31891.6 & 1.30 & 2.2 & 600.0 \\ 
		5 & 41355.9 & 43025 & 4.04 &  & 43025 & 4.04 & 22.5 &  & 42155 &  & 42177.3 & 1.99 & 1.2 & 600.0 \\ 
		7 & 45553.2 & 46997 & 3.17 &  & 46997 & 3.17 & 44.6 &  & 46523 &  & 46527.8 & 2.14 & 11.2 & 600.0 \\ 
		9 & 41775.6 & 42846 & 2.56 &  & 42846 & 2.56 & 12.4 &  & 42308 &  & 42350.0 & 1.37 & 2.5 & 600.0 \\ 
		11 & 9912.7 & 10890 & 9.86 &  & 10890 & 9.86 & 113.4 &  & 10447 &  & 10538.9 & 6.32 & 27.5 & 600.0 \\ 
		13 & 12624.3 & 14257 & 12.93 &  & 14257 & 12.93 & 604.7 &  & 13457 &  & 13537.3 & 7.23 & 11.0 & 600.0 \\ 
		15 & 7758.6 & 8788 & 13.27 &  & 8788 & 13.27 & 171.6 &  & 8421 &  & 8504.7 & 9.62 & 45.9 & 600.0 \\
\end{longtable}
}
\end{center}

\clearpage
\subsection{$R||\sum w^\prime_jE_j + w_jT_j$}
\label{app4}

Detailed results obtained for the instances of \cite{SenBulbul2015} (see Tables \ref{RwET40_2}-\ref{RwET200_5}).

\begin{center}
{\scriptsize
\onehalfspacing
\setlength{\tabcolsep}{0.5mm}
\setlength{\LTcapwidth}{10in}
\begin{longtable}{ccccccccccccccc}
  \caption{Results for $R||\sum w^\prime_jE_j + w_jT_j$ with $40$ jobs and $2$ machines}\\
	\hline
	\multirow{4}{*}{Inst}& \multicolumn{7}{c}{\cite{SenBulbul2015}} &  & \multicolumn{6}{c}{UILS}\\
	\cline{2-8}\cline{10-15}
	& \multicolumn{3}{c}{Best} &  & \multicolumn{3}{c}{PR+Benders+SiPSi} & & Best & & \multicolumn{4}{c}{Average} \\
	\cline{2-4} \cline{6-8} \cline{10-10} \cline{12-15}
	& \multirow{2}{*}{LB} & \multirow{2}{*}{UB} & Gap$_{\text{LB}}$ & & \multirow{2}{*}{UB} & Gap$_{\text{LB}}$ & Time & & \multirow{2}{*}{UB} &  & \multirow{2}{*}{UB} & Gap$_{\text{LB}}$ & Time to & Time \\ 
	& & & ($\%$) & & & ($\%$) & (s) & & & & & ($\%$) & BKS (s) & (s) \\
	\hline
	\endfirsthead
	\caption[]{(\textit{Continued})}\\
	\hline
	\multirow{4}{*}{Inst}& \multicolumn{7}{c}{\cite{SenBulbul2015}} &  & \multicolumn{6}{c}{UILS}\\
	\cline{2-8}\cline{10-15}
	& \multicolumn{3}{c}{Best} &  & \multicolumn{3}{c}{PR+Benders+SiPSi} & & Best & & \multicolumn{4}{c}{Average} \\
	\cline{2-4} \cline{6-8} \cline{10-10} \cline{12-15}
	& \multirow{2}{*}{LB} & \multirow{2}{*}{UB} & Gap$_{\text{LB}}$ & & \multirow{2}{*}{UB} & Gap$_{\text{LB}}$ & Time & & \multirow{2}{*}{UB} &  & \multirow{2}{*}{UB} & Gap$_{\text{LB}}$ & Time to & Time \\ 
	& & & ($\%$) & & & ($\%$) & (s) & & & & & ($\%$) & BKS (s) & (s) \\
	\hline
	\endhead
	\hline
	\multicolumn{15}{r}{\textit{Continued on next page}} \\
	\endfoot
	\endlastfoot
		1 & 38605.2 & 38609 & 0.01 &  & 38942 & 0.87 & 16.3 &  & 38609 &  & 38609.0 & 0.01 & 2.9 & 32.7 \\ 
		3 & 35400.4 & 35447 & 0.13 &  & 35688 & 0.81 & 19.3 &  & 35442 &  & 35443.4 & 0.12 & 8.3 & 38.8 \\ 
		5 & 37499.9 & 37574 & 0.20 &  & 37639 & 0.37 & 14.2 &  & 37574 &  & 37574.6 & 0.20 & 16.8 & 28.7 \\ 
		7 & 52040.8 & 52046 & 0.01 &  & 52514 & 0.91 & 17.8 &  & 52046 &  & 52046.0 & 0.01 & 2.7 & 28.8 \\ 
		9 & 37147.6 & 37150 & 0.01 &  & 37512 & 0.98 & 14.6 &  & 37150 &  & 37150.0 & 0.01 & 2.4 & 27.3 \\ 
		11 & 29243.7 & 29333 & 0.31 &  & 29783 & 1.84 & 22.8 &  & 29330 &  & 29330.0 & 0.30 & 0.9 & 25.3 \\ 
		13 & 29267.8 & 29270 & 0.01 &  & 29438 & 0.58 & 24.4 &  & 29270 &  & 29270.0 & 0.01 & 6.9 & 34.0 \\ 
		15 & 27727.0 & 27727 & 0.00 &  & 27727 & 0.00 & 28.2 &  & 27727 &  & 27727.0 & 0.00 & 1.7 & 27.5 \\
\end{longtable}
}
\end{center}

\begin{center}
{\scriptsize
\onehalfspacing
\setlength{\tabcolsep}{0.5mm}
\setlength{\LTcapwidth}{10in}
\begin{longtable}{ccccccccccccccc}
  \caption{Results for $R||\sum w^\prime_jE_j + w_jT_j$ with $60$ jobs and $2$ machines}\\
	\hline
	\multirow{4}{*}{Inst}& \multicolumn{7}{c}{\cite{SenBulbul2015}} &  & \multicolumn{6}{c}{UILS}\\
	\cline{2-8}\cline{10-15}
	& \multicolumn{3}{c}{Best} &  & \multicolumn{3}{c}{PR+Benders+SiPSi} & & Best & & \multicolumn{4}{c}{Average} \\
	\cline{2-4} \cline{6-8} \cline{10-10} \cline{12-15}
	& \multirow{2}{*}{LB} & \multirow{2}{*}{UB} & Gap$_{\text{LB}}$ & & \multirow{2}{*}{UB} & Gap$_{\text{LB}}$ & Time & & \multirow{2}{*}{UB} &  & \multirow{2}{*}{UB} & Gap$_{\text{LB}}$ & Time to & Time \\ 
	& & & ($\%$) & & & ($\%$) & (s) & & & & & ($\%$) & BKS (s) & (s) \\
	\hline
	\endfirsthead
	\caption[]{(\textit{Continued})}\\
	\hline
	\multirow{4}{*}{Inst}& \multicolumn{7}{c}{\cite{SenBulbul2015}} &  & \multicolumn{6}{c}{UILS}\\
	\cline{2-8}\cline{10-15}
	& \multicolumn{3}{c}{Best} &  & \multicolumn{3}{c}{PR+Benders+SiPSi} & & Best & & \multicolumn{4}{c}{Average} \\
	\cline{2-4} \cline{6-8} \cline{10-10} \cline{12-15}
	& \multirow{2}{*}{LB} & \multirow{2}{*}{UB} & Gap$_{\text{LB}}$ & & \multirow{2}{*}{UB} & Gap$_{\text{LB}}$ & Time & & \multirow{2}{*}{UB} &  & \multirow{2}{*}{UB} & Gap$_{\text{LB}}$ & Time to & Time \\ 
	& & & ($\%$) & & & ($\%$) & (s) & & & & & ($\%$) & BKS (s) & (s) \\
	\hline
	\endhead
	\hline
	\multicolumn{15}{r}{\textit{Continued on next page}} \\
	\endfoot
	\endlastfoot
		1 & 91481.0 & 92240 & 0.83 &  & 92240 & 0.83 & 72.4 &  & 91651 &  & 91651.0 & 0.19 & 2.1 & 203.3 \\ 
		3 & 86233.9 & 87132 & 1.04 &  & 87132 & 1.04 & 62.9 &  & 86571 &  & 86589.4 & 0.41 & 3.6 & 222.3 \\ 
		5 & 83874.8 & 84729 & 1.02 &  & 84729 & 1.02 & 46.8 &  & 84163 &  & 84163.0 & 0.34 & 1.2 & 173.6 \\ 
		7 & 79950.6 & 80538 & 0.73 &  & 80599 & 0.81 & 40.8 &  & 80144 &  & 80183.7 & 0.29 & 3.4 & 199.4 \\ 
		9 & 77645.3 & 77653 & 0.01 &  & 77663 & 0.02 & 28.7 &  & 77651 &  & 77733.2 & 0.11 & 160.4 & 185.4 \\ 
		11 & 51203.9 & 51799 & 1.16 &  & 51799 & 1.16 & 116.9 &  & 51403 &  & 51407.2 & 0.40 & 3.1 & 183.8 \\ 
		13 & 50924.0 & 50924 & 0.00 &  & 50962 & 0.07 & 171.7 &  & 50924 &  & 50924.0 & 0.00 & 7.0 & 170.2 \\ 
		15 & 52080.2 & 53245 & 2.24 &  & 53245 & 2.24 & 254.5 &  & 52686 &  & 52703.6 & 1.20 & 6.6 & 211.9 \\
\end{longtable}%
}
\end{center}


\end{appendices}

\end{document}